%% file: 0-main.tex
\documentclass{article} 
\usepackage{iclr2026_conference,times}

\input{math_commands.tex}

\usepackage{hyperref}
\usepackage{url}

\title{On the Surprising Effectiveness of a Single Global Merging in Decentralized Learning}

\author{
Tongtian Zhu\textsuperscript{1*}, \ Tianyu Zhang\textsuperscript{2,3*}, \ Mingze Wang\textsuperscript{4}, 
\textbf{Zhanpeng Zhou\textsuperscript{5†}, \ Can Wang\textsuperscript{1}} \\[2ex]
\textsuperscript{1}Zhejiang University \ \textsuperscript{2}Mila, Quebec AI Institute \ \textsuperscript{3}Université de Montréal \\
\textsuperscript{4}Peking University \ \textsuperscript{5}Shanghai Jiao Tong University \\[2ex]
\texttt{\{raiden, wcang\}@zju.edu.cn} \qquad
\texttt{tianyu.zhang@mila.quebec} \\
\texttt{mingzewang@stu.pku.edu.cn} \qquad \ \ 
\texttt{zzp1012@sjtu.edu.cn}
\vphantom{\thanks{Equal contribution. \textsuperscript{†}Corresponding author.}}
}

\usepackage[utf8]{inputenc} 
\usepackage[T1]{fontenc}    
\usepackage{hyperref}       
\usepackage{url}            
\usepackage{booktabs}       
\usepackage{amsfonts}       
\usepackage{nicefrac}       
\usepackage{microtype}      
\usepackage{xcolor}         

\usepackage{xcolor}
\usepackage{colortbl}
\usepackage{hyperref}
\usepackage{url}
\usepackage{dsfont}

\usepackage{amsmath}
\usepackage{amssymb}
\usepackage{mathtools}
\usepackage{amsthm}

\usepackage[shortlabels]{enumitem} 
\setlist[itemize]{topsep=0em, itemsep=0em, partopsep=0em, parsep=0.5em}

\allowdisplaybreaks
\usepackage{mdframed}
\usepackage[normalem]{ulem}
\usepackage[most]{tcolorbox}   

\usepackage{tikz}
\usetikzlibrary{backgrounds}
\usetikzlibrary{arrows,shapes}
\usetikzlibrary{tikzmark}
\usetikzlibrary{calc}

\usepackage{graphicx}
\usepackage{svg}
\usepackage{microtype}
\usepackage{booktabs} 

\usepackage{caption}
\usepackage{subcaption}

\usepackage{booktabs, multicol, multirow, diagbox}

\usepackage{fontawesome5}

\usepackage{footmisc}

\newcommand{\orange}{\color{orange}}

\colorlet{LightBlue}{blue!39!white} 
\colorlet{DarkBlue}{blue!70!black} 
\colorlet{VeryLightBlue}{blue!30!white} 
\colorlet{LightRed}{red!35!white} 

\definecolor{darkgrey}{rgb}{0.53,0.53,0.53}
\definecolor{middlegrey}{rgb}{0.75,0.75,0.75}
\definecolor{mygrey}{rgb}{0.9,0.9,0.9}
\definecolor{mydarkblue}{rgb}{0,0.08,0.45}
\definecolor{darkdarkblue}{rgb}{0.0,0.0,0.3}
\definecolor{darkblue}{rgb}{0.0,0.0,0.7}
\definecolor{darkred}{rgb}{0.4,0,0.3}
\definecolor{lightblue}{HTML}{F9FEFE}
\definecolor{verylightpurple}{HTML}{FBFAFF}
\definecolor{lightred}{HTML}{FFFAFA}
\definecolor{fancyblue}{HTML}{4771E3}
\definecolor{grey}{rgb}{0.95,0.95,0.95}
\definecolor{myred}{HTML}{7A1410}
\definecolor{lightpurple}{HTML}{ECE5F3}
\definecolor{myorange}{HTML}{FFDD81}
\definecolor{mypink}{HTML}{ffaec9}

\definecolor{boxcyan}{RGB}{225, 245, 250}
\definecolor{textcyan}{RGB}{0,78,181}
\definecolor{boxpink}{RGB}{250, 225, 245}
\definecolor{textpink}{RGB}{120, 0, 100}

\hypersetup{
    colorlinks=true,       
    linkcolor=darkred,        
    citecolor=darkblue,  
    filecolor=darkblue,
    urlcolor=darkblue       
}

\usepackage{algorithm}
\usepackage{algorithmic} 

\renewcommand{\algorithmiccomment}[1]{\hfill $\triangleright$ #1}

\usepackage[capitalize,nameinlink]{cleveref}

\usepackage{aliascnt} 

\theoremstyle{plain}
\newtheorem{theorem}{Theorem}

\newaliascnt{proposition}{theorem}
\newtheorem{proposition}[proposition]{Proposition}
\aliascntresetthe{proposition} 
\newaliascnt{lemma}{theorem}
\newtheorem{lemma}[lemma]{Lemma}
\aliascntresetthe{lemma}
\newaliascnt{corollary}{theorem}
\newtheorem{corollary}[corollary]{Corollary}
\aliascntresetthe{corollary}

\newtheorem{definition}{Definition}

\newtheorem{assumption}{Assumption}

\theoremstyle{definition}
\newtheorem{remark}{Remark}

\newenvironment{inequality}
    {\crefalias{equation}{inequality}\begin{equation}}
    {\end{equation}}

\crefname{definition}{Definition}{Definitions}
\crefname{assumption}{Assumption}{Assumptions}
\crefname{theorem}{Theorem}{Theorems}
\crefname{remark}{Remark}{Remarks}
\crefname{lemma}{Lemma}{Lemmas}
\crefname{corollary}{Corollary}{Corollaries}
\crefname{proposition}{Proposition}{Propositions}
\crefname{section}{Section}{Sections}
\crefname{subsection}{Subsection}{Subsections}
\crefname{example}{Example}{Examples}
\crefname{table}{Table}{Tables}
\crefname{problem}{Problem}{Problems}
\crefname{algorithm}{Algorithm}{Algorithms}
\crefname{figure}{Figure}{Figures}
\crefname{property}{Property}{Properties}

\crefformat{equation}{#2Equation~(#1)#3}
\crefformat{inequality}{#2Inequality~{}(#1){}#3}
\crefrangeformat{inequality}{#3Inequality~(#1)#4--#5(#2)#6}
\Crefformat{section}{#2Section~#1#3}
\Crefformat{page}{#2Page~#1#3}

\newcommand{\acref}[1]{\hyperref[#1]{Appendix~\ref*{#1}}}

\newcommand{\bluemain}[1]{\textcolor[HTML]{3B87EA}{#1}}
\newcommand{\orangemain}[1]{\textcolor[HTML]{CE5319}{#1}}
\newcommand{\greenmain}[1]{\textcolor[HTML]{4B9929}{#1}}

\usepackage[textsize=tiny]{todonotes}

\usepackage{pifont}

\iclrfinalcopy 
\begin{document}

\maketitle

\input{Section/0-Abstract}
\input{Section/1-Introduction}
\input{Section/2-Related_work}

\input{Section/3-Preliminaries}
\input{Section/4-MainDiscovery}

\input{Section/5-Theory}
\input{Section/8-Implications}
\bibliography{iclr2026_conference}
\bibliographystyle{iclr2026_conference}
\input{Section/Appendix}


\end{document}

%% file: math_commands.tex
\usepackage{amsmath,amsfonts,bm}

\def\eqref#1{equation~\ref{#1}}

\def\1{\bm{1}}

\DeclareMathAlphabet{\mathsfit}{\encodingdefault}{\sfdefault}{m}{sl}
\SetMathAlphabet{\mathsfit}{bold}{\encodingdefault}{\sfdefault}{bx}{n}

\DeclareMathOperator{\Tr}{Tr}

%% file: Section/0-Abstract.tex
\begin{abstract}

Decentralized learning provides a scalable alternative to parameter-server-based training, yet its performance is often hindered by limited peer-to-peer communication. 
In this paper, we study how communication should be scheduled over time, including determining when and how frequently devices synchronize. 
Counterintuitive empirical results show that concentrating communication budgets in the later stages of decentralized training remarkably improves global test performance.
Surprisingly, we uncover that fully connected communication at the final step, implemented by a single global merging, can significantly improve the performance of decentralized learning under high data heterogeneity. 
Our theoretical contributions, which explain these phenomena, are the first to establish that the globally merged model of decentralized SGD can match the convergence rate of parallel SGD.
Technically, we reinterpret part of the discrepancy among local models, which were previously considered as detrimental noise, as constructive components essential for matching this rate. 
This work provides evidence that  decentralized learning is able to generalize under high data heterogeneity and limited communication, while offering broad new avenues for model merging research. 
Blog post and code are available at \href{https://paper-list.notion.site/ICLR-26-Oral-The-Grokking-Moment-in-Decentralized-Learning-On-The-Surprising-Effectiveness-of-A--2f43218102c0805d99d6e56d2934fac4}{\faLink~Grokking in Decentralized Learning} and \href{https://github.com/Raiden-Zhu/ICLR-2026-Grokking-in-Decentralized-Learning}{\faGithub~Code}, respectively.
\end{abstract}

%% file: Section/1-Introduction.tex
\section{Introduction}
\label{sec: introduction}


Decentralized learning offers a promising approach to crowdsource computational workloads across geographically distributed compute \citep{yuan2022decentralized,NEURIPS2023_28bf1419,jaghouar2024intellect}.\
A defining characteristic of this setting is the reliance on peer-to-peer communication during training, involving the peer-level exchange of model parameters or gradients.
However, such communication is often  constrained in practice due to limited bandwidth between geographically distant nodes, making it a scarce resource. 
These constraints can significantly degrade the performance of decentralized learning, both theoretically and empirically \citep{NIPS2017_f7552665, pmlr-v119-koloskova20a, vogels2021relaysum}. As a result, efficiently allocating limited communication resources becomes a fundamental challenge in decentralized learning, especially in heterogeneous environments where varying local data distributions intensify communication demands \citep{10251949}.

To date, most efforts addressing this challenge have focused on optimizing communication allocation at the \textit{spatial level}, particularly through the design of communication graphs \citep{ying2021exponential, Li_2022_CVPR, takezawa2023beyond, kharrat2024decentralized}.
In contrast, the \textit{temporal} allocation of communication, i.e., deciding when and how frequently agents synchronize with others, remains a significant yet underexplored direction for improving decentralized learning.
Although temporal communication allocation has been studied in federated learning (FL) \citep{tang2020communication}, this problem remains largely untouched in the fully decentralized setting, which is fundamentally different due to the lack of a central server for global aggregation (see discussions in \cref{sec: related_work} and \cref{remark: global_test}).

\begin{tcolorbox}[notitle, rounded corners, colframe=middlegrey, colback=lightblue, 
       boxrule=2pt, boxsep=0pt, left=0.15cm, right=0.17cm, enhanced, 
       toprule=2pt,
    ]
\textbf{Question}: \textit{\fontsize{9.5pt}{12pt}\selectfont How to allocate communication budget in decentralized learning over temporal levels?}
\end{tcolorbox}

To answer this question, we design a series of experiments that allocate communication budgets across different time windows during training (see \cref{fig: sliding_window_cifar100}). 
Specifically, we divide the training process into consecutive windows, each with of a fixed number of communication rounds. We assign higher communication budgets to selected windows using global synchronization via \texttt{AllReduce} \citep{sergeev2018horovod}, while keeping communication low otherwise by infrequent synchronization with random peer agents.\footnote{Agents refer to participants in decentralized learning.  ``Communication'' and ``synchronization'' are used interchangeably.}
This reveals how the temporal communication allocation affects performance under constrained budgets.
We observe that allocating higher communication budgets toward the later stages of training consistently leads to improved final test performance (see \cref{def: avg_gen}). 
More surprisingly, we observe the remarkable effect of a single round of fully-connected communication.\footnote{Fully-connected communication refers to global synchronization via \texttt{AllReduce}. In this paper, fully-connected communication is realized through parameter averaging over the models on all agents, namely \emph{global merging}.}

\begin{tcolorbox}[notitle, rounded corners, colframe=middlegrey, colback=lightblue, 
       boxrule=2pt, boxsep=0pt, left=0.15cm, right=0.17cm, enhanced, 
       toprule=2pt,
    ]
\textbf{Surprising Phenomenon}: \textit{\fontsize{9.5pt}{12pt}\selectfont A single global merging of decentralized models, under severely constrained communication and high data heterogeneity, can significantly improve global test performance.}
\end{tcolorbox}

\begin{figure}[t!]
    \centering

    \begin{subfigure}{.32\textwidth}
        \centering
        \includegraphics[width=\linewidth]{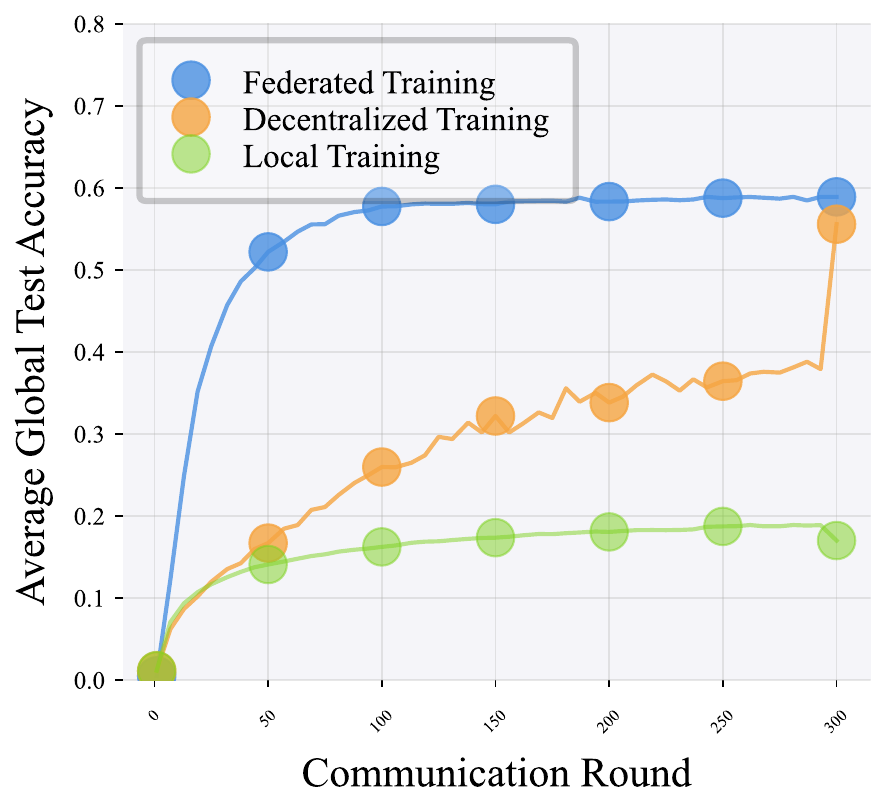}
        \caption{CLIP ViT-B/32}
        \label{fig: main_clip}
    \end{subfigure}
    \hfill
    \begin{subfigure}{.32\textwidth}
        \centering
        \includegraphics[width=\linewidth]{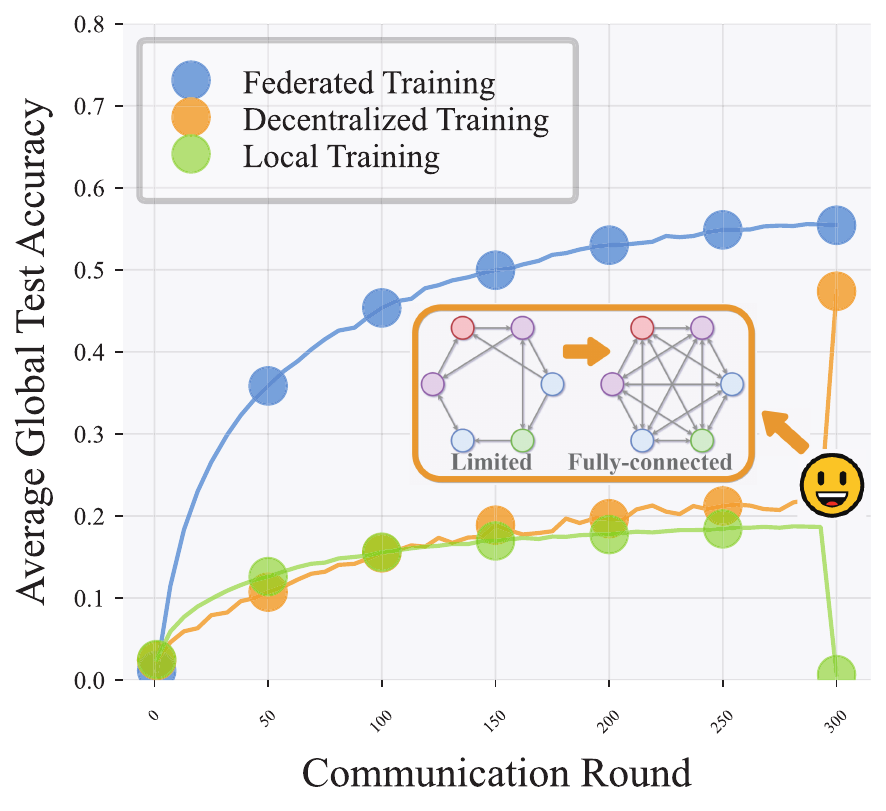}
        \caption{ResNet-18 (w/o pretraining)}
        \label{fig: main_resnet}
    \end{subfigure}
    \hfill
    \begin{subfigure}{.31\textwidth}
        \centering
        \includegraphics[width=\linewidth]{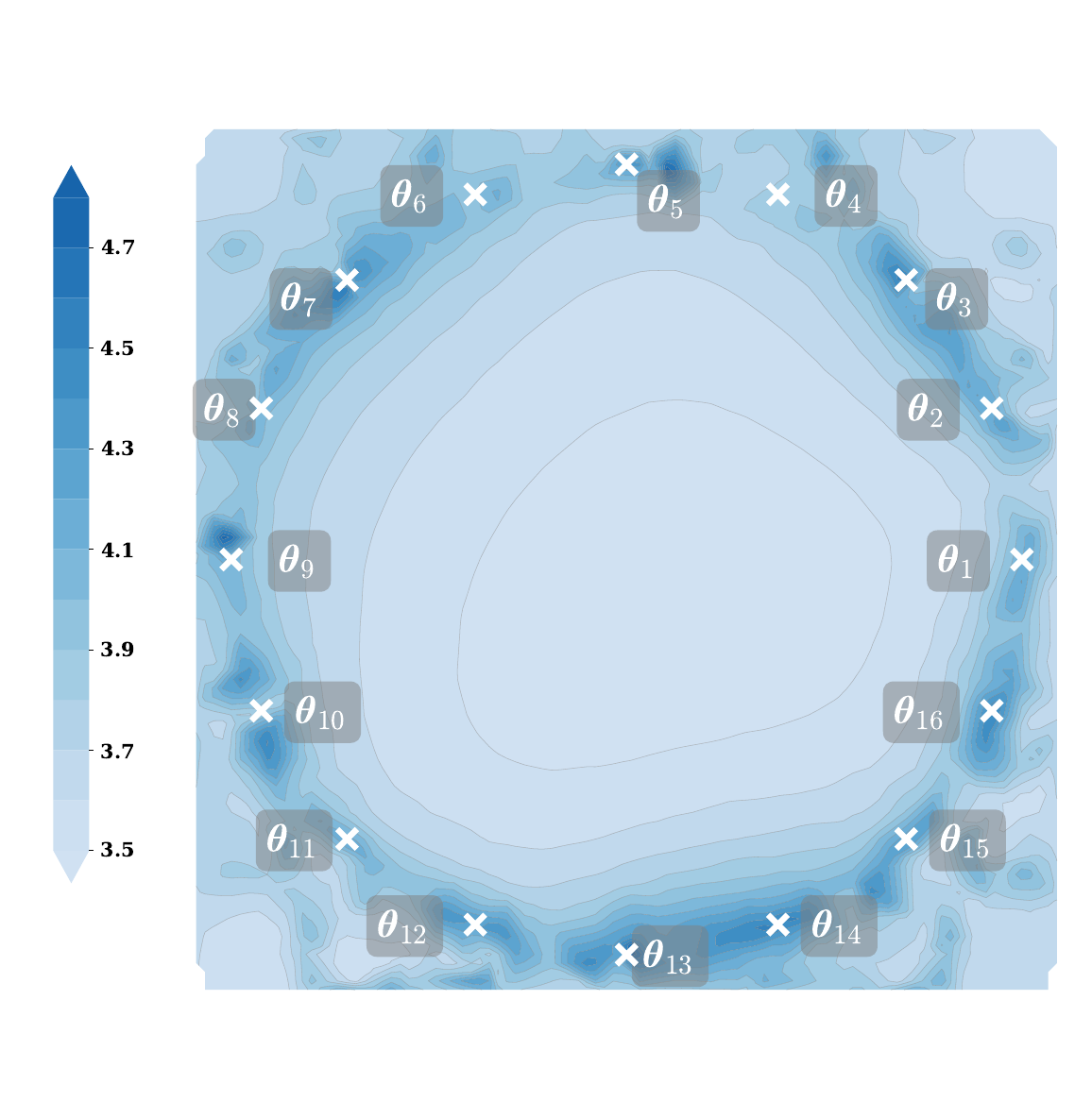}
        \caption{Landscape before final merging}
        \label{fig: resnet_landscape}
    \end{subfigure}
    
    \vspace{1em} 

    \begin{subfigure}{\textwidth}
        \centering
        \includegraphics[width=0.96\linewidth]{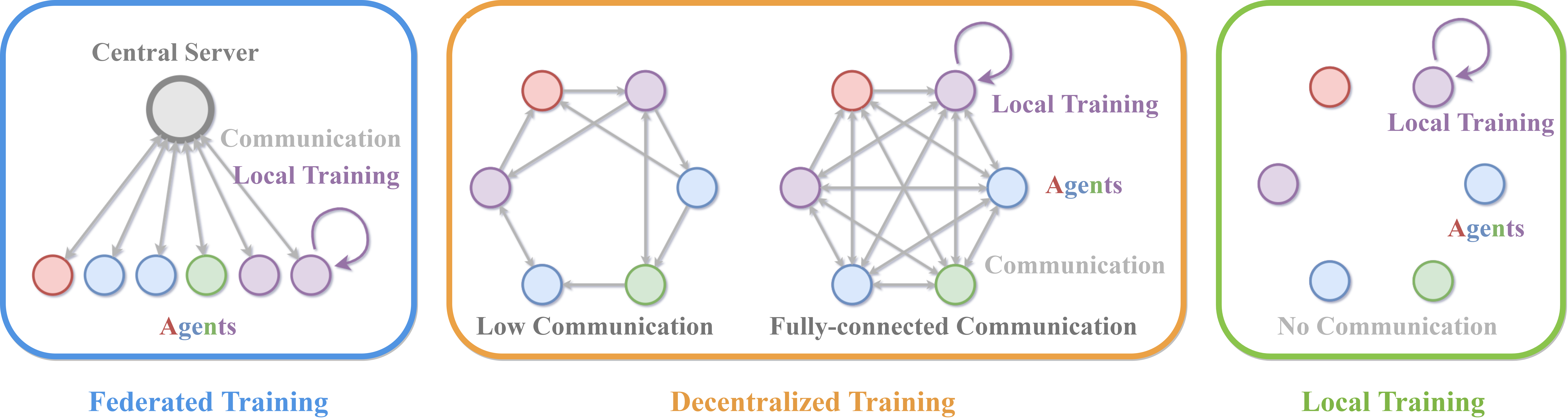}
        \caption{A comparative illustration of federated, decentralized, and local training.}
        \label{fig: topology}
    \end{subfigure}
    \caption{
    \textbf{(a, b)} Global test accuracy (see \cref{def: avg_gen}) of CLIP ViT-B/32 (a) and ResNet-18 (b) on Tiny ImageNet under a 32-agent non-IID setup (Dirichlet $\alpha=0.1$), where decentralized SGD communicates with one random peer per round with probability 0.2 and performs a final global merge.
    \textbf{(c)} Loss landscape before the final merge for decentralized SGD.\protect\footnotemark
    \textbf{(d)} Illustration of federated, decentralized, and local training.
    Experimental details are provided in \acref{sec: setup}.
    }
    \label{fig: main_combined}
\end{figure}

\footnotetext{We use $16$ agents for loss landscape visualization to ensure visual clarity.}

\textbf{Our Contributions} are summarized below.

\begin{itemize}[leftmargin=*]
    \item \textbf{Empirical Observations}. 
    \textbf{(1)}: We highlight the critical role of a single global merging in decentralized training, showing that it can achieve performance close to federated learning, even under severe communication constraints and data heterogeneity (see \cref{fig: main_clip}, \cref{fig: main_resnet}). 
    \textbf{(2)}: We observe that limited but non-zero communication preserves a challenging cross-initialization, cross-distribution ``mergeability'' of local models throughout training (see \cref{def:mergeable}, \cref{fig: resnet_landscape}, and the blue curve in \cref{fig: sliding_window_cifar100_Average}), which does not hold under complete local training (green curve in \cref{fig: main_clip}, \cref{fig: main_resnet}). These findings remain consistent across datasets, heterogeneity levels, model architectures, and communication topologies (see \acref{sec: add_exp}), and provide a first systematic study of global merging in decentralized learning.
    \item \textbf{Theoretical Contributions}. We investigate the underlying mechanism that enables the \emph{mergeability} of local models in decentralized learning. Specifically, we provide the first convergence analysis showing that the globally merged model of decentralized SGD can match the rate of parallel SGD 
    (\cref{thm:nonconvex-convergence-descent-lemma} and \cref{prop: critical}). 
    Furthermore, we offer a theoretical explanation for  why limited but nonzero communication can ensure mergeability, and why communication should be concentrated in the later stages of training (see \cref{prop: sufficient}).
\end{itemize}

We anticipate that this work will pave the way for principled decentralized training algorithms capable of generalizing under communication constraints and data heterogeneity, while also advancing model merging research (see discussions in \cref{sec: implications} and insights in the \textbf{Q\&A Section}, \acref{sec: QA}).

%% file: Section/2-Related_work.tex
\section{Related Work}
\label{sec: related_work}

\textbf{Temporal Communication Allocation in Parallel, Federated, and Decentralized Learning.}
Communication allocation is well-studied in both data-centric parallel learning \citep{li2014communication}, and Federated Learning (FL) \citep{pmlr-v54-mcmahan17a}. 
In parallel learning settings, \citet{gu2024a} proposed a novel strategy for scheduling local steps by analyzing the implicit bias of Local SGD \citep{localsgdgeneralizes2023}. FL extends this server-based paradigm to handle not identically and independently distributed (non-IID) data, but it critically retains a global model. This reliance on a global model has shaped a broad consensus in the FL literature: frequent, early-stage communication is considered essential for aligning local models  \citep{8664630, tang2020communication}.

In contrast, our work addresses fully decentralized learning, a fundamentally different setting that lacks a central server. 
Instead of optimizing a generic global model, the goal is to make local models generalize to the global distribution. 
Despite extensive work focusing on communication allocation at the spatial level in decentralized learning (e.g., designing communication topologies) \citep{ying2021exponential, Li_2022_CVPR, takezawa2023beyond, kharrat2024decentralized}, few studies examine temporal allocation; a pioneering IID study \citep{pmlr-v139-kong21a} showed that stronger early alignment to the global average can modestly improve test performance. 
These findings do not directly transfer to non-IID settings, where $\mathcal{L}(\cdot)\equiv\mathcal{L}_k(\cdot)$ no longer holds (see \cref{eq:val_loss} and \cref{def: avg_local_gen_app}), so they mainly characterize local rather than global generalization (see \cref{def: avg_gen}).
Due to space constraints, we refer readers to \acref{sec:implicit_bias} and \acref{sec:model_merging} for related work on the implicit bias of decentralized learning, and on the topic of model merging.

%% file: Section/3-Preliminaries.tex
\section{Notations and Preliminaries}
\label{sec: preliminaries}


\subsection{Non-IID Decentralized Learning}
Decentralized learning formalizes distributed learning as an optimization problem over a connected graph $G = (\mathcal{V}, \mathcal{E})$, where $\mathcal{V}$ contains $m$ agents and $\mathcal{E}$ denotes the communication links. Each agent $k \in \mathcal{V}$ samples data from a local distribution $\mathcal{D}_k$ and maintains a local model $\theta_k \in \mathbb{R}^d$. The objective is to learn a consensus model $\theta$ that minimizes the global population risk \citep{pmlr-v119-koloskova20a}:
\begin{align}\label{eq:val_loss}
\min_{\theta \in \mathbb{R}^d} \left[ \mathcal{L}(\theta) \triangleq \frac{1}{m}\sum_{k \in \mathcal{V}} \mathbb{E}_{\xi_k \sim \mathcal{D}_k} \mathcal{L}(\theta; \xi_k) \right],
\end{align}
where $\mathbb{E}_{\xi_k \sim \mathcal{D}_k} \mathcal{L}(\theta; \xi_k) \triangleq \mathcal{L}_k(\theta)$ denotes the local population risk of $\theta$ on unseen instance ${\xi_k \sim \mathcal{D}_k}$.

In practice, the optimization of \cref{eq:val_loss} is performed under the empirical risk minimization framework, leveraging $m$ local datasets  $S \triangleq\bigcup_{k=1}^{m} S_k$, where $S_k=$ $\left\{\xi_{k,1}, \ldots, \xi_{k,\zeta}\right\}$ denotes the dataset of agent $k$ sampled from $\mathcal{D}_k$. The resulting optimization problem is given by:
\begin{align}\label{eq:training_loss}
\min_{\theta \in \mathbb{R}^d} \left[ \mathcal{L}_\mathcal{S}(\theta) \triangleq \frac{1}{m}\sum_{k \in \mathcal{V}} \sum_{\zeta=1}^{n_k} \mathcal{L}(\theta; \xi_{k,\zeta}) \right].
\end{align}
To solve the optimization problem in \cref{eq:training_loss}, decentralized algorithms minimize the global empirical risk with only local computations and peer-to-peer communication \citep{1104412, nedic2009distributed}.
The communication graph is governed by a  weighted adjacency matrix $W^{(t)} \in [0,1]^{m \times m}$, sampled from a distribution $\mathcal{W}^{(t)}$, where each entry $W^{(t)}_{k,l} \geq 0$ reflects the influence of agent $l$ on agent $k$.\footnote{Our framework incorporates a randomized decentralized learning setting where the weighted adjacency matrix $W^{(t)}$ can change during training \citep{1638541, pmlr-v119-koloskova20a, vos2023epidemic}.} 
Decentralized learning algorithms operate by alternating between local updates and model aggregation through communication with neighbors, as outlined in \cref{alg:decentralized_learning}. 


\begin{algorithm}[ht]
	\caption{Decentralized Learning}\label{alg:decentralized_learning}
	\let\oldfor\algorithmicfor
	\renewcommand{\algorithmicfor}{\textbf{in parallel on all agents $k \in \mathcal{V}$, for}}
	\let\oldendfor\algorithmicendfor
	\renewcommand{\algorithmicendfor}{\algorithmicend\ \textbf{parallel for}}
	\begin{algorithmic}[1]
		\INPUT{Initialize values $\theta_k^{(0)} \in \mathbb{R}^d$ on each agent $k \in \mathcal{V}$,
       number of steps $T$, mixing matrix $W$}\\[1ex]
		\FOR{$t=0,\dots, T-1$}
            \STATE Sample training data $\xi_{k}^{(t)}$ from $\mathcal{D}_k$,
            $\theta_k^{(t+\frac{1}{2})} \gets \textit{Optimizer}(\theta_k^{(t)}, \xi_{k}^{(t)})$ \hfill $\triangleright$ Local update
            \STATE Send $\theta_k^{(t)}$ to out-neighbor(s) and receive $\{\theta_l^{(t)}\}_{l \in \mathcal{N}_{\text{in}}(k)}$ from in-neighbor(s) \hfill $\triangleright$ Communication
            \STATE Sample mixing matrix ${W}^{(t)} \sim \mathcal{W}^{(t)}$, $\theta_k^{(t+1)} \gets \sum_{l \in \mathcal{N}_{\text{in}}(k)} W^{(t)}_{k,l} \theta_l^{(t+\frac{1}{2})}$ \hfill $\triangleright$ Gossip averaging
		\ENDFOR
	\end{algorithmic}
\end{algorithm}

\textbf{Practical Evaluation Metrics}. 
In decentralized learning, data heterogeneity and limited training time often prevent a full consensus model $\theta$. We therefore propose to adopt \textit{average global test accuracy}, a proxy for global population risk, as the primary metric to quantify how well local models generalize to the global data distribution.
\begin{definition}[Average Global Test Accuracy]\label{def: avg_gen}
 The average accuracy of agents \(\scalebox{0.9}{$k \in \mathcal{V}$}\) is defined as:
\begin{align*}
\underbrace{
\overline{\operatorname{Acc}}(\{\theta_k^{(t)}\}_{k \in \mathcal{V}}) = \frac{1}{m}\sum_{k \in \mathcal{V}}\operatorname{Acc}(\theta_k^{(t)})
}_{\text{Average Accuracy across agents}}
,\quad 
\text{where } 
\operatorname{Acc}(\cdot) \triangleq  
\underbrace{
\frac{1}{m}\sum_{l \in \mathcal{V}} \mathbb{E}_{\xi_l \sim \mathcal{D}_l} \operatorname{Acc}(\cdot; \xi_l)
}_{\text{Test accuracy on the global distribution}}.
\end{align*}
\end{definition}

\begin{tcolorbox}[notitle, rounded corners, colframe=middlegrey, colback=lightred, 
       boxrule=2pt, boxsep=0pt, after skip=7pt, left=0.15cm, right=0.17cm, enhanced, 
       toprule=2pt,
    ]
\begin{remark}[Metric Justification]\label{remark: global_test}
    This metric is specifically designed to address a core question in fully decentralized learning: \textit{how well do {local models $\{\theta_k^{(t)}\}_{k \in \mathcal{V}}$}, trained with limited peer-to-peer synchronization, generalize to the {global data distribution $\mathcal{D}$}?} This metric offers a more realistic evaluation for decentralized settings without a global model. See discussions in \acref{sec: eval}.
\end{remark}
\end{tcolorbox}

\subsection{Mergeability}


\begin{definition}[Mergeability under Global Population Risk]
\label{def:mergeable}
A set of local models $\{\theta_k\}_{k \in V}$ is globally mergeable if there exist combination weights $\{w_k\}_{k \in V} \in [0,1]$ such that:
\begin{equation}
\mathcal{L}\left(\sum_{k \in V} w_k \theta_k\right) \leq \sum_{k \in V} w_k \mathcal{L}(\theta_k),
\end{equation}
where $\mathcal{L}(\cdot)$ denotes the global population risk.
\end{definition}


\cref{def:mergeable} formalizes the intuition that a linearly interpolated model performs no worse than the original local models. 
This Definition is inherently non-trivial due to the \textit{non-convexity} of $\mathcal{L}$. 

%% file: Section/4-MainDiscovery.tex
\section{Empirical Observations}
\label{sec: main_discovery}

\subsection{Increasing Impact of Communication in the Later Stages of Training}

To investigate potential solutions of communication scheduling, we explore a direct strategy: Concentrate communication in a small subset of communication rounds.
To this end, we divide the training process into consecutive windows, each consisting of a fixed length of communication rounds.
Specifically, the communication scheme is as follows: (1) fully-connected communication (see \cref{fig: topology} (b)) is activated only within specific communication windows (i.e., global synchronization via \texttt{AllReduce}~\citep{sergeev2018horovod}\footnote{We note that \texttt{AllReduce} can be efficiently realized in a decentralized manner such as \texttt{Ring-ALLReduce}.}); 
(2) while in all other rounds, each agent communicates only with \textit{one} random peer with a probability of $0.2$ (see ``Communication Graph" in \acref{sec: setup}).\footnote{\textbf{Code}: \url{https://github.com/Raiden-Zhu/ICLR-2026-Grokking-in-Decentralized-Learning}}

\begin{figure*}[t!]
\centering
\begin{subfigure}{.32\textwidth}
    \centering
    \includegraphics[width=\linewidth]{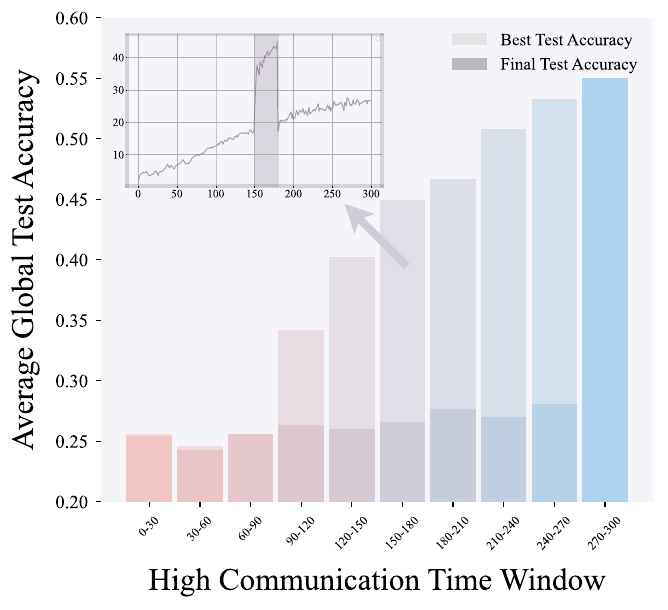}
    \caption{}
    \label{fig: sliding_window_cifar100_10}
\end{subfigure}
\hfill
\begin{subfigure}{.32\textwidth}
    \centering
    \includegraphics[width=\linewidth]{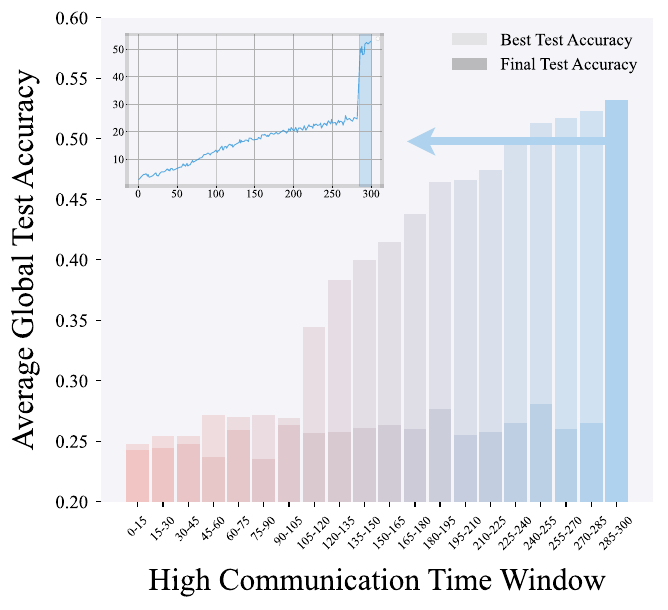}
    \caption{}
    \label{fig: sliding_window_cifar100_20}
\end{subfigure}
\hfill
\begin{subfigure}{.32\textwidth}
    \centering
    \includegraphics[width=\linewidth]{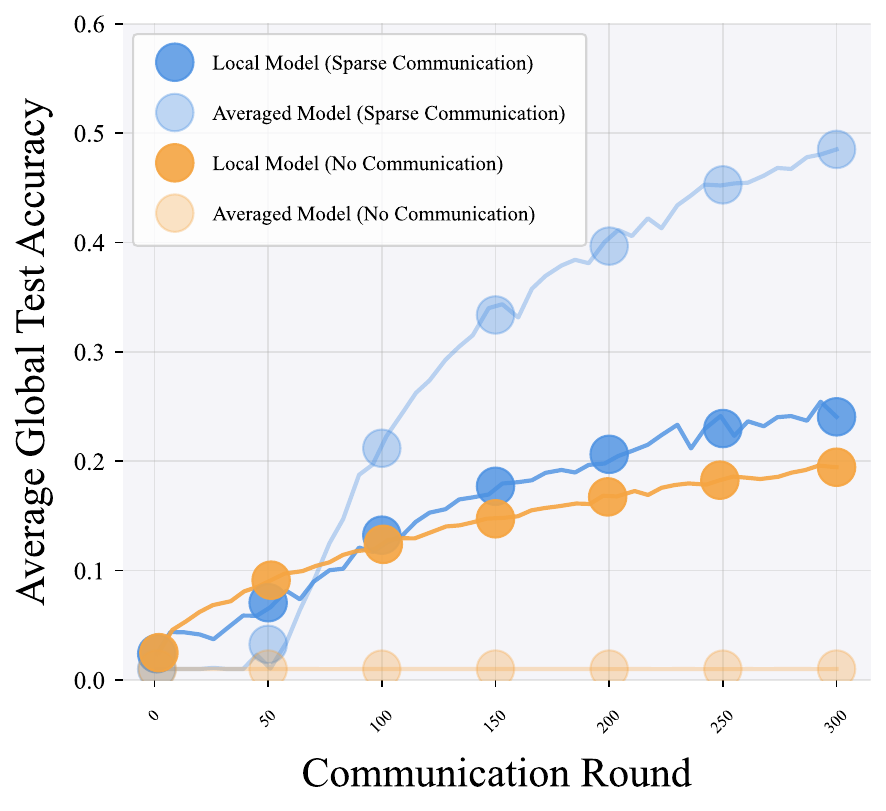}
    \caption{}
    \label{fig: sliding_window_cifar100_Average}
\end{subfigure}
\caption{
\textbf{(a, b)}: Comparisons of global test accuracy (see \cref{def: avg_gen}) in decentralized training of ResNet-18 on CIFAR-100 with AdamW, distributed across 16 agents with Dirichlet $\alpha$ = 0.1 (see details in \acref{sec: setup}). 
Fully-connected communication (i.e., AllReduce) is activated only in specific windows, while low communication with one random peer with a probability of $0.2$ is used elsewhere. 
\textbf{(a)}: Fully-connected communication in $1/10$ of total rounds. 
\textbf{(b)}: Fully-connected communication in $1/20$ of total rounds. 
In both, lighter bars show peak accuracy, darker bars show final accuracy.
\textbf{(c)}: Global test accuracy curves for local models and the globally averaged model (counterfactual) under persistent low communication \bluemain{(blue)} and no communication \orange{(orange)}.\protect\footnotemark
  }
\label{fig: sliding_window_cifar100}
\end{figure*}

As shown in \cref{fig: sliding_window_cifar100}, training is divided into $10$ (a) and $20$ (b) communication windows, respectively. 
The bars in \cref{fig: sliding_window_cifar100} show both the best global test accuracy achieved during training (lighter-colored bars) and the final test accuracy at the end of training (darker-colored bars).
Each bar corresponds to one communication window, where fully connected communication is applied \textit{only} to the rounds within that window, while random peer communication is used in all other rounds. 
For instance, the inset in \cref{fig: sliding_window_cifar100_10} presents the complete test accuracy trajectory when fully-connected communication is applied during rounds 150 to 180. 
A consistent trend emerges: \textit{allocating communication budgets toward the later stages of training yields substantial improvements, particularly in final test accuracy}.

\footnotetext{The term “counterfactual” refers to the fact that no global merging occurs during decentralized training. Instead, we manually compute the test accuracy of the hypothetical globally averaged model to quantify the “mergeability” of local models.}

\subsection{A Single Global Merging Significantly Improves Global Test Performance}
In \cref{fig: sliding_window_cifar100_20}, we reduce the fully-connected communication window length to 10 rounds, yet still observe substantial improvements in global test performance.
This observation naturally raises the question: \textit{What happens if the fully-connected window is reduced to a single round?}

To investigate this, we conduct experiments where fully-connected communication is applied only once, implemented by a single global merging.
As shown in \cref{fig: main_clip} and \cref{fig: main_resnet},  a single global merging is sufficient to significantly improve global test performance. Consistent gains are observed across a wide range of settings, including different datasets, architectures, and communication topologies (see additional experiments in \acref{sec: add_exp}).
The significant increase in performance suggests that the  potential of decentralized learning might be considerably underestimated. 


\textbf{Comparisons.} D-PSGD \citep{NIPS2017_f7552665} introduced final global merging under IID settings, but did not analyze the performance gap before and after merging. In contrast, we provide a systematic study of this recovery in challenging non-IID scenarios. \citet{pmlr-v139-chen21y} showed the benefit of periodic global averaging, but their method requires frequent global communication (every $H=48$ steps), whereas we recover performance with only \textit{a single} merging. We also compare with Skew-Compensated Sparse Push (SCSP) \citep{aketi2021sparse}, which also includes a final global merging step. While both works aim to reduce communication, they differ from ours in methodology and setting: (1 ) SCSP uses \textit{gradient sparsification} (top-$k$ gradients) over a fixed topology, whereas we study mergeability under \textit{topological sparsification} (sparse gossip); (2) SCSP focuses on one local step ($H=1$), whereas we show robust mergeability with multiple local steps (e.g., $H=100$) under high heterogeneity. While these works share the broader goal of improving communication efficiency, our work offers a new perspective by investigating the \textit{mergeability} itself: Why local models retain this property despite extremely limited communication and high data heterogeneity.

\begin{tcolorbox}[notitle, rounded corners, colframe=middlegrey, colback=lightred, 
       boxrule=2pt, boxsep=0pt, after skip=7pt, left=0.15cm, right=0.17cm, enhanced, 
       toprule=2pt,
    ]
\begin{remark}[Non-Triviality of the Performance Gain]
One may attribute the final improvement to the cumulative effect of sparse gossip. In our setup, each communication round activates one random peer exchange with probability $R=0.2$ over $T=300$ rounds, yielding about $60$ peer exchanges per agent in expectation. This may appear analogous, on average, to multiple implicit global aggregations. If this additive interpretation were sufficient to explain the gain, local models should already be better aligned and achieve performance close to that of the post-merge model. However, our empirical results challenge this interpretation: as shown in \cref{fig: main_resnet}, local models under sparse gossip still exhibit poor global test performance, close to the no-communication case, while a single final global merging produces a large performance gain. This gap indicates that the effect of single merging is non-trivial, rather than a simple result of accumulated local gossip.
\end{remark}
\end{tcolorbox}

\textbf{Communication Cost Comparison.}
Let $P$ be the model size, $m$ the number of agents, and $T$ the number of training rounds. A standard \texttt{Ring-AllReduce}-based protocol incurs a total communication cost of $\mathcal{O}(2 m P T)$. In contrast, our decentralized setup has a cost of $\mathcal{O}(m R PT + 2m P)$, where $R\ll 2$ denotes the expected number of peers per round, and the $\mathcal{O}(2m P)$ term arises from final global merging via \texttt{Ring-AllReduce}.
We also note that while a global merging  may appear less practical in some cases, it can be effectively approximated via multiple rounds of gossip synchronization among local agents (see our supplementary experiments in \acref{sec: gossip_merging}). 


\subsection{Mergeability Persists under Limited but Nonzero Communication}\label{sec: mergeability_origin}
A follow-up question is whether the effectiveness of the global merging is specific to the end of training.
To investigate this, we assess the \textit{counterfactual} performance of the globally averaged model at each training round, as depicted by the light-blue curve in \cref{fig: sliding_window_cifar100_Average}.
The experiments are conducted under a lower-communication setting, where each agent communicates with one random peer at each round with probability 0.2 (see “Communication Graph” in \cref{sec: setup}).
A consistent superiority of the merged model (light-blue curve) over the local models (dark-blue curve) is observed throughout training, suggesting that local models remain mergeable at all stages (see \cref{def:mergeable}).

As an ablation, we conduct an experiment where all models are trained locally without any communication (see \cref{fig: sliding_window_cifar100_Average}). In this case, the counterfactual test performance of the globally averaged model remains close to zero (light-orange curve), indicating that without communication, local models are not mergeable. This suggests that mergeability does \textit{not} arise inherently from the local models themselves.
Interestingly, under the low-communication setting, the performance of local models before merging (dark-blue curve) remains similar to that in the no-communication case (dark-orange curve). However, after global merging, the models show significant performance improvement. This clear contrast implies that extremely limited but nonzero communication enables mergeability.

\textbf{Mergeability without Consensus.} Prior work on gossip algorithms has suggested that local models may converge to a similar state even in minimal communication regimes~\citep{10.1145/1082469.1082470}. In contrast, our work addresses a more challenging heterogeneous data setting where we find that local models do not reach a single consensus point, yet remain mergeable. Specifically, we identify an emergent geometric structure where decentralized training guides local models to a ring-like high-loss region surrounding a central low-loss basin (see \cref{fig: resnet_landscape}). Notably, this corresponds to a more challenging cross-initialization, cross-distribution merging scenario, where local models do not start from a shared pretrained checkpoint and are trained on non-IID data distributions.

%% file: Section/5-Theory.tex
\section{Theoretical Analysis}
\label{sec: theory}
In this section, we examine the underlying mechanisms that enable the mergeability of local models in decentralized learning.
As an initial step, we conduct a fine-grained convergence analysis of the globally merged models trained by Decentralized SGD (DSGD).\footnote{DSGD refers to standard decentralized SGD where the optimizer in \cref{alg:decentralized_learning} is replaced with SGD.}
To substantiate the mergeability of local models, we compare the convergence rate of the merged model trained by DSGD with that of parallel SGD. Remarkably, we prove that the merged model in decentralized learning can match the rate of parallel SGD \citep{JMLR:v13:dekel12a, li2014communication}. This supports the empirical findings that the merged model can preserve the performance of individual local models (see \cref{def:mergeable}).

\subsection{Assumptions}
We start by introducing the commonly used assumptions \citep{pmlr-v139-kong21a, pmlr-v119-koloskova20a}.

\begin{assumption}[Mixing matrix]\label{ass: mixing}
Each sample of the (randomized) mixing matrix 
\(W  \in \mathbb{R}^{m \times m}\)
is doubly stochastic. Moreover, there exists \(p > 0\) such that
\begin{align}
\mathbb{E}_{W}\|\Theta W - \bar{\Theta}\|_F^2
\le
(1 - p)\|\Theta - \bar{\Theta}\|_F^2,
\ \forall\, \Theta \in \mathbb{R}^{d \times m}.
\end{align}
Here
${\Theta} = [{\theta}_1,\ldots,{\theta}_m]$,
\(\bar{\Theta} = [\bar{\theta},\ldots,\bar{\theta}]
\equiv 
\Theta\,\tfrac{1}{m}\mathbf{1}\mathbf{1}^\top\)
where
\(\bar{\theta} = \frac{1}{m}\sum_{k=1}^m \theta_k\).
\end{assumption}

\begin{assumption}[Smoothness]\label{ass: regularity}
The objective function \(\mathcal{L}\) is four-times continuously differentiable (i.e., \(\mathcal{L} \in \mathcal{C}^4\)) and there exist constants \(L_q \ge 0\) for \(q \in \{1, \ldots, 4\}\) such that:
\begin{align}
\|\nabla^q \mathcal{L}(\theta)\|
\le
L_q,
\quad
\forall\, \theta \in \mathbb{R}^d.
\end{align}
\end{assumption}
We note that given $\mathcal{L} \in \mathcal{C}^2$, the boundedness of the Hessian norm (i.e., the case \(q=2\)) implies that $\mathcal{L}$ is \(L_2\)-smooth, thereby recovering \cref{ass: l-smooth} with ($L=L_2$).

\begin{assumption}[Bounded noise and diversity]\label{ass: bounded_noise_diversity}
There exist \(\sigma^2, \zeta^2 \ge 0\) such that for any $\theta_k\in\{\theta_k\}_{k=1}^m$:
\begin{align}
\frac{1}{m}\sum_{k=1}^m \mathbb{E}_{\xi_k}
\|\nabla \mathcal{L}_k(\theta_k; \xi_k) - \nabla \mathcal{L}_k(\theta_k)\|_2^2 
\le
\sigma^2, \quad
\frac{1}{m}\sum_{k=1}^m 
\|\nabla \mathcal{L}_k(\theta_k) - \nabla \mathcal{L}(\theta_k)\|_2^2
\le
\zeta^2,
\end{align}
where 
\(\mathcal{L}(\theta) = 
\tfrac{1}{m}\sum_{k=1}^m \mathcal{L}_k(\theta)\). 
\end{assumption}
Here $\sigma$ measures the local noise level and $\zeta$ is a measure of the heterogeneity among agents.

\subsection{Convergence Analysis}

\begin{tcolorbox}[notitle, rounded corners, colframe=middlegrey, colback=lightblue, 
       boxrule=2pt, boxsep=2pt, left=0.2cm, right=0.2cm, top=0.2cm, bottom=0.2cm, enhanced, 
       toprule=2pt
    ]
\begin{theorem}[Non-convex Convergence Rate of DSGD]
\label{thm:nonconvex-convergence-descent-lemma}
Suppose \cref{ass: regularity} and \cref{ass: bounded_noise_diversity} hold. 
Consider decentralized SGD (DSGD) with initializations \(\theta_k^{(0)} = \theta^{(0)}\) for all \(k \in \mathcal{V}\), and a constant learning rate satisfying \(\eta < \tfrac{2}{L_2}\). 
Let \(\bar{\theta}^{(t)} = \tfrac{1}{m}\sum_{k=1}^m \theta_{k}^{(t)}\) denote the averaged model at the \(t\)-th step. 
To achieve an \(\varepsilon\)-stationary point such that 
\(\frac{1}{T}\sum_{t=0}^{T-1} \mathbb{E}\bigl[\|\nabla \mathcal{L}(\bar{\theta}^{(t)})\|_{2}^2\bigr] \le \varepsilon\), 
the total number of steps \(T\) satisfies:
\[
  T 
  =\mathcal{O}\Bigl(
    \tfrac{\sigma^2}{m \varepsilon^2}+
    \tfrac{1}{\varepsilon}
    +\colorbox{cyan!10}{$\sum_{t=0}^{T-1}U^{(t)}$}
  \Bigr)
  \cdot
  \bigl(\mathcal{L}(\theta^{(0)}) - \mathcal{L}^{\star}\bigr),
\]
where  \(U^{(t)} = (\eta L_2 - 1)(\nabla\mathcal L(\bar\theta^{(t)}))^\top \nabla\operatorname{Tr}\bigl(\nabla^2\mathcal L(\bar\theta^{(t)})\,\Gamma^{(t)}\bigr) + \Theta(\Xi_{t}^3)\),
with  $\Gamma^{(t)} = \tfrac{1}{m}\sum_{k=1}^m (\theta_k^{(t)} - \bar{\theta}^{(t)})(\theta_k^{(t)} - \bar{\theta}^{(t)})^\top$ and the consensus distance \(\Xi^2_t = \Tr(\Gamma^{(t)})\).
\end{theorem}
\end{tcolorbox}

\begin{remark}
    We note that \cref{thm:nonconvex-convergence-descent-lemma} gives an implicit bound depending on $U^{(t)}, t\in\{1,2,\dots, T-1\}$, rather than a closed-form expression. It primarily serves to bridge convergence with the per-iteration dynamics of $U^{(t)}$, facilitating the subsequent derivation of the conditions on consensus and communication required to recover the parallel SGD rate (see \cref{prop: critical} and \cref{prop: sufficient}).
\end{remark}

\begin{table*}[t!]
\begin{minipage}{\textwidth}
\caption{Comparison of non-convex convergence rates for parallel SGD and DSGD, both run with $m$ agents under non-IID data.}
\label{tab:sgd_comparison}
\centering
\resizebox{\textwidth}{!}{%
\begin{tabular}{lccc}
\toprule[1pt]
Algorithm & Parallel SGD & DSGD \citep{pmlr-v119-koloskova20a} & DSGD (ours) \\
\midrule
Rate & $\mathcal{O}\Bigl(\colorbox{myorange!30}{$
    \tfrac{\sigma^2}{m \varepsilon^2}+
    \tfrac{1}{\varepsilon}$}
  \Bigr)$
& $\mathcal{O}\Bigl(\colorbox{myorange!30}{$
    \tfrac{\sigma^2}{m \varepsilon^2}+\frac{1}{p\varepsilon}$} +
    \colorbox{green!8}{$
     \frac{\sqrt{p}\,\sigma + \zeta}{p\,\varepsilon^{3/2}}$}
  \Bigr)$
& $\mathcal{O}\Bigl(\colorbox{myorange!30}{$
    \tfrac{\sigma^2}{m \varepsilon^2}+
    \tfrac{1}{\varepsilon}$}+
    \colorbox{cyan!10}{$\frac{1}{\varepsilon}\sum_{t=0}^{T-1}U^{(t)}$}
  \Bigr)$ \\
\bottomrule[1pt]
\end{tabular}%
}
\end{minipage}
\end{table*}

\textbf{Comparison and Technical Novelty}. As summarized in \cref{tab:sgd_comparison}, unified analysis by \citet{pmlr-v119-koloskova20a} showed that DSGD suffers from additional terms of order $\mathcal{O}\left(\frac{1-p}{p\varepsilon} + \frac{\sqrt{p}\,\sigma + \zeta}{p\,\varepsilon^{3/2}}\right)$ in the convergence rate compared to parallel SGD. 
The core idea behind their analysis is to separate the effects of three key factors: the descent force (i.e., the squared gradient norm), gradient noise, and parameter discrepancy among agents. Each of these components is then analyzed and controlled separately. Among them, both the gradient noise and the model discrepancy are treated as detrimental to convergence.
In contrast, we adopt a new proof framework that leverages the implicit bias of decentralized learning (see \cref{prop: implicit_bias} \citep{pmlr-v202-zhu23e} and  \acref{sec:implicit_bias}). Rather than treating the discrepancy among agents purely as noise, we partially incorporate it as a constructive component essential for matching the rate of parallel SGD. This intuition is formalized through the convergence guarantee provided in \cref{thm:nonconvex-convergence-descent-lemma}, which introduces an additional term of $\mathcal{O}\left(\frac{1}{\varepsilon}\sum_{t=0}^{T-1}U^{(t)}\right)$. In what follows, we conduct a fine-grained analysis on the sign of $U^{(t)}$.

\begin{remark}[Reduction to Standard Rates]
    We consider two special cases where the term $U^{(t)}$ vanishes because the consensus error is identically zero ($\Xi_t \equiv 0$):
\begin{itemize}[leftmargin=*]
    \item The {single-agent case} ($m = 1$);
    \item The {fully synchronous Parallel SGD case}, where perfect synchronization ensures identical local models ($\theta_k^{(t)} \equiv \bar{\theta}^{(t)}$ for all $k$).
\end{itemize}
    In both settings, the auxiliary term $U^{(t)}$ in \cref{thm:nonconvex-convergence-descent-lemma} strictly equals zero. Consequently, \cref{thm:nonconvex-convergence-descent-lemma} naturally recovers the rate of standard (Parallel) SGD, which is of the order
    \(
        \mathcal{O}\left({\frac{\sigma^2}{m\varepsilon^2} + \frac{1}{\varepsilon}} \right).
    \)
\end{remark}
To better characterize how the high-order loss landscape affects the dynamics of \(U^{(t)}\) , we introduce a new assumption that is theoretically novel yet empirically supported by prior literature.

\begin{assumption}[Progressive sharpening]\label{ass: sharpening}
For any positive semi-definite matrix $\Sigma$, the gradient of population risk negatively aligns with the gradient of sharpness. Formally, $\forall \theta\in\mathbb{R}^{d}$,
\begin{align}
    \nabla \mathcal{L}(\theta)^\top \nabla \operatorname{Tr}(\nabla^2 \mathcal{L}(\theta)\Sigma) < 0. 
\end{align}
\end{assumption}
By the density of $\mathbb{R}$, there exists a $\gamma > 0$ such that $\nabla \mathcal{L}(\theta)^\top \nabla \operatorname{Tr}(\nabla^2 \mathcal{L}(\theta)\Sigma) < -\gamma \|\nabla \mathcal{L}(\theta)\|< 0$ holds strictly. We refer to $\gamma^*\triangleq \sup\left\{ \gamma > 0 \mid \nabla \mathcal{L}(\theta)^\top \nabla \text{Tr}(\nabla^2 \mathcal{L}(\theta)\Sigma) < -\gamma \|\nabla \mathcal{L}(\theta)\| \right\}$ as the \textit{degree of progressive sharpening}.
\begin{remark}
    $\operatorname{Tr}(\nabla^2 \mathcal{L}(\theta)\Sigma)$ can be interpreted as an ``average sharpness'' around $\theta$; see similar metrics in \citep{gu2023why, pmlr-v202-zhu23e}.
    \cref{ass: sharpening} reflects a widely observed phenomenon in deep learning:  
    The loss gradient exhibits a negative correlation with the gradient of sharpness \citep{wang2022analyzing,  damian2023selfstabilization, Cohen2025understanding}.
    Intuitively, this condition implies that as the optimizer moves to reduce the loss, it simultaneously moves in a direction that increases the sharpness.
\end{remark}
\cref{ass: sharpening} ensures $\nabla\mathcal L(\bar\theta^{(t)})^\top \nabla\operatorname{Tr}\bigl(\nabla^2\mathcal L(\bar\theta^{(t)})\,\Gamma^{(t)}\bigr)$ in $U^{(t)}$ remains negative.
In the following, we establish that this term can dominate the other terms in $U^{(t)}$, thereby ensuring $U^{(t)}<0$.

\begin{tcolorbox}[notitle, rounded corners, colframe=middlegrey, colback=lightblue, 
       boxrule=2pt, boxsep=0pt, left=0.15cm, right=0.17cm, enhanced, 
       toprule=2pt]
\begin{proposition}\label{prop: critical}
Suppose \cref{ass: regularity} and \cref{ass: sharpening} hold. Assume  $\eta$ satisfies $\eta > 1/L_2$, and assume \(\|\nabla\mathcal L(\bar\theta^{(t)})\|\ge \mu_t>0\) for all \(t\). 
Consider the  matrix \(\Gamma^{(t)} = \tfrac{1}{m}\sum_{k=1}^m (\theta_k^{(t)} - \bar{\theta}^{(t)})(\theta_k^{(t)} - \bar{\theta}^{(t)})^\top\) and its trace \(\Xi_t^2=\operatorname{Tr}(\Gamma^{(t)})\).
Then, for any fixed \(m>0\), there exists \(\Xi_t^2 > 0\) such that 
\begin{inequality}\label{eq:critical}
\colorbox{cyan!10}{$U^{(t)}$}\triangleq \frac{1}{2}(\eta L_2 - 1)\nabla\mathcal L(\bar\theta^{(t)})^\top \nabla\operatorname{Tr}\bigl(\nabla^2\mathcal L(\bar\theta^{(t)})\,\Gamma^{(t)}\bigr) + O(\Xi_{t}^3)< 0.
\end{inequality}
\end{proposition}
\end{tcolorbox}

\textbf{Explanations for Assumptions.} 
We assume a lower bound on the global gradient norm evaluated at the averaged parameters $\bar\theta^{(t)}$, i.e., $\|\nabla\mathcal L(\bar\theta^{(t)})\|\ge \mu_t>0$. We note that this applies to the gradient on the global data set, which can remain significant even if individual local gradients vanish. The assumption is motivated by the Polyak-Lojasiewicz (PL) condition \citep{polyak1963gradient}, $\frac{1}{2}\|\nabla\mathcal{L}(\theta)\|^2 \ge \mu(\mathcal{L}(\theta) - \mathcal{L}^*)$, which ensures the gradient is bounded from zero before reaching the optimum. Our new assumption formalizes this property for the pre-convergence phase by denoting this lower bound at iteration $t$ as $\frac{1}{2}\mu_t^2$. We also note that \cref{ass: regularity} requires that the norm of the loss derivatives is bounded up to the fourth order, $\|\nabla^q \mathcal{L}(\theta)\| \le L_q$ for $q=1,2,3,4$. These bounds are necessary to analyze the interaction between the consensus error $\Xi_t$ and higher-order landscape.

At a high level, \cref{prop: critical} highlights the potential of leveraging decentralized training to accelerate distributed training beyond communication efficiency. This theoretical insight aligns with the empirical gains observed in \cref{fig: main_clip_adamw_16}.
Specifically, \cref{prop: critical} implies that satisfying $U^{(t)}<0$ necessitates careful control of both $\eta$ and $\Xi_t^2$. We proceed by discussing the learning rate constraint below, and subsequently analyze the role of $\Xi_t^2$, based on which we provide a theoretical justification for allocating more communication in the later stages of training in \cref{sec: com_allocation}.
\begin{tcolorbox}[notitle, rounded corners, colframe=middlegrey, colback=lightred, 
       boxrule=2pt, boxsep=0pt, after skip=7pt, left=0.15cm, right=0.17cm, enhanced, 
       toprule=2pt,
    ]
\begin{remark}[Acceleration Under Larger Learning Rate]\label{remark: acceleartion_unstable}
    Note that the coefficient of the term $\nabla\mathcal L(\bar\theta^{(t)})^\top \nabla\operatorname{Tr}\bigl(\nabla^2\mathcal L(\bar\theta^{(t)})\,\Gamma^{(t)}\bigr)$ in \cref{eq:critical} is $\frac{1}{2}(\eta L_2 - 1)$. To ensure the negativity of this term, we require the condition $\eta > \frac{1}{L_2}$. Notably, the resulting interval $\frac{1}{L_2} < \eta < \frac{2}{L_2}$ coincides exactly with the regime of ``oscillatory convergence" in classical optimization theory for a quadratic objective $\mathcal L (\theta)= \frac{1}{2} \theta^\top H \theta$ \citep[Chapter~1, p.~26]{polyak1987introduction}.
\end{remark}
\end{tcolorbox}
To provide deeper intuition, we outline the proof sketch of \cref{thm:nonconvex-convergence-descent-lemma} below. This analysis establishes a novel descent lemma tailored for decentralized SGD, demonstrating how learning rate control harnesses progressive sharpening to drive acceleration.
\begin{align}\label{eq:descent_lemma_dec-2}
    \mathbb{E}_{\xi^{(t)}} \left[ \mathcal{L}(\bar{\theta}^{(t+1)}) \right] \leq \, & \mathcal{L}(\bar{\theta}^{(t)}) 
     -  \underset{\textcolor{textcyan}{>0}}{\colorbox{boxcyan!70}{$\displaystyle \textcolor{textcyan}{\bigl(\eta - \frac{\eta^2 L_2}{2}\bigl)}$}} 
      \underbrace{ \left\| \nabla \mathcal{L}(\bar{\theta}^{(t)}) \right\|^2 }_{\text{Standard Descent}} \nonumber\\
    & + \underset{\textcolor{textpink}{>0}}{\colorbox{boxpink!40}{$\displaystyle \textcolor{textpink}{\bigl(\eta^2 L_2 - \eta\bigl)}$}} 
      \underbrace{ \nabla \mathcal{L}(\bar{\theta}^{(t)})^\top \nabla \text{Tr}\bigl( \nabla^2 \mathcal{L}(\bar{\theta}^{(t)}) \Gamma^{(t)} \bigl)}_{< 0, \text{ \textcolor{textpink}{progressive sharpening}}} 
     + \frac{\eta^2 L_2 \sigma^2}{2m} + \mathcal{O}(\Xi_t^3).
\end{align}

We note that this third-order effect differs from prior analyses, which were typically established in a near-minima regime \citep{li2022what}. By contrast, we show that this mechanism emerges whenever local models are inconsistent, i.e., $\Xi_t > 0$.

\cref{eq:descent_lemma_dec-2} further illuminates the critical role of the consensus violation term $\Xi_t = \sqrt{\operatorname{Tr}(\Gamma^{(t)})}$. Crucially, the progressive sharpening term scales with $\mathcal{O}(\Xi_t^2)$, whereas the higher-order residual error is $\mathcal{O}(\Xi_t^3)$. Consequently, provided $\Xi_t$ is properly controlled such that the $\mathcal{O}(\Xi_t^2)$ gain dominates the $\mathcal{O}(\Xi_t^3)$ error, decentralized SGD can match or even surpass the convergence rate of parallel SGD.
According to \cref{coro: consensus_distance}, \(\mathbb{E}\bigl[\Xi_t^2\bigr]\) is bounded by
\begin{align}\label{eq:consensus_distance}
  \mathbb{E}\bigl[\Xi_t^2\bigr]
  \le
  \mathcal{O}{\left(\,\frac{(1-p)}{p^2}\right)},
\end{align}
where the parameter \(p (\text{with } p \in (0,1]\) reflects connectivity in the communication graph (see \cref{ass: mixing}). A larger \(p\) indicates better connectivity and faster consensus, while a smaller \(p\) implies a sparse communication graph (i.e., lower communication) and slower information propagation.
For example, \(p = 1\) corresponds to a fully connected topology, enabling perfect communication, whereas \(p = 0\) represents the extreme case of complete local training with no communication. 

\textbf{Why Limited but Nonzero Communication Enables  Mergeability.} Notably, random communication graphs can achieve \(p = \Theta(1)\), striking a favorable trade-off: they require relatively low communication overhead while still maintaining efficient information mixing due to randomized edge sampling, which ensures a rapid decrease of $\Xi_t$ \citep{vos2023epidemic}. This is why we adopt random topologies as the primary setup in our experiments: They can satisfy the condition in \cref{prop: critical} even under extremely limited communication, thereby ensuring mergeability (see \cref{fig: main_combined}).
However, in the case of full local training where \(p = 0\) (see \cref{fig: topology}), the right-hand side of \cref{eq:consensus_distance} increases to infinity, indicating that \(\Xi_t\) may diverge. As a consequence, the condition of \(\Xi_t\) in \cref{prop: critical} can no longer be satisfied, which explains why local models after complete local training cannot be reliably merged (see the green curve in \cref{fig: main_resnet}).



\subsection{A Theoretical Explanation for Communication Allocation}\label{sec: com_allocation}

\cref{prop: critical} highlights the importance of  $\Xi_t^2$ to satisfy \cref{eq:critical}. This motivates the question of how small $\Xi_t^2$ (or how large $p$) should be, which we answer in the following sufficient condition.
\begin{tcolorbox}[notitle, rounded corners, colframe=middlegrey, colback=lightblue, 
       boxrule=2pt, boxsep=0pt, left=0.15cm, right=0.17cm, enhanced, 
       toprule=2pt]
\begin{proposition}[Critical Consensus Edge]\label{prop: sufficient}
Suppose \cref{ass: mixing}-\cref{ass: sharpening} hold. 
Assume $\eta>\frac{1}{L_2}$, and the consensus error  \(\Xi_t \le 1\) for all $t$.
Then, the following condition ensures that the critical \cref{eq:critical} is satisfied:
\begin{align}\label{ineq:sufficient}
  \sqrt{\frac{24\colorbox{cyan!10}{$\textcolor{textcyan}{\bm{\bigl(1-p)}}$}\eta^2}{\colorbox{cyan!10}{$\textcolor{textcyan}{\bm{p^2}}$}}
  \bigl(\phi^2 + \sigma^2\bigr)} 
  < 
  \min \left\{ 
     \frac{(\eta L_2 - 1)\gamma^* \colorbox{boxpink!50}{$\textcolor{textpink}{\bm{\mu_t}}$}}{2(\eta L_2 + \frac{L_4}{24})\sqrt{m}L_1}, \quad 
     \sqrt{\frac{(\eta L_2 - 1)\gamma^* \colorbox{boxpink!50}{$\textcolor{textpink}{\bm{\mu_t}}$}}{2 \Sigma_{\mathrm{high}}}} 
  \right\},
\end{align}
where \(\Sigma_{\mathrm{high}} = \frac{1}{8}\eta L_2 L_3^2 + \frac{1}{2}\eta \sqrt{m} L_2 L_3 + \frac{\eta m L_2 L_4^2}{1152}\). 
Here, \(\gamma^*\) denotes the degree of progressive sharpening (see \cref{ass: sharpening}),  $\phi^2$ denotes the uniform upper bound of the averaged squared local gradient norm (i.e., \(\frac{1}{m}\sum_{k=1}^m\|\nabla\mathcal{L}_k(\theta_k^{(t)})\|^2\leq \phi^2\)), and \(\mu_t\) is the lower bound on the global gradient norm (i.e., \(\|\nabla\mathcal L(\bar\theta^{(t)})\|\ge \mu_t>0\)).
\end{proposition}
\end{tcolorbox}

\textbf{Practical Guidance}. 
\cref{prop: sufficient} provides guidance for allocating communication to ensure $U^{(t)}\leq 0$, thereby guaranteeing the contribution of each step to the cumulative sum in \cref{thm:nonconvex-convergence-descent-lemma}. From \cref{ineq:sufficient}, note that $\phi$, $\sigma^2$, $\gamma^*$, $m$, and~$L_q$ ($q \in \{1, \ldots, 4\}$) are time-independent, while the key time-varying factor is the gradient lower bound $\mu_t$, which tracks the optimization status of the averaged parameters $\bar{\theta}^{(t)}$ on the global landscape.
Under this interpretation, \cref{ineq:sufficient} implies that the communication-related term $p$ should be dynamically adjusted in response to the changing landscape geometry captured by~$\mu_t$. Analytically, the left-hand side of \cref{ineq:sufficient} is a strictly decreasing function of $p$, while the right-hand side is an increasing function of $\mu_t$. This implies a fundamental trade-off: more frequent communication (larger $p$) relaxes the condition, whereas a vanishing gradient (smaller $\mu_t$) tightens the allowable error bound.
Specifically,

\begin{itemize}[leftmargin=*]
    \item Early, High-Gradient Regime: In the starting phase of training, when the globally averaged model is far from a minimum, the lower bound on gradient norm \(\mu_t\) is large. This corresponds to a relaxed consensus requirement in \cref{ineq:sufficient}, which permits low-frequency communication (i.e., smaller \(p\)) without significantly impacting the performance of the globally merged model.
    \item {Late, Low-Gradient Regime}: As models approach a solution and training enters a convergence phase, the gradient norm $\mu_t$ decreases. This tightens the constraint in \cref{ineq:sufficient}. In this regime, frequent communication (i.e., larger $p$) becomes critical.
\end{itemize}

We note that this theoretically motivated guidance aligns well with our empirical findings in \cref{sec: main_discovery} that more communication should be concentrated in the later stages of training.


%% file: Section/8-Implications.tex
\section{Implications and Discussions}\label{sec: implications}
\textbf{Model Merging.} The effectiveness of a single merging of decentralized models has significant implications for model merging. A recent work showed that pre-trained models occupy a large, flat ``basic capability basin'', within which fine-tuning creates smaller ``specific capability basins'' \citep{chen2025understanding}. The observed ``mergeability'' of local models in our paper implies that decentralized learning may guide agents into connected specific capability basins, allowing simple permutation-free merging to integrate specialized knowledge. This suggests a practical direction: lightweight synchronization during local training may improve basin connectivity and simplify later merging into a more capable model.

\textbf{Decentralized Learning.} Our work provides promising empirical and theoretical evidence that  decentralized learning can generalize under high data heterogeneity and limited communication. More importantly, our findings could directly motivate a new class of adaptive, communication-efficient decentralized learning algorithms, which dynamically allocate their communication budget by monitoring training dynamics to satisfy the critical consensus edge condition in \cref{ineq:sufficient}.

%% file: Section/Appendix.tex
\onecolumn
\appendix

\numberwithin{equation}{section}
\numberwithin{theorem}{section}
\numberwithin{lemma}{section}
\numberwithin{proposition}{section}
\numberwithin{corollary}{section}
\numberwithin{remark}{section}
\numberwithin{definition}{section}
\numberwithin{assumption}{section}
\numberwithin{figure}{section}
\numberwithin{table}{section}
\renewcommand{\thesection}{{\Alph{section}}}
\renewcommand{\thesubsection}{\Alph{section}.\arabic{subsection}}
\renewcommand{\thesubsubsection}{\Alph{section}.\arabic{subsection}.\arabic{subsubsection}}

\section*{LLM Usage Statement}

We use large language models (LLMs) as writing-assistance tools. Their role is confined to proofreading and language polishing.

\section*{Impact Statement}

This paper studies the problem of temporal communication allocation in decentralized distributed learning, a topic of very high significance in the era of communication-intensive large model training.
Specifically, we aim to contribute to the development of communication-efficient decentralized learning without compromising performance. The potential positive social impact are twofold:
\begin{itemize}[leftmargin=*]
    \item \textbf{Democratizing Access.} For individuals and organizations with constrained infrastructure, our work contributes to the democratization of access to large-scale collaborative training. By reducing communication requirements, we lower the barrier to entry for participating in advanced model development. Such inclusivity can extend the applicability of distributed learning systems to edge environments, thereby promoting more equitable contributions to models trained at scale.
    \item \textbf{Reducing Training Costs.} In data center environments, our approach can alleviate communication bottlenecks of distributed training. This reduction directly translates to shorter total wall-clock training time, thereby lowering the overall costs and energy consumption associated with large-scale distributed training.
\end{itemize}
No negative societal impacts are identified.

\section*{Ethics statement}

Our research strictly adheres to the ICLR Code of Ethics. The work is foundational, focusing on the algorithmic and theoretical properties of decentralized learning, and does not involve human subjects or the collection of new sensitive data. All experiments were conducted on publicly available, standard academic datasets. We foresee no direct negative societal impacts; on the contrary, by reducing communication overhead, our findings may contribute positively by democratizing access to large-scale distributed training and lowering the associated resource footprint.

\section*{Reproducibility statement}

We are committed to the reproducibility of our research. Our theoretical claims, including all assumptions and their justifications, are presented in \cref{sec: theory} with complete, step-by-step proofs provided in \acref{sec: proof}. Comprehensive details for reproducing our empirical results, including model architectures, data processing, hyperparameter settings, and communication configurations, are well documented in \acref{sec: setup}.

\section{Limitations and Potential Questions}\label{sec: QA}
\begin{tcolorbox}[notitle, rounded corners, colframe=middlegrey, colback=verylightpurple, 
       boxrule=2pt, boxsep=0pt, after skip=7pt, left=0.15cm, right=0.17cm, enhanced, 
       toprule=2pt,
    ]
\textit{\textbf{Q}: Why use decentralized \textbf{AdamW} in some experiments when the theory is on decentralized \textbf{SGD}?}
\end{tcolorbox}
\textbf{A}:We use decentralized AdamW in some of our experiments for its superior performance in Non-IID settings. Crucially, all reported empirical observations remain fully consistent when using decentralized SGD, directly aligning with our theoretical analysis (see \cref{fig: main_combined} and \cref{sec: add_exp}).

\begin{tcolorbox}[notitle, rounded corners, colframe=middlegrey, colback=verylightpurple, 
       boxrule=2pt, boxsep=0pt, after skip=7pt, left=0.15cm, right=0.17cm, enhanced, 
       toprule=2pt,
    ]
\textit{\textbf{Q}: How does the theory explain local models in decentralized learning are globally mergeable?}
\end{tcolorbox}

\textbf{A}: The theoretical explanation for the ``mergeability'' of local models in decentralized learning is supported by our result that a globally merged model converges faster to the optimum than individual local models. Specifically, we provide a fine-grained convergence analysis showing that the model merged from decentralized SGD (DSGD) can match the convergence rate of parallel SGD, despite limited communication.
Since the rate of $m$-agent parallel SGD is superior to that of a single local model, this result transitively justifies the merged model's superior performance relative to any individual model, thereby providing theoretical support for mergeability.

\begin{tcolorbox}[notitle, rounded corners, colframe=middlegrey, colback=verylightpurple, 
       boxrule=2pt, boxsep=0pt, after skip=7pt, left=0.15cm, right=0.17cm, enhanced, 
       toprule=2pt,
    ]
\textit{\textbf{Q (Hyperparameter Tuning)}: How were the baselines tuned in terms of hyperparameters?}
\end{tcolorbox}

\textbf{A:} All hyperparameters were tuned via grid search based on global test performance, with the batch size searched over $\{64, 128\}$. For ResNet-18 trained from scratch on Tiny ImageNet, we searched the learning rate over $\{1 \times 10^{-4}, 5 \times 10^{-4}, 1 \times 10^{-3}\}$ for AdamW and $\{1 \times 10^{-3}, 5 \times 10^{-3}, 1 \times 10^{-2}\}$ for SGD. For CLIP ViT-B/32 on Tiny ImageNet, we searched the learning rate over $\{1 \times 10^{-4}, 5 \times 10^{-4}, 1 \times 10^{-3}\}$ for AdamW and $\{5 \times 10^{-4}, 1 \times 10^{-3}, 5 \times 10^{-3}, 1 \times 10^{-2}\}$ for SGD. For the optimal hyperparameters selected for our main experiments, please refer to the \textbf{Implementation Details} in \acref{sec: setup} and the additional empirical results in \cref{sec: add_exp}).

\begin{tcolorbox}[notitle, rounded corners, colframe=middlegrey, colback=verylightpurple, 
       boxrule=2pt, boxsep=0pt, after skip=7pt, left=0.15cm, right=0.17cm, enhanced, 
       toprule=2pt,
    ]
\textit{\textbf{Q (Comparison with Model Soup)}: How do initialization schemes affect results? Performance gains from merging have been observed in Model Soup \citep{wortsman2022model}.}
\end{tcolorbox}

\textbf{A: } We use different initialization schemes and observe consistent performance gains from global merging, whether models start from different random initializations or from a pretrained state. The majority of our experiments use different initializations, demonstrating that local models in decentralized learning can be effectively merged regardless of their starting points. This is quite surprising, as it contrasts with methods like \textbf{Model Soup}, which require models to be fine-tuned from an identical pretrained state.
Furthermore, our experiments with a shared pretrained state confirm that the performance gains hold in that setting as well (see \cref{fig: main_clip} and \cref{sec: add_exp}). 

\begin{tcolorbox}[notitle, rounded corners, colframe=middlegrey, colback=verylightpurple, 
    boxrule=2pt, boxsep=0pt, after skip=7pt, left=0.15cm, right=0.17cm, enhanced, 
    toprule=2pt,
]
\textit{\textbf{Q (Methodology for Landscape Visualization)}: Please clarify the methodology for visualizing the loss landscape in \cref{fig: resnet_landscape}, including the basis for the visualization grid.}
\end{tcolorbox}
\textbf{A}: We adopt the visualization tool from \citep{crisostomi2024cm}, positioning 16 trained models at the vertices of a regular hexadecagon. Any point within this polygon is an interpolated model whose parameters are determined by Wachspress barycentric coordinates; we then evaluate its cross-entropy loss to generate the contour map. Unlike methods that use random directions, our visualization grid is \textbf{deterministically} defined by the models themselves, allowing a direct investigation of their geometric connectivity. The full implementation is available in their official code repository \href{https://github.com/crisostomi/cycle-consistent-model-merging/blob/master/notebooks/plots/plot_loss_contours_n_models.ipynb}{https://github.com/crisostomi/cycle-consistent-model-merging/blob/master/notebooks/plots/plot\_loss\_contours\_n\_models.ipynb}.

\begin{tcolorbox}[notitle, rounded corners, colframe=middlegrey, colback=verylightpurple, 
    boxrule=2pt, boxsep=0pt, after skip=7pt, left=0.15cm, right=0.17cm, enhanced, 
    toprule=2pt,
]
\textit{\textbf{Q (Experimental Scope)}: The empirical findings are restricted to visual tasks.}
\end{tcolorbox}
\textbf{A}: Our empirical findings primarily focus on tasks within the vision domain. We note that this is consistent with most existing decentralized learning literature \citep{pmlr-v139-lin21c, pmlr-v139-kong21a, ying2021exponential, vogels2021relaysum, Li_2022_CVPR, zehtabi2025decentralized}. Extending the experimental setup to broader tasks is a meaningful direction for future research.

\begin{tcolorbox}[notitle, rounded corners, colframe=middlegrey, colback=verylightpurple, 
    boxrule=2pt, boxsep=0pt, after skip=7pt, left=0.15cm, right=0.17cm, enhanced, 
    toprule=2pt,
    ]
\textit{\textbf{Q}: The finding in \cref{fig: sliding_window_cifar100} (c), that local models eventually converge to a similar state even with limited communication, was also observed in prior work \citep{10.1145/1082469.1082470}.}
\end{tcolorbox}
\textbf{A}:  In our setting, the local models do not, in fact, converge to a similar state or a single consensus point. This is because our work addresses a more challenging heterogeneous data regime, which differs from the setting in the cited prior work.
Instead, we identify an emergent geometric structure where decentralized training guides local models to a shared ``high-loss ring'' surrounding a central low-loss basin (see \cref{fig: resnet_landscape}). Although the models do not reach a consensus, they remain surprisingly mergeable within this region. This geometric arrangement allows their average, i.e., the globally merged model, to fall directly into the low-loss basin. To the best of our knowledge, we are the first to identify this emergent phenomenon in decentralized learning.

\section{Additional Background and Related Work}\label{sec: background}

\subsection{Decentralized Learning}\label{sec: dec}
Modern large-scale model training and inference are predominantly conducted within centralized, high-cost data centers.
Driven by mounting constraints on computational resources and power availability \citep{pilz2025trends}, both academia and industry are increasingly exploring decentralized training approaches \citep{openai_stargate, grandview2024ai}.
This paradigm, drawing inspiration from swarm intelligence systems \citep{bonabeau1999swarm, MAVROVOUNIOTIS20171}, offers a more economical and scalable approach by distributing computational tasks across globally distributed nodes, rather than relying solely on a single central server \citep{yuan2022decentralized, NEURIPS2023_28bf1419,jaghouar2024intellect, ramasinghe2025protocol}.
A notable illustration of the computational potential through decentralization is the Bitcoin system, which sustains workloads equivalent to a 16 GW power draw \citep{CBECI}, surpassing by a factor of three the estimated 5 GW consumption of the largest AI supercluster under development \citep{microsoft_openai_stargate, openai_stargate}.

To provide context, we summarize key algorithmic and theoretical advances in decentralized learning. While our discussion highlights several notable contributions, it is not exhaustive; readers are referred to recent advances and surveys \citep{zhu2025dice, 10251949, singhaperspective, 10542323, he2025democratic, ramasinghe2025protocol, kolehmainen2025noloco}.

\textbf{Algorithmic Progress in Decentralized Learning}.
The advancement of decentralized learning algorithms has been primarily driven by the need for communication-efficiency in practical distributed learning. Decentralized algorithms have been refined to handle a variety of realistic scenarios, including time-varying communication topologies \citep{nedic2014distributed,pmlr-v119-koloskova20a,ying2021exponential,takezawa2023beyond}, asynchronous updates \citep{lian2018asynchronous,xu2021dp,nadiradze2021asynchronous,bornstein2023swift,pmlr-v238-even24a}, statistical heterogeneity \citep{tang2018d,vogels2021relaysum,pmlr-v206-le-bars23a}, and robustness to Byzantine failures \citep{he2022byzantine, ye2025generalization}. Moreover, recent works extended beyond standard empirical risk minimization to more structured problem classes, such as compositional \citep{gao2021fast}, minimax \citep{xian2021faster,zhu2023stability,chen2024}, and bi-level optimization \citep{yang2022decentralized,pmlr-v206-gao23a,pmlr-v202-chen23n}. Additionally, privacy concerns in decentralized learning are also critical, with efforts focusing on differentially privacy \citep{pmlr-v235-cyffers24a, pmlr-v235-allouah24b} and data reconstruction attacks \citep{pmlr-v235-mrini24a}.

\textbf{Theoretical Progress in Decentralized Learning.}
Foundational work on decentralized optimization \citep{nedic2009distributed,sayed2014adaptation,yuan2016decentralized,NIPS2017_f7552665} laid the groundwork for understanding convergence. Building on this, \citet{pmlr-v139-lu21a} proposed a hierarchical abstraction of decentralization, distinguishing it into three layers, providing a unified view across federated and decentralized paradigms. \citet{pmlr-v119-koloskova20a} consolidated synchronous decentralized SGD algorithms with changing communication topologies and local updates, and \citet{pmlr-v238-even24a} extended the unifying perspective to asynchronous protocols. More recently, \citet{zehtabi2025decentralized} developed these frameworks further by considering the sporadicity of both communication and computations.
On the generalization front, \citet{richards2020graph} derived stability-based bounds for decentralized SGD in convex settings, while \citet{Sun_Li_Wang_2021} extended these to non-convex objectives, revealing a dependency on the spectral gap of the communication graph. This dependency was subsequently refined by \citet{zhu2022topology}, who introduced a Gaussian weight difference assumption to tighten the bound. Complementary results showed that in convex regimes, the generalization of decentralized SGD matches that of centralized SGD \citep{pmlr-v235-le-bars24a}, while in non-convex landscapes, decentralization primarily impacts worst-case generalization behavior.
To account for unexplained generalization behaviors in decentralized training \citep{pmlr-v139-kong21a, gurbuzbalaban2022heavy, JMLR:v24:22-1471}, \citet{pmlr-v202-zhu23e} linked decentralized SGD to random sharpness‐aware minimization (SAM), revealing a bias toward flatter minima.
Notably, akin to our finding that decentralized learning generalizes when allocated high communication late in training, \citet{zhou2025sharpnessaware} has shown that SAM efficiently selects flatter minima in the later stage of training.

\textbf{Towards Decentralized Training of Foundation Models.}
Recent advances have shown the feasibility of training large-scale foundation models in decentralized environments. DT-FM \citep{yuan2022decentralized} introduced tasklet-based scheduling for Transformer training under bandwidth-constrained settings, enabling efficient resource allocation. SWARM Parallelism \citep{pmlr-v202-ryabinin23a} scaled decentralized training through resilient pipeline design and adaptive load balancing. CocktailSGD \citep{pmlr-v202-wang23t} further improved efficiency via a combination of decentralization, gradient sparsification, and quantization for LLM fine-tuning. On the inference side, Petal \citep{borzunov-etal-2023-petals} exploited peer-to-peer networks to amortize computational costs across heterogeneous nodes. Most recently, Intellect \citep{jaghouar2024intellect}, building on Diloco \citep{douillard2023diloco}, leveraged hybrid parallelism, i.e., both data and model parallelism, to collaboratively train models with billions of parameters.
NoLoCo \citep{kolehmainen2025noloco} further extended Diloco to gossip-type decentralized settings.
For a broad survey of large-scale deep learning practice, see \citet{10.1145/3700439, shen2025will}.

\subsection{Implicit Bias of Decentralized Learning}\label{sec:implicit_bias}
The concept of implicit bias, i.e., the intrinsic preference of learning algorithms for solutions with certain properties, has emerged as a key concept in explaining the empirical success of modern deep learning \citep{li2022what, 10.1145/3571070, lyu2024implicit}.
Recent studies have highlighted intriguing distinctions between decentralized stochastic gradient descent (DSGD) and its centralized counterpart (CSGD). \citet{gurbuzbalaban2022heavy} demonstrated that under certain conditions, DSGD operating on large, sparse topologies exhibits heavier-tailed parameter distributions compared to CSGD. \citet{zhang2021loss} showed that decentralization introduces landscape-dependent noise, which can improve tolerance to larger learning rates. This observation aligns with findings by \citet{JMLR:v24:22-1471}, who revealed that collaboration in decentralized settings permits the use of larger learning rates. \citet{pmlr-v202-zhu23e} first explicitly characterized the implicit bias of decentralized SGD by establishing its connection with random sharpness-aware minimization, proving the existence of flatness bias in decentralized training. Complementing this, \citet{cao2024trade} offered a detailed analysis of the interplay between flatness and optimization in DSGD, particularly its ability to escape local minima. More recently, \citet{wu2024implicit} investigated the implicit regularization properties of decentralized optimization in non-convex sparse regression problems, recovering the convergence rates achieved by gradient descent in centralized settings.

\textbf{Comparison with \citet{pmlr-v202-zhu23e}}.
We note that  \citet{pmlr-v202-zhu23e} has highlighted the generalization benefits of decentralized learning, but key differences exist in terms of the experimental setup and the insights derived. While \citet{pmlr-v202-zhu23e} focused on IID scenarios and specific cases involving exceptionally large batch sizes, we consider the more realistic non-IID setting using standard batch sizes. This shift in focus allows us to uncover phenomena not observed by \citet{pmlr-v202-zhu23e}, including insights into communication allocation strategies.

\subsection{Model Merging}\label{sec:model_merging}

\textbf{Mode Connectivity and Model Merging Techniques.}
Recent works on \emph{(Linear) Mode Connectivity} have advanced our understanding of the complex loss landscape in neural networks.
\citet{freeman2017topology,draxler2018essentially,garipov2018loss,nagarajan2019uniform,frankle2020linear} discovered that 
different solutions of deep neural networks can be merged together by simply averaging their parameters. \citet{sonthalia2025do} further showed that the solutions may form a star domain.
We note that these phenomena are  observed in the following scenarios:
\begin{itemize}[leftmargin=*]
    \item \emph{Shared initialization~\citep{frankle2020linear,fort2020deep,zhou2023going}.} Models are initialized from a pretrained checkpoint.  
    \item   \emph{Homogeneous data distribution \citep{wortsman2022model}}. Models are trained on homogeneous data distribution.  
    \item \emph{Permutation~\citep{ainsworth2023git,entezari2022the}.} Models are independently trained. The neurons of one model are permuted to match the neurons of the other while maintaining a functionally equivalent network.
\end{itemize}
These findings have inspired a range of model merging techniques for various applications.
\citet{DBLP:conf/uai/IzmailovPGVW18,matena2022merging,rame2022diverse,pmlr-v202-rame23a,wortsman2022model,Wortsman_2022_CVPR} found that merging the parameters of models that start from the same pretrained model and finetune over the same task leads to improved generalization and robustness.
Furthermore, \citet{ilharco2022patching,li2022branchtrainmerge,ilharco2023editing,ortiz-jimenez2023task,yadav2023tiesmerging} showed that merging models that finetune over different tasks enables multi-task abilities.

\textbf{Comparisons with Model Merging Literature.} Our results show that mode connectivity, or mergeability, can still emerge in decentralized learning, even when the local models are initialized \emph{differently}, trained on highly \emph{heterogeneous} data, and merged \emph{without} any permutation.
Our findings offer new insights into both model merging techniques and the geometry of the neural network loss landscape, which we anticipate will motivate further advances in both areas.

\section{Additional Experiments}
\subsection{Experimental Setups}\label{sec: setup}
\textbf{Computational Resources.} The experiments were conducted on a computing facility equipped with 80 GB NVIDIA\textsuperscript{\textregistered} A100\textsuperscript{\texttrademark} GPUs. All implementations are based on PyTorch, and computations are distributed across multiple GPUs for efficiency. 

\textbf{Dataset.}  
We use three widely adopted image classification datasets: CIFAR-10, CIFAR-100 \citep{krizhevsky2009learning}, and Tiny ImageNet \citep{le2015tiny}. CIFAR-10 consists of 60,000 RGB images across 10 classes, while CIFAR-100 contains 60,000 RGB images across 100 classes. The images in both datasets have a spatial resolution of $32 \times 32$ pixels. Tiny ImageNet is a subset of the ImageNet dataset, comprising 100,000 images drawn from 200 classes, with each image resized to $64 \times 64$ pixels. It provides a mid-scale benchmark that is more challenging than CIFAR datasets but less computationally demanding than training full ImageNet. To incorporate data augmentation, we employ a combination of RandomCrop with 4-pixel padding, RandomHorizontalFlip, and RandAugment with \texttt{num\_ops}=2 and \texttt{magnitude}=9.

\textbf{Details of Decentralized Learning}. 
We simulate a heterogeneous decentralized learning environment.
For our main experiments (\cref{fig: main_clip} and \cref{fig: main_resnet}), we use $m=32$ agents, while for other experiments, including the sliding window experiments (\cref{fig: sliding_window_cifar100}) and the loss landscape visualizations (\cref{fig: resnet_landscape}), we use $m=16$ agents. The number of agents for the visualization was chosen as $16$ for clarity, as a plot with $32$ models would be visually crowded.
In all configurations, we employ a Dirichlet distribution characterized by $\alpha=0.1$ to partition the data among agents.
The Dirichlet distribution is commonly used to partition data in federated learning scenarios, as it allows for the control of label distribution skew among agents \citep{pmlr-v97-yurochkin19a, hsu2019measuring}. A smaller $\alpha$ results in more imbalanced data distributions, where some agents predominantly receive data from a limited number of classes, while a larger $\alpha$ results in more uniform label distributions across agents. This configuration effectively captures the realistic non-IID nature of decentralized learning, where different agents may have access to personalized data reflective of their local environments. 


\begin{itemize}[leftmargin=*]
    \item \textbf{Communication Graph.}  
    We evaluate three decentralized communication topologies: random graph, ring graph, and exponential graph. In the random graph setting, during each communication round, each agent selects a random subset of its neighbors for gossip averaging. For ``R 1'', each agent selects exactly one random neighbor in each round. For ``R 0.2'', each agent selects one neighbor with probability 0.2 and continues local training without communication with probability 0.8. The ring graph enforces a fixed cyclic communication structure, while the exponential graph ensures connectivity by allowing agents to communicate at exponentially increasing distances.

    \item \textbf{Communication Rounds and Local Steps.}  
    The decentralized learning process is conducted over $T = 300$ communication rounds. We use a local training step size of $H=100$ batches per communication round to balance communication and computation costs.
    \item \textbf{Local Data per Agent.}  
    Each agent is assigned a subset of the dataset with a fixed size of $4096$ samples, drawn according to a Dirichlet distribution to simulate realistic non-IID scenarios. 
\end{itemize}


\textbf{Model Architecture.}  
To ensure a representative comparison across different model families, we adopt ResNet-18 \citep{he2016identity} and CLIP ViT-B/32 \citep{pmlr-v139-radford21a} as backbone architectures in our experiments. ResNet-18 is a widely used lightweight convolutional neural network that serves as a canonical example of traditional CNN-based architectures. In contrast, CLIP ViT-B/32 is a transformer-based vision model pre-trained on large-scale image–text pairs. For experiments on Tiny ImageNet, where images are resized to 64×64 pixels, we adjust the CLIP visual encoder to handle the lower resolution. With a patch size of 32, each image yields 4 visual tokens arranged in a 2×2 grid, plus a [CLS] token, resulting in a 5-token input sequence.

\textbf{Implementation Details.} All hyperparameters are tuned through grid search based on global test performance (see \cref{def: avg_gen}). For experiments using decentralized SGD, the optimal learning rates were found to be $1 \times 10^{-2}$ for ResNet-18 (trained from scratch) and $1 \times 10^{-3}$ for CLIP ViT-B/32. When using decentralized AdamW, the optimal learning rate is $5 \times 10^{-4}$ for ResNet-18 (both when trained from scratch and fine-tuned from ImageNet-pretrained weights) and $1 \times 10^{-5}$ for the pretrained CLIP ViT-B/32 on Tiny ImageNet. For all experiments, weight decay is set to $5 \times 10^{-4}$ and the batch size is selected as 128. The key empirical results remain consistent across these optimizer and hyperparameter choices, indicating that our conclusions are stable and not sensitive to specific hyperparameter configurations.  

\textbf{Details of Loss Landscape Visualization in \cref{fig: resnet_landscape}}. To analyze the geometric connections among models after decentralized training, we visualize the loss landscape spanning their parameter spaces. We adopt the visualization tool from \citep{crisostomi2024cm}, which is specifically designed to analyze the interpolation space within the convex hull formed by a given set of models. In our implementation, we position the 16 trained models at the vertices of a regular hexadecagon. Any point within this polygon represents an interpolated model, whose parameters are a weighted sum of the parameters of the 16 vertex models; the weights are determined by the point's Wachspress barycentric coordinates. We then evaluate the cross-entropy loss of each interpolated model on the entire test set to generate the final loss contour map, as shown in \cref{fig: resnet_landscape}. The implementation is available in their notebook \href{https://github.com/crisostomi/cycle-consistent-model-merging/blob/master/notebooks/plots/plot_loss_contours_n_models.ipynb}{https://github.com/crisostomi/cycle-consistent-model-merging/blob/master/notebooks/plots/plot\_loss\_contours\_n\_models.ipynb} within the official code repository for \citep{crisostomi2024cm}. We note two key aspects of this visualization approach:
\begin{itemize}[leftmargin=*]
    \item \textbf{Focus on Convex Combinations.} For points outside the polygon, one or more of their barycentric coordinates become negative, corresponding to an extrapolation, which is often unstable. This visualization approach is consistent with \cref{def:mergeable}, focusing on the space of convex combinations among the models.
    \item \textbf{Deterministic Grid vs. Random Directions.} Notably, the visualization method differs from approaches that use random directions to probe the landscape of a single model, as our visualization grid is defined directly by the 16 models themselves. This allows us to directly investigate the geometric connectivity and interpolation properties among this predefined set of models.
\end{itemize}

\textbf{Computational Resource Requirements and Runtime.} 
To enhance accessibility for researchers working with diverse computational environments, our code includes a centralized simulation of decentralized training. This enables the reproduction and extension of our decentralized learning experiments using fewer GPUs.
A single decentralized AdamW training experiment with 16 agents using ResNet-18 on the Tiny ImageNet dataset requires approximately 15 GB of GPU memory and can be conducted on a single GPU with sufficient memory, such as an NVIDIA V100, RTX 3090, RTX 4090, or A100. On an A100 GPU, the typical runtime is approximately 8 hours for 300 communication rounds, each comprising 100 local steps. For the CLIP ViT-B/32 model, the memory demand rises to about 30 GB, yet it remains feasible on a single A100 GPU, with a runtime of approximately 12 hours under the same configuration of 300 communication rounds and 100 local steps per round.


\subsection{Practical Evaluation Metrics}\label{sec: eval}

The standard evaluation metric of parallel and federated learning is the accuracy of the global model.
\begin{definition}[Test Accuracy of Global Model]\label{def: local_gen_global_model_app}
 The accuracy of the global model $\theta$ is defined as:
\begin{align*}
{\operatorname{Acc}}(\theta)\triangleq \frac{1}{m}\sum_{k \in \mathcal{V}} \mathbb{E}_{\xi_k \sim \mathcal{D}_k} \operatorname{Acc}(\theta; \xi_k)
\overset{\textnormal{\scriptsize if IID}}{=}  \mathbb{E}_{\xi \sim \mathcal{D}} \operatorname{Acc}(\theta; \xi).
\end{align*}
\end{definition}

In decentralized learning, models are often evaluated in the absence of a full consensus model $\theta$ due to data heterogeneity and limited training time.  Two major metrics are adopted in this scenario.

\begin{definition}[Average Local Test Accuracy]\label{def: avg_local_gen_app}
 The average accuracy of agents \(\scalebox{0.9}{$k \in \mathcal{V}$}\) is defined as:
\begin{align*}
\underbrace{
\overline{\operatorname{Acc}}(\{\theta_k\}_{k \in \mathcal{V}}) \triangleq  
\frac{1}{m}\sum_{k \in \mathcal{V}} \mathbb{E}_{\xi_k \sim \mathcal{D}_k} \operatorname{Acc}(\theta_k; \xi_k)
}_{\text{Average Test accuracy on the local distribution across agents}} \overset{\textnormal{\scriptsize if IID}}{=} \frac{1}{m}\sum_{k \in \mathcal{V}} \mathbb{E}_{\xi \sim \mathcal{D}} \operatorname{Acc}(\theta_k; \xi).
\end{align*}
\end{definition}
\begin{tcolorbox}[notitle, rounded corners, colframe=middlegrey, colback=lightred, 
       boxrule=2pt, boxsep=0pt, after skip=7pt, left=0.15cm, right=0.17cm, enhanced, 
       toprule=2pt,
    ]
\begin{remark}[Local Generalization]\label{remark: local_test_app}
    This metric aims to address the following question in decentralized learning: \textit{how well do {local models $\{\theta_k\}_{k \in \mathcal{V}}$}, with the aid of peer-to-peer communication, generalize to their {local (personalized) data distribution $\mathcal{D}_l$}?} This is the standard evaluation metric in personalized decentralized settings, where the goals are to optimize local objectives.
\end{remark}
\end{tcolorbox}

However, in real-world scenarios, local data distributions are often heterogeneous and not guaranteed to be IID across agents.
In such settings, an important goal is to understand how well local models, trained on limited local data, generalize to the global data distribution.
To account for this, we adopt the following \textit{average global test accuracy}, a proxy of average global population risk, as the primary evaluation metric, which quantifies how well local models generalize to the global distribution.
\begin{definition}[Average Global Test Accuracy]\label{def: avg_gen_app}
 The average accuracy of agents \(\scalebox{0.9}{$k \in \mathcal{V}$}\) is defined as:
\begin{align*}
\underbrace{
\overline{\operatorname{Acc}}(\{\theta_k\}_{k \in \mathcal{V}}) = \frac{1}{m}\sum_{k \in \mathcal{V}}\operatorname{Acc}(\theta_k)
}_{\text{Average Accuracy across agents}}
,\quad 
\text{where } 
\operatorname{Acc}(\cdot) \triangleq  
\underbrace{
\frac{1}{m}\sum_{l \in \mathcal{V}} \mathbb{E}_{\xi_l \sim \mathcal{D}_l} \operatorname{Acc}(\cdot; \xi_l)
}_{\text{Test accuracy on the global distribution}}.
\end{align*}
\end{definition}

\begin{tcolorbox}[notitle, rounded corners, colframe=middlegrey, colback=lightred, 
       boxrule=2pt, boxsep=0pt, after skip=7pt, left=0.15cm, right=0.17cm, enhanced, 
       toprule=2pt,
    ]
\begin{remark}[Global Generalization]\label{remark: global_test_app}
    This metric is specifically designed to address a core research question in fully decentralized learning with non-IID data: \textit{how well do {local models $\{\theta_k\}_{k \in \mathcal{V}}$}, trained with limited peer-to-peer synchronization, generalize to the {global data distribution $\mathcal{D}$}?} 
    We note that this objective is particularly critical in the highly non-IID scenarios we study, where local models drift significantly apart. 
    Unlike federated learning that measures the performance of a {global model}, this metric offers a more realistic evaluation for decentralized settings where no central server is present.
\end{remark}
\end{tcolorbox}



\subsection{Additional Experiments}\label{sec: add_exp}
\subsubsection{Different number of agents and optimizers}
We conduct additional experiments by varying the number of agents (from $16$ to $32$) and comparing different optimizers (SGD to AdamW). The effect of single merging remains consistent.
\begin{figure}[ht!]
    \centering

    \begin{subfigure}{.32\textwidth}
        \centering
        \includegraphics[width=\linewidth]{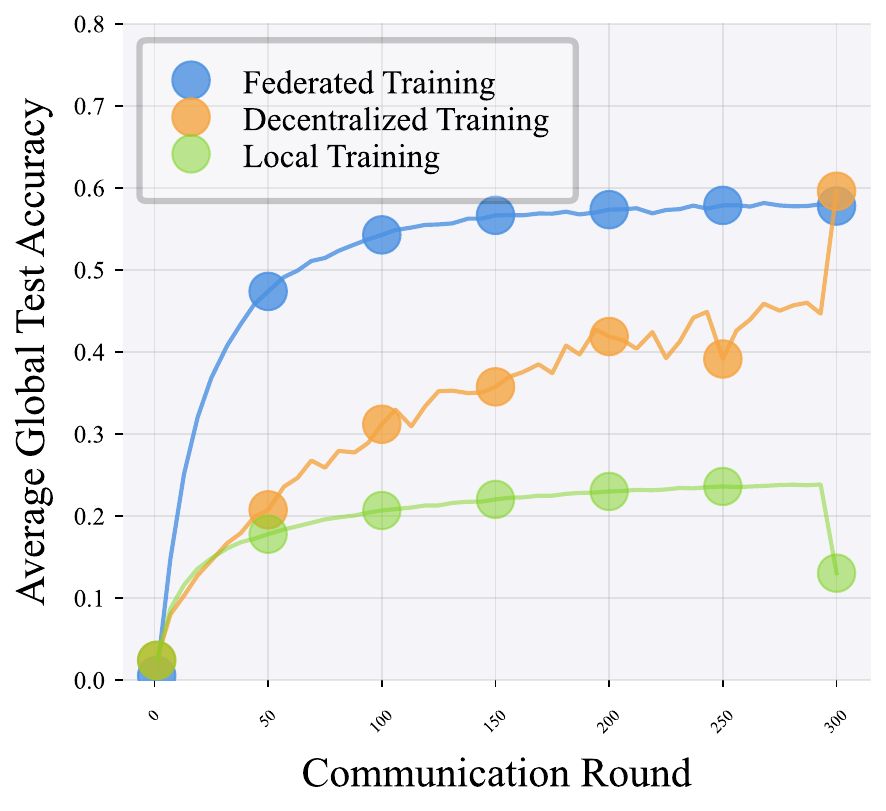}
        \caption{CLIP ViT-B/32}
        \label{fig: main_clip_adamw_16}
    \end{subfigure}
    \hspace{0.05\textwidth} 
    \begin{subfigure}{.32\textwidth}
        \centering
        \includegraphics[width=\linewidth]{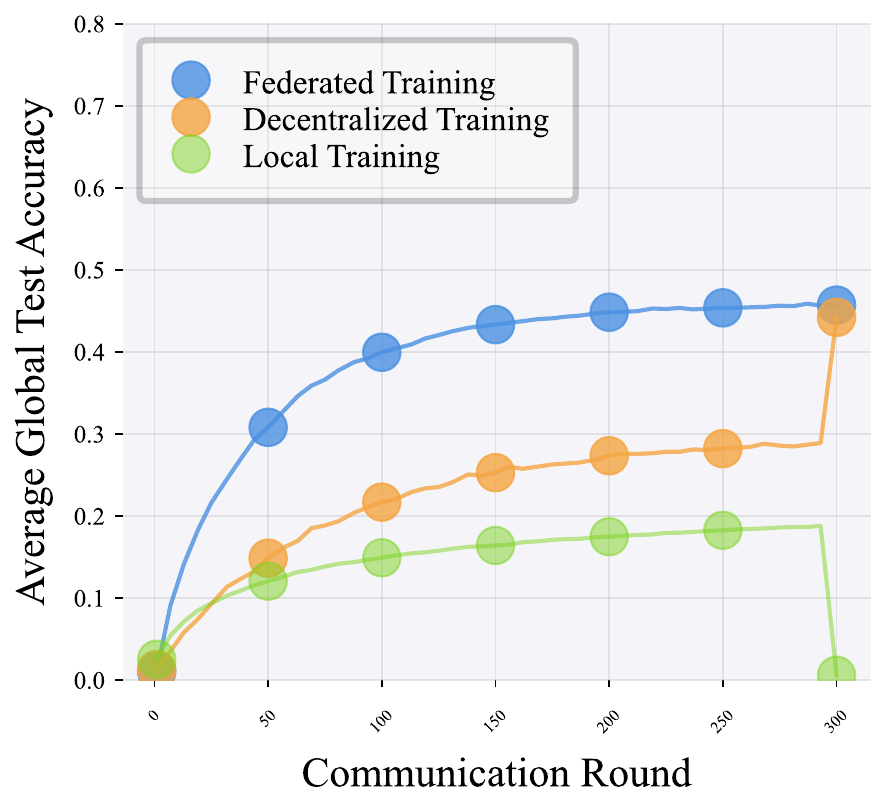}
        \caption{ResNet-18 (w/o pretraining)}
        \label{fig: main_resnet_adamw_16}
    \end{subfigure}

    \caption{
        \textbf{(a, b)}: 
        Global test accuracy (see \cref{def: avg_gen}) of CLIP ViT-B/32 (a) and ResNet-18 (b) trained on Tiny ImageNet using \bluemain{FedAdamW (blue)}, \orangemain{decentralized AdamW (orange)}, and \greenmain{one-shot FedAdamW (green)}, distributed across \textbf{16} agents with high data heterogeneity (Dirichlet $\alpha=0.1$). Decentralized training involves each agent syncing model parameters with a random peer per round with a probability of 0.2, with a single global merging at the final round (see details in \acref{sec: setup}). 
    }
    \label{fig: main_combined_adamw_16}
\end{figure}

\begin{figure}[ht!]
    \centering

    \begin{subfigure}{.32\textwidth}
        \centering
        \includegraphics[width=\linewidth]{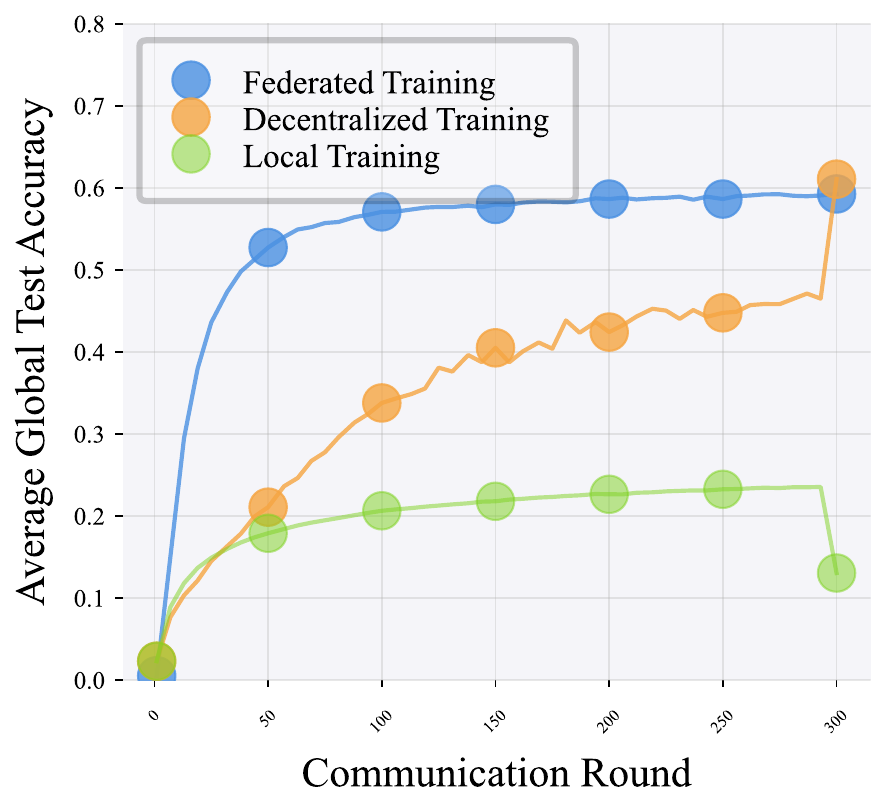}
        \caption{CLIP ViT-B/32}
        \label{fig: main_clip_adamw_32}
    \end{subfigure}
    \hspace{0.05\textwidth} 
    \begin{subfigure}{.32\textwidth}
        \centering
        \includegraphics[width=\linewidth]{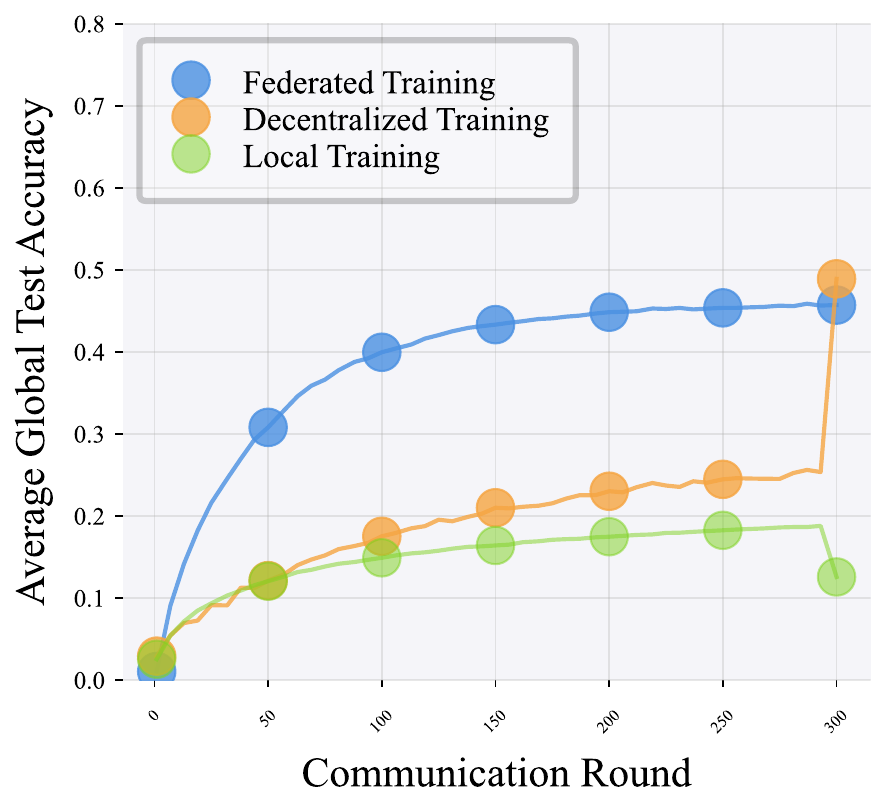}
        \caption{ResNet-18 (w/o pretraining)}
        \label{fig: main_resnet_adamw_32}
    \end{subfigure}
    \caption{
        \textbf{(a, b)}: 
        Global test accuracy (see \cref{def: avg_gen}) of CLIP ViT-B/32 (a) and ResNet-18 (b) trained on Tiny ImageNet using \bluemain{FedAdamW (blue)}, \orangemain{decentralized AdamW (orange)}, and \greenmain{one-shot FedAdamW (green)}, distributed across \textbf{32} agents with high data heterogeneity (Dirichlet $\alpha=0.1$). Decentralized training involves each agent syncing model parameters with a random peer per round with a probability of 0.2, with a single global merging at the final round (see details in \acref{sec: setup}). 
    }
    \label{fig: main_combined_adamw_32}
\end{figure}

\subsubsection{Different communication topologies}
We also conduct additional experiments with different communication topologies to examine whether the empirical results remain consistent. New observations are summarized below.

\begin{figure*}[ht!]
\centering
    \begin{subfigure}{.32\textwidth}
        \centering
        \includegraphics[width=\linewidth]{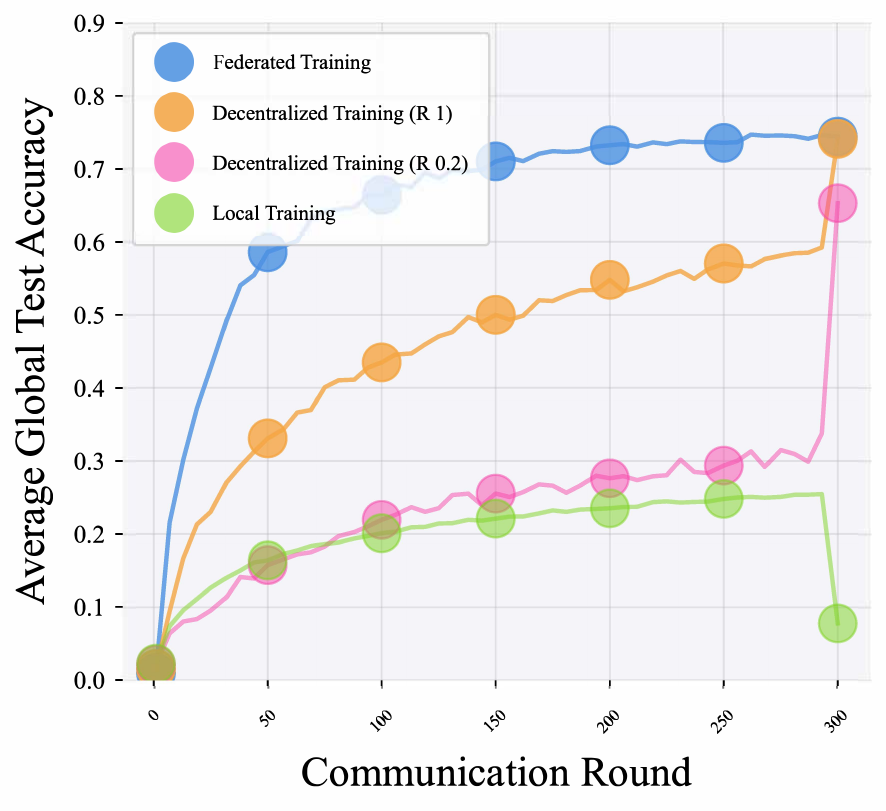}
    \caption{Different Number of Peers $R$}
    \label{fig: resnet18_pretrain}
    \end{subfigure}
    \hspace{0.05\textwidth} 
    \begin{subfigure}{.32\textwidth}
        \centering
        \includegraphics[width=\linewidth]{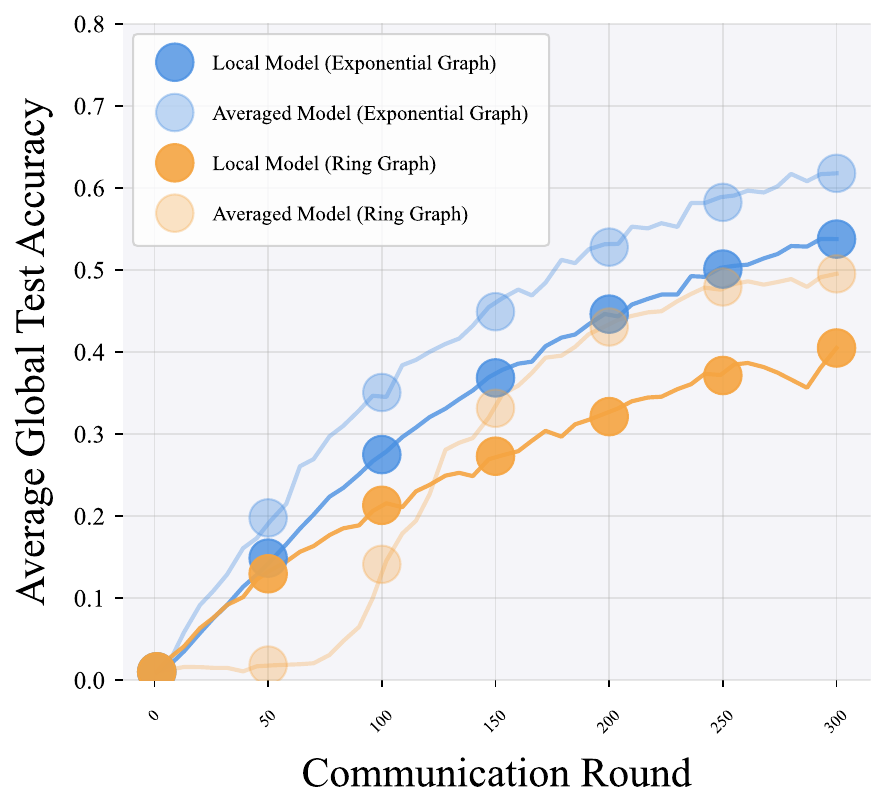}
    \caption{Different Topologies}
    \label{fig: other_topology}
    \end{subfigure}
\caption{Global test accuracy (see \cref{def: avg_gen}) of training ResNet-18 on Tiny ImageNet, distributed across 16 agents with high heterogeneity (Dirichlet $\alpha$ = 0.1; see details in \acref{sec: setup}). We evaluate the effects of different \textbf{(a)} number of peers $R$, and \textbf{(b)} communication topologies. Pretrained weights are used only in (a).}
\label{fig: cifar100_discussion}
\end{figure*}

\textbullet~\textbf{Models remain mergeable under different numbers of peers.} We evaluate two settings (random topology with $R=0.2$ and $R=1$; see ``Communication Graph'' in \acref{sec: setup}).
As shown in \cref{fig: resnet18_pretrain}, performance improvements are consistently observed, with more significant gains in the $R=0.2$ case.

\textbullet~\textbf{Models remain mergeable across different communication topologies.}  
We evaluate two topologies: exponential and ring graphs.  
As shown in \cref{fig: other_topology}, both topologies preserve the mergeability of local models, with exponential graphs yielding slightly better test performance for both local and merged models.  
The trend of mergeability persists across topologies throughout training, though performance may vary.



\subsubsection{Different hyperparameters, dataset, and heterogeneity level}

We further conduct supplementary experiments in which the final global merge is approximated by topology-constrained gossip merging on an exponential graph \citep{ying2021exponential}.

\begin{figure*}[ht!]
\centering
\begin{subfigure}{.31\textwidth}
    \centering
    \includegraphics[width=\linewidth]{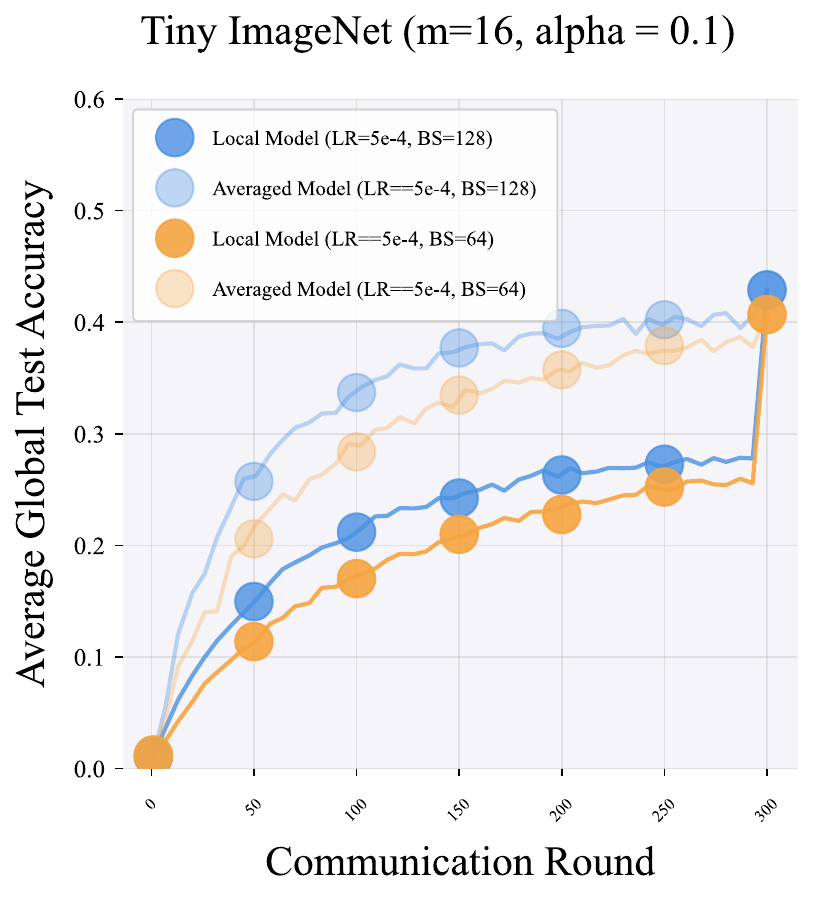}
    \caption{Different Batch Sizes}
    \label{fig: tinyimagenet_bs}
\end{subfigure}
\begin{subfigure}{.31\textwidth}
    \centering
    \includegraphics[width=\linewidth]{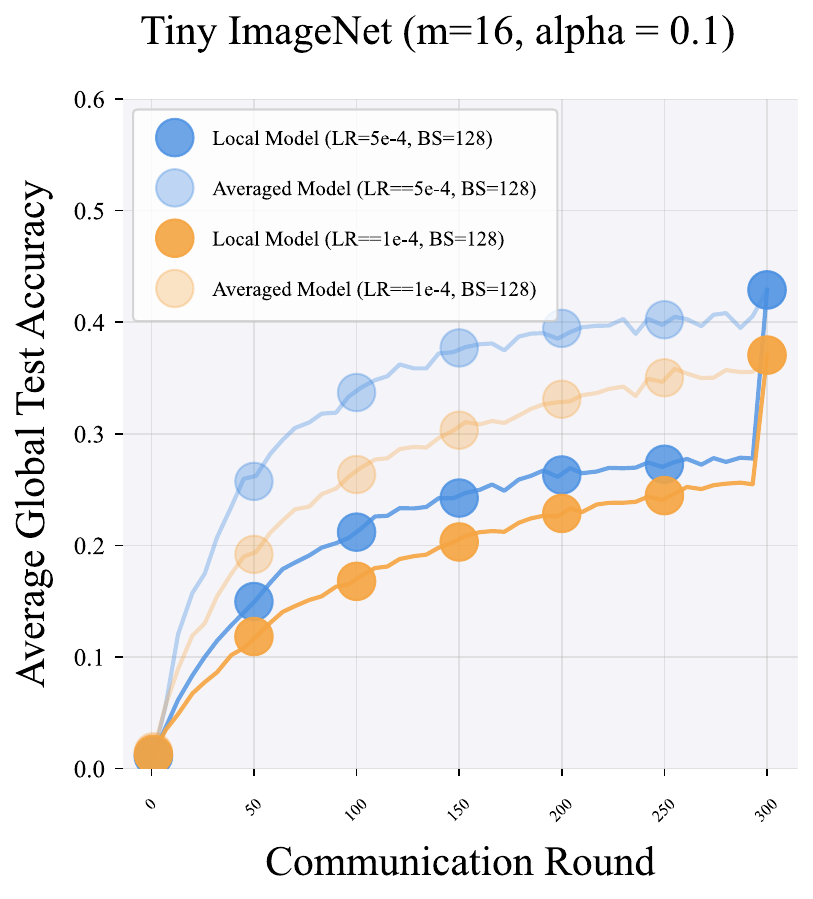}
    \caption{Different Learning Rates}
    \label{fig: tinyimagenet_lr}
\end{subfigure}
\begin{subfigure}{.35\textwidth}
    \centering
    \includegraphics[width=\linewidth]{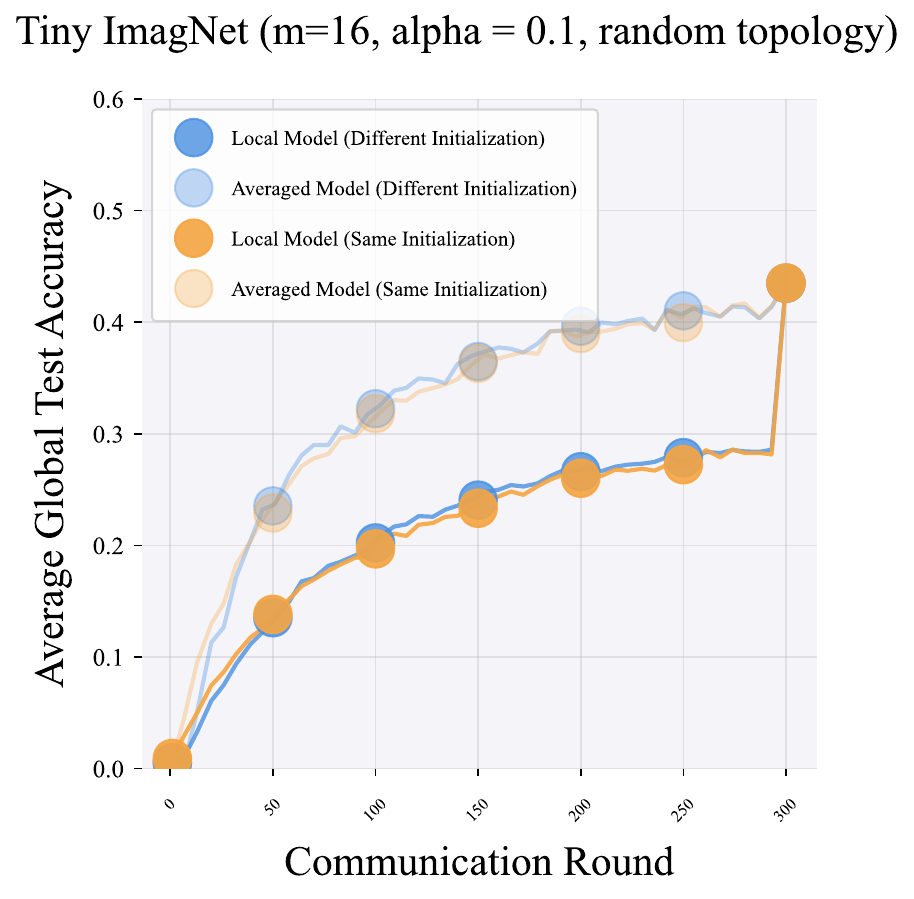}
    \caption{Different Initialization Schemes}
    \label{fig: tinyimagenet_init}
\end{subfigure}
\caption{Global test accuracy (see \cref{def: avg_gen}) of training ResNet-18 on Tiny ImageNet with decentralized AdamW, distributed across 16 agents with high heterogeneity (Dirichlet $\alpha$ = 0.1; see details in \acref{sec: setup}). We evaluate the effects of different  \textbf{(a)} batch sizes  (64 vs. 128),  \textbf{(b)} learning rates ($5 \times 10^{-4}$ vs. $1 \times 10^{-4}$), and \textbf{(c)} different initialization schemes.}
\label{fig: tinyimagenet_lr_bs}
\end{figure*}

\begin{figure*}[ht!]
\centering
\begin{subfigure}{.24\textwidth}
    \centering
    \includegraphics[width=\linewidth]{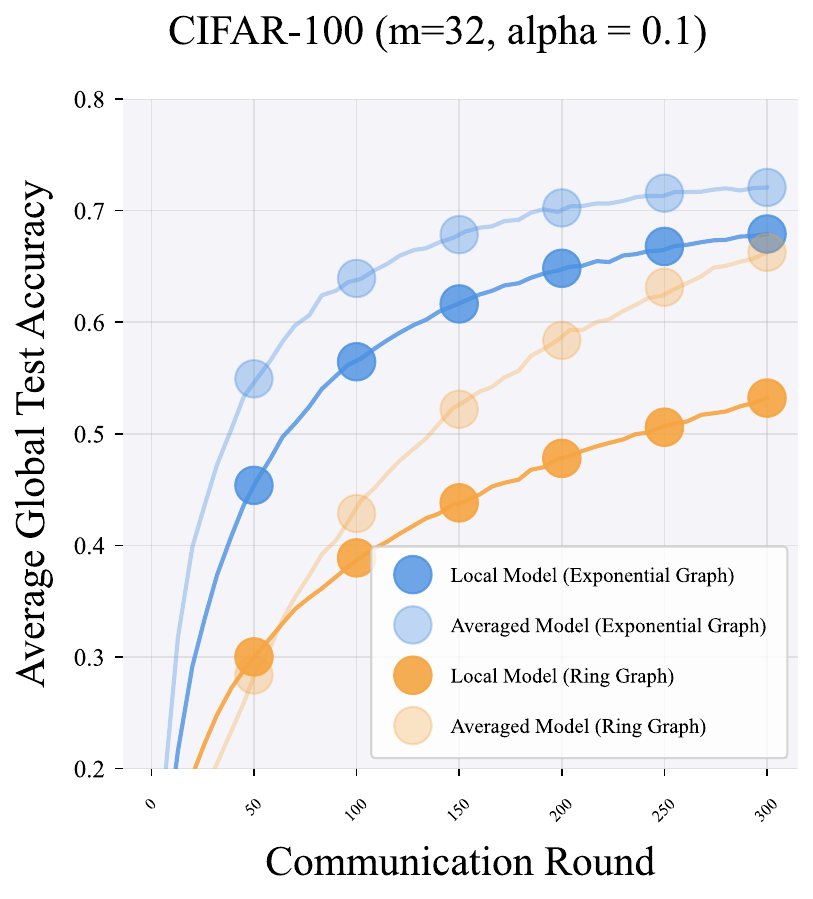}
    \caption{}
    \label{fig: cifar100_alpha_01}
\end{subfigure}
\hfill
\begin{subfigure}{.24\textwidth}
    \centering
    \includegraphics[width=\linewidth]{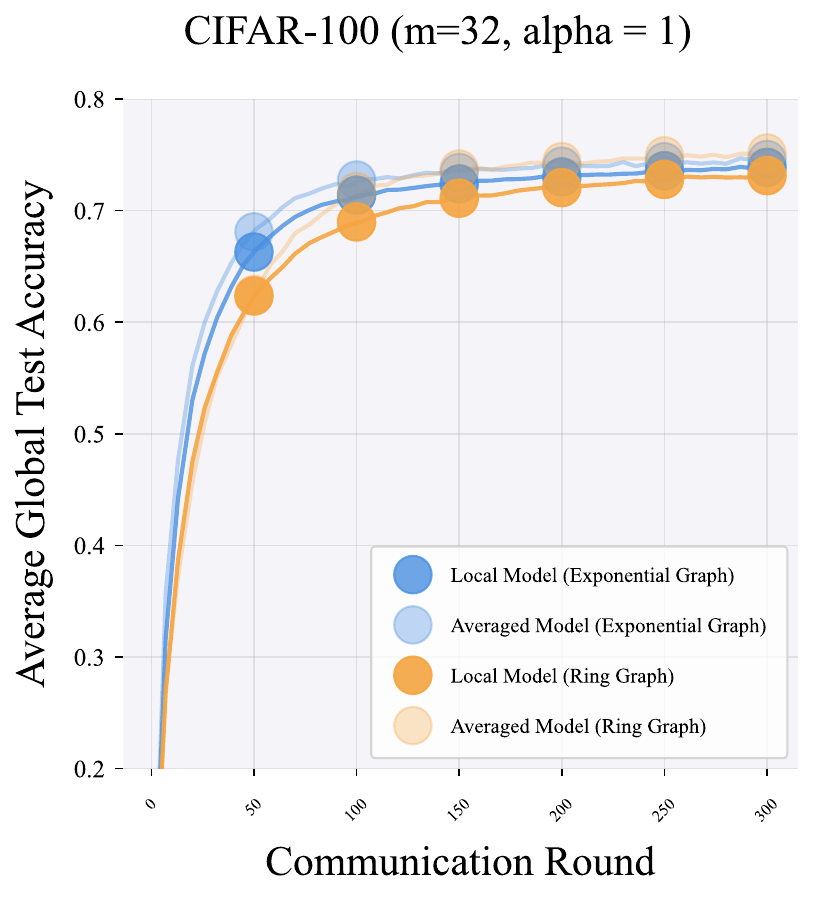}
    \caption{}
    \label{fig: cifar100_alpha_1}
\end{subfigure}
\hfill
\begin{subfigure}{.24\textwidth}
    \centering
    \includegraphics[width=\linewidth]{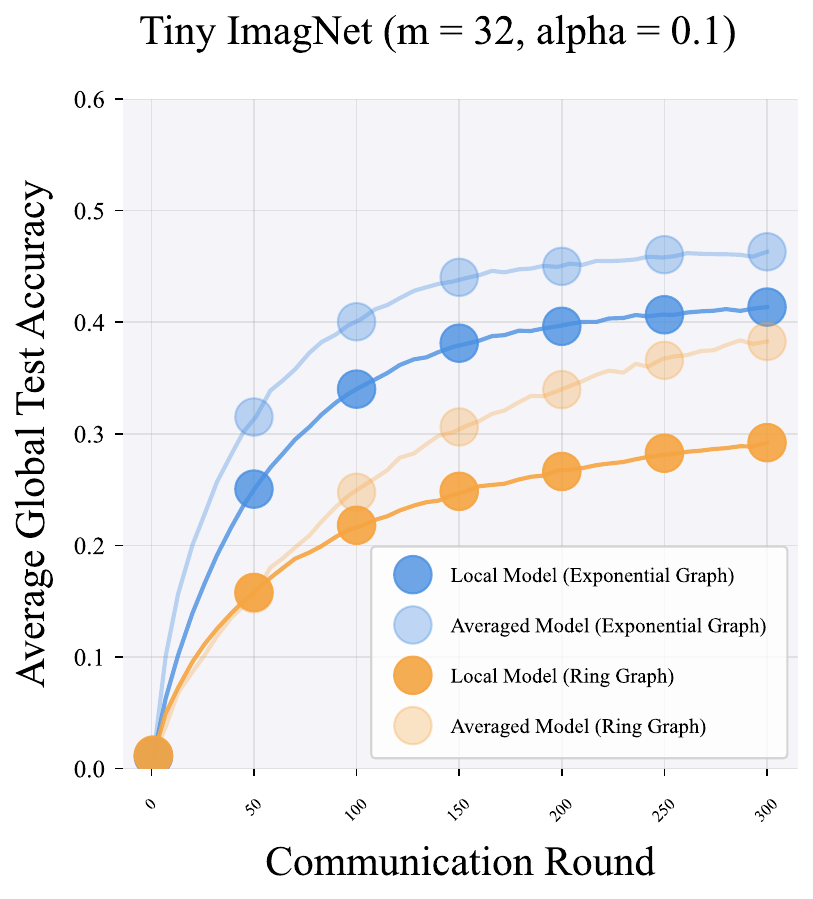}
    \caption{}
    \label{fig: tinyimagenet_alpha_01}
\end{subfigure}
\hfill
\begin{subfigure}{.24\textwidth}
    \centering
    \includegraphics[width=\linewidth]{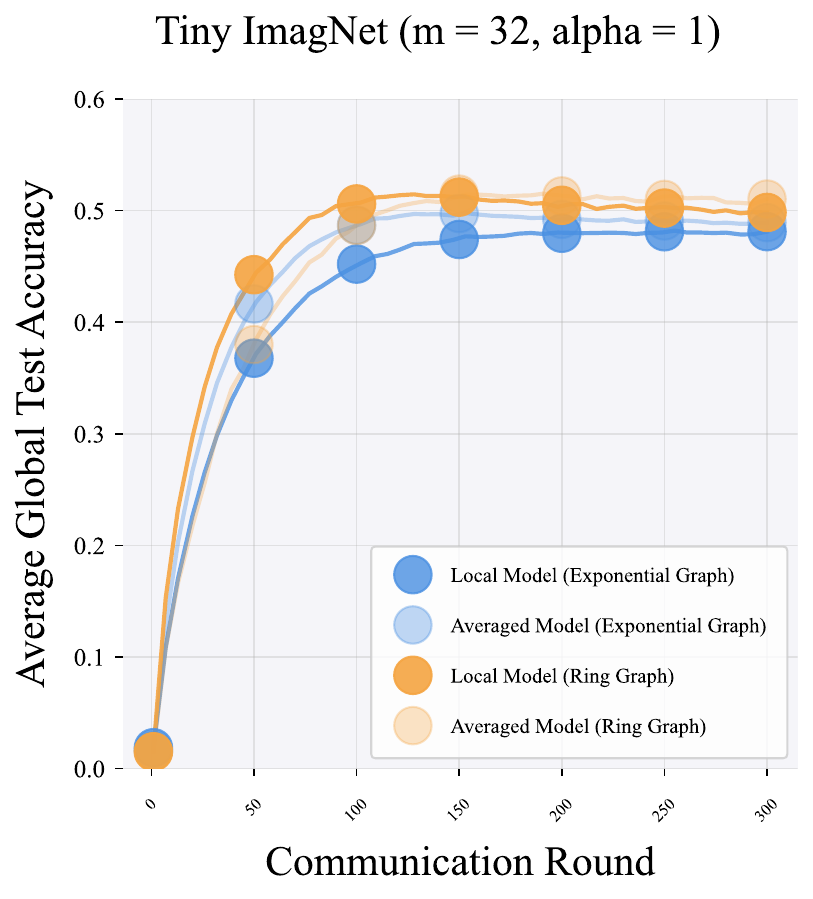}
    \caption{}
    \label{fig: tinyimagenet_alpha_1}
\end{subfigure}

\caption{
Global test accuracy (see \cref{def: avg_gen}) for ResNet-18 trained with decentralized AdamW across 32 agents under different levels of data heterogeneity (Dirichlet $\alpha = 0.1$ (\textbf{a}, \textbf{c}) vs. $\alpha = 1.0$ (\textbf{b}, \textbf{d}); see \acref{sec: setup}). Results are reported on both CIFAR-100 (\textbf{a}, \textbf{b}) and Tiny ImageNet (\textbf{c}, \textbf{d}).
}
\label{fig: alpha_sensitivity}
\end{figure*}





\textbf{Summary}. Consistent test performance improvement of a single global merging across a wide range of settings are observed, including different hyperparameter setups, datasets, degree of data heterogeneity, model architectures, optimizers, initialization schemes, and communication topologies.


\subsubsection{Realizing Final Global Gossip via Topology-Constrained Merging}\label{sec: gossip_merging}

We also perform additional experiments where the final global merging is approximated by multiple rounds of gossip merging, i.e.,  a specific exponential topology \citep{ying2021exponential}.

\begin{figure*}[ht!]
\centering
\begin{subfigure}{.31\textwidth}
    \centering
    \includegraphics[width=\linewidth]{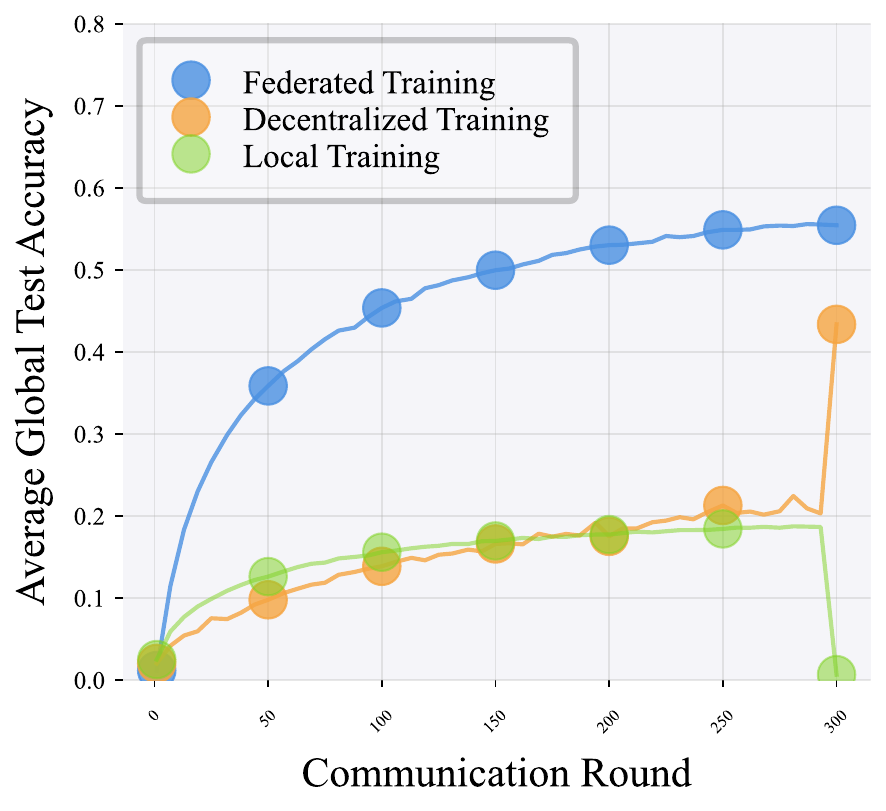}
    \caption{1 Round Final Gossip Merging}
    \label{fig: gossip_merge_1}
\end{subfigure}
\begin{subfigure}{.31\textwidth}
    \centering
    \includegraphics[width=\linewidth]{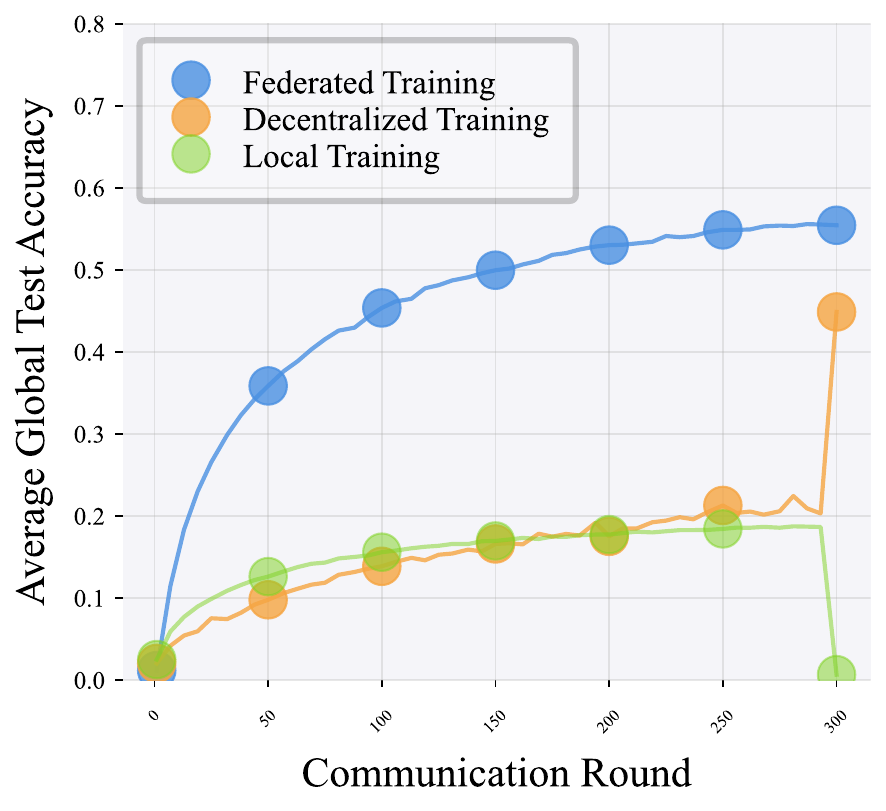}
    \caption{5 Rounds Final Gossip Merging}
    \label{fig: gossip_merge_2}
\end{subfigure}
\begin{subfigure}{.31\textwidth}
    \centering
    \includegraphics[width=\linewidth]{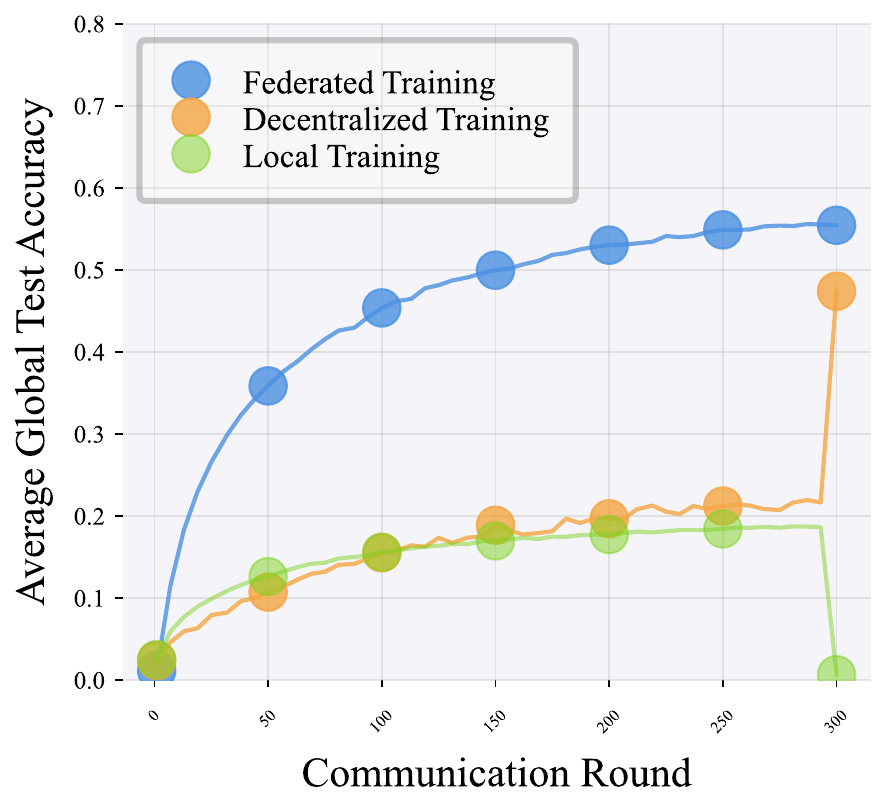}
    \caption{1 Round Final Global Merging}
    \label{fig: gossip_merge_baseline}
\end{subfigure}
\caption{Global test accuracy (see \cref{def: avg_gen}) for ResNet-18 trained with decentralized AdamW on Tiny ImageNet (32 agents). Left: regular decentralized training followed by one round of topology-constrained final gossip merging on an exponential graph. Middle: the same topology-constrained setting, but with five rounds of final gossip merging to better approximate global aggregation. Right: baseline setting (our original approach), with random communication among all agents during training followed by one perfect global merge.}
\label{fig: gossip_merge}
\end{figure*}

\textbf{Summary}. We observe that (1) even a single round of topology-constrained final gossip merging substantially improves global test accuracy, and (2) the resulting performance is comparable to the baseline that uses random communication among all agents followed by one perfect global merge.

\newpage
\section{Theory}\label{sec: proof}
This section provides the proofs of the main theoretical results presented in this paper. For simplicity, and following the setup in the existing literature, we assume that the sample size of local agents is $n_k = n$ for all $k \in \mathcal{V}$.

\begin{tcolorbox}[notitle, rounded corners, colframe=middlegrey, colback=lightblue, 
       boxrule=2pt, boxsep=0pt, left=0.15cm, right=0.17cm, enhanced, 
       toprule=2pt,
    ]
\begin{lemma}[Consensus Distance Recursion under Local Updates \citep{pmlr-v139-kong21a}]
\label{lemma:consensus_distance_local_updates}
Suppose \cref{ass: mixing}–\cref{ass: bounded_noise_diversity} hold. Let $\theta_k^{(t)}$ be the local parameter on client $k$ at $t$-th step, and denote their average by $\bar\theta^{(t)}=\frac1m\sum_{k=1}^m\theta_k^{(t)}$. Define the \emph{consensus distance} and the \emph{average gradient norm} at round $t$ by
$\Xi_t^2=\frac1m\sum_{k=1}^m\|\theta_k^{(t)}-\bar\theta^{(t)}\|^2$
and
$\phi_t^2=\frac1m\sum_{k=1}^m\|\nabla\mathcal{L}_k(\theta_k^{(t)})\|^2$,
where $\mathcal{L}_k(\theta)=\mathbb{E}_{\xi_k\sim\mathcal{D}_k}[\mathcal{L}(\theta;\xi_k)]$. Let $\eta>0$ the learning rate, and $\sigma^2$ the variance bound from \cref{ass: bounded_noise_diversity}. Then there exists a constant $p>0$ (see \cref{ass: mixing}) such that for all $t\ge0$, the following inequality holds:
\begin{align}
    \mathbb{E}\left[\Xi_{t+1}^2 \right] \leq \left(1 - \frac{p}{2}\right)\Xi_t^2 + \frac{12(1-p)}{p} \eta^2 \left(\phi_t^2 + \sigma^2\right),
\end{align}
where the expectation is taken over the stochastic gradients in the \(t\)-th update phase.
\end{lemma}
\end{tcolorbox}

\begin{proof}
For completeness, we provide the proof of \cref{lemma:consensus_distance_local_updates}, with minor corrections and additional details.
In decentralized SGD (\cref{alg:decentralized_learning} with SGD as the local optimizer), each agent \(k\in\mathcal V\) performs at each iteration
\begin{align*}
\theta_k^{(t+1)}
=\sum_{l=1}^m W_{k,l}\bigl(\theta_l^{(t)}-\eta\,\nabla\mathcal L_l(\theta_l^{(t)};\,\xi_l^{(t)})\bigr).   
\end{align*}

In matrix form, letting
\begin{align*}
\Theta^{(t)}=[\theta_1^{(t)},\dots,\theta_m^{(t)}]\in\mathbb R^{d\times m},
\quad
\nabla\mathcal L(\Theta^{(t)};\xi^{(t)})=[\nabla\mathcal L_1(\theta_1^{(t)};\xi_1^{(t)}),\dots,\nabla\mathcal L_m(\theta_m^{(t)};\xi_m^{(t)})],
\end{align*}
we have
\begin{align*}
\Theta^{(t+1)}
=\bigl(\Theta^{(t)}-\eta\,\nabla\mathcal L(\Theta^{(t)};\xi^{(t)})\bigr)\,W.
\end{align*}

The consensus matrix after mixing is
\begin{align*}
\bar\Theta^{(t+1)}
=\Theta^{(t+1)}\;\tfrac1m\mathbf1\mathbf1^\top
=\bigl(\Theta^{(t)}-\eta\,\nabla\mathcal L(\Theta^{(t)};\xi^{(t)})\bigr)\;\tfrac1m\mathbf1\mathbf1^\top,
\end{align*}
since \(\mathbf1^\top W=\mathbf1^\top\).

Thus the consensus distance satisfies
\begin{align*}
m\,\Xi_{t+1}^2
=\bigl\|\Theta^{(t+1)}-\bar\Theta^{(t+1)}\bigr\|_F^2
=\bigl\|\bigl(\Theta^{(t)}-\eta\,\nabla\mathcal L(\Theta^{(t)};\xi^{(t)})\bigr)
\,(W-\tfrac1m\mathbf1\mathbf1^\top)\bigr\|_F^2.
\end{align*}

By \cref{ass: mixing}, for any \(\Theta\in\mathbb R^{d\times m}\),
\begin{align*}
\mathbb E_W\bigl\|\Theta W-\bar\Theta\bigr\|_F^2
\le(1-\rho)\bigl\|\Theta-\bar\Theta\bigr\|_F^2,
\end{align*}
we obtain,
\begin{align*}
m\,\Xi_{t+1}^2
\le(1-p)\Bigl\|\Theta^{(t)}(I-\tfrac1m\mathbf1\mathbf1^\top)
-\eta\,\nabla\mathcal L(\Theta^{(t)};\xi^{(t)})(I-\tfrac1m\mathbf1\mathbf1^\top)\Bigr\|_F^2.
\end{align*}

Applying the inequality \(\|A+B\|_F^2\le(1+\alpha)\|A\|_F^2+(1+1/\alpha)\|B\|_F^2\) with \(\alpha=\tfrac p2\) gives
\begin{align*}
m\,\Xi_{t+1}^2
&\le(1-p)\Bigl[(1+\tfrac p2)\big\|\Theta^{(t)}(I-\tfrac1m\mathbf1\mathbf1^\top)\big\|_F^2
+(1+\tfrac2p)\,\eta^2\,\big\|\nabla\mathcal L(\Theta^{(t)};\xi^{(t)})\big\|_F^2\Bigr]\\
&\le\Bigl(1-\tfrac p2\Bigr)\,m\,\Xi_t^2
\;+\;\frac{6(1-p)}{p}\,\eta^2\,\big\|\nabla\mathcal L(\Theta^{(t)};\xi^{(t)})\big\|_F^2,
\end{align*}
where we use \((1+p/2)\le1+p\) and \((1+2/p)\le6/p\) for \(p\in(0,1)\).

We now decompose the stochastic gradient as
\begin{align*}
\nabla\mathcal L(\Theta^{(t)};\xi^{(t)})
=\nabla\mathcal L(\Theta^{(t)})+\bigl[\nabla\mathcal L(\Theta^{(t)};\xi^{(t)})-\nabla\mathcal L(\Theta^{(t)})\bigr],
\end{align*}
so by Young's Inequality, we have
\begin{align*}
\big\|\nabla\mathcal L(\Theta^{(t)};\xi^{(t)})\big\|_F^2
\le2\big\|\nabla\mathcal L(\Theta^{(t)})\big\|_F^2
+2\big\|\nabla\mathcal L(\Theta^{(t)};\xi^{(t)})-\nabla\mathcal L(\Theta^{(t)})\big\|_F^2.
\end{align*}

Taking expectation over \(\xi^{(t)}\) and invoking \cref{ass: bounded_noise_diversity}, we get
\begin{align*}
\mathbb E\bigl[\|\nabla\mathcal L(\Theta^{(t)};\xi^{(t)})\|_F^2\bigr]
\le2\|\nabla\mathcal L(\Theta^{(t)})\|_F^2+2\,\sigma^2m.
\end{align*}

Substituting back and dividing by \(m\) yields
\begin{align*}
\mathbb E\bigl[\Xi_{t+1}^2\bigr]
\le\Bigl(1-\tfrac p2\Bigr)\,\Xi_t^2
\;+\;\frac{12(1-p)}{p}\,\eta^2\bigl(\phi_t^2+\sigma^2\bigr),
\end{align*}
which completes the proof.
\end{proof}

\begin{tcolorbox}[notitle, rounded corners, colframe=middlegrey, colback=lightblue, 
       boxrule=2pt, boxsep=0pt, left=0.15cm, right=0.17cm, enhanced, 
       toprule=2pt,
    ]
\begin{corollary}[Upper Bounds of Consensus Distance \citep{pmlr-v139-kong21a}]\label{coro: consensus_distance}
Define the \emph{consensus distance} and the \emph{average gradient norm} at round $t$ by
$\Xi_t^2=\frac1m\sum_{k=1}^m\|\theta_k^{(t)}-\bar\theta^{(t)}\|^2$
and
$\phi_t^2=\frac1m\sum_{k=1}^m\|\nabla\mathcal{L}_k(\theta_k^{(t)})\|^2$,
where $\mathcal{L}_k(\theta)=\mathbb{E}_{\xi_k\sim\mathcal{D}_k}[\mathcal{L}(\theta;\xi_k)]$.
Under the conditions of \cref{lemma:consensus_distance_local_updates}, suppose that for all iterations \( t \), the gradient norms are uniformly bounded by a constant \( \phi \), i.e.\ \( \phi_t^2 \le \phi^2,\ \forall t\in\{1,\dots,T\}\). Then the expected consensus distance satisfies
\[
  \mathbb{E}\bigl[\Xi_t^2\bigr]
  \;\le\;
  \frac{24\,(1-p)\,\eta^2}{p^2}
  \Bigl({\phi^2} + {\sigma^2}\Bigr).
\]
In the general case where the gradient‐norms change slowly, i.e.,
\(\phi_t^2 \le (1 + \tfrac p4)\,\phi_{t+1}^2\), we have
\[
  \mathbb{E}\bigl[\Xi_t^2\bigr]
  \;\le\;
  \frac{48\,(1 - p)\eta^2 }{p^2}  \left({\phi^2_{t-1}} + {\sigma^2}\right).
\]
The expectation here is taken over the stochastic gradients in the \(t\)-th update phase.
\end{corollary}
\end{tcolorbox}

\begin{proof}
Consider the key recursion from \cref{lemma:consensus_distance_local_updates}:
\[
  \mathbb{E}\bigl[\Xi_{t+1}^2\bigr]
  \;\le\;
  \Bigl(1 - \tfrac p2\Bigr)\,\Xi_t^2
  \;+\;
  \frac{12\,(1-p)}{p}\,\eta^2
  \bigl(\phi_t^2 + \sigma^2\bigr).
\]

\textbf{(1) Special Case: uniformly bounded gradient norms.}

Assume \(\phi_t^2\le\phi^2\).  Unrolling the above gives
\[
  \mathbb{E}\bigl[\Xi_{t+1}^2\bigr]
  \;\le\;
  \sum_{i=0}^{t-1}
    \Bigl(1 - \tfrac p2\Bigr)^i
    \frac{12\,(1-p)}{p}\,\eta^2\,(\phi^2 + \sigma^2).
\]
Since 
\(\sum_{i=0}^{t-1}(1-\tfrac p2)^i \le \frac{2}{p}\),
we can bound the consensus distance as
\[
  \mathbb{E}\bigl[\Xi_{t+1}^2\bigr]
  \;\le\;
  \frac{12\,(1-p)}{p}\,\eta^2(\phi^2+\sigma^2)
  \;\times\;
  \frac{2}{p}
  \;=\;
  \frac{24\,(1-p)\,\eta^2}{p^2}\,
  (\phi^2 + \sigma^2),
\]
which yields the first claim.

\textbf{(2) Special Case: slowly changing gradient norms.}

If \(\phi_t^2\le(1+\tfrac p4)\,\phi_{t+1}^2\), and since
\[
  \Bigl(1 - \tfrac p2\Bigr)^i
  \Bigl(1 + \tfrac p4\Bigr)^i
  \;\le\;
  \Bigl(1 - \tfrac p4\Bigr)^i,
\]
 the consensus distance satisfies
\begin{align}
    \mathbb{E}\left[ \Xi_{t+1}^2 \right] &\leq \sum_{i=0}^{t-i-1} \left(1 - \frac{p}{2}\right)^i \frac{12(1 - p) \eta^2  \left(\phi_{t-1}^2 +  \sigma^2\right)}{p} \nonumber\\
    &\leq \sum_{i=0}^{t-1} \left(1 - \frac{p}{4}\right)^i \frac{12(1 - p) \eta^2 (\phi_{t-1}^2 +  \sigma^2)}{p} \leq \frac{48(1 - p)\eta^2 }{p^2}  \left({\phi^2_{t-1}} + {\sigma^2}\right).
\end{align}
\end{proof}

\begin{tcolorbox}[notitle, rounded corners, colframe=middlegrey, colback=lightblue, 
       boxrule=2pt, boxsep=0pt, left=0.15cm, right=0.17cm, enhanced, 
       toprule=2pt,
    ]
\begin{proposition}[Implicit Bias of Decentralized SGD \citep{pmlr-v202-zhu23e}]\label{prop: implicit_bias}
Suppose \cref{ass: regularity} hold,
the globally averaged model of decentralized SGD (DSGD), defined by
\(\bar{\theta}^{(t)} = \frac{1}{m}\sum_{k=1}^m \theta_k^{(t)},\)
follows the following gradient descent direction:
\begin{align*}
\mathbb{E}_{\xi^{(t)}}[\bar{\theta}^{(t+1)}]
= \bar{\theta}^{(t)} - \eta \cdot \mathbb{E}_{\epsilon^{(t)} \sim \mathcal{N}(0, {\Gamma}^{(t)})}\left[\nabla\mathcal{L}({\bar{\theta}^{(t)}+\epsilon^{(t)}})\right]
+ {\delta}^{(t)},
\end{align*}
where \({\Gamma}^{(t)} = \frac{1}{m}\sum_{k=1}^{m}(\theta_k^{(t)} - \bar{\theta}^{(t)})(\theta_k^{(t)} - \bar{\theta}^{(t)})^{\top} \in \mathbb{R}^{m\times m}\)
denotes the consensus distance matrix, and ${\delta}^{(t)} = O\left( \frac{\eta}{m}\sum_{k=1}^{m}\|\theta_k^{(t)} - \bar{\theta}^{(t)}\|_2^3\right)$ denotes the high-order terms. The first expectation eliminates the randomness from sampled data $\xi^{(t)}=\{\xi^{(t)}_k\}_{k\in\mathcal{V}}$  at step $(t)$.
\end{proposition}
\end{tcolorbox}
We can then control the expected squared distance between two consecutive steps of the globally averaged model with \cref{coro: distance}.

\begin{tcolorbox}[notitle, rounded corners, colframe=middlegrey, colback=lightblue, 
       boxrule=2pt, boxsep=0pt, left=0.15cm, right=0.17cm, enhanced, 
       toprule=2pt,
    ]
\begin{corollary}\label{coro: distance}
   Under the assumptions in \cref{prop: implicit_bias}, the expected squared distance between two consecutive iterates of decentralized SGD can be bounded as follows:
    \begin{align}
        \mathbb{E}_{\xi^{(t)}}\,
        \big\|\bar{\theta}^{(t+1)}-\bar{\theta}^{(t)}\big\|^{2} \leq 
        \frac{\eta^2\sigma^2}{m} + \eta^2 \left\| \nabla\mathcal{L}\big(\bar{\theta}^{(t)}\big)+ \frac{1}{2}\nabla\,\operatorname{Tr}\big(\nabla^2 \mathcal{L}(\bar{\theta}^{(t)}) \Gamma^{(t)}\big)
+ {\delta}^{(t)}\right\|^2.
    \end{align}
\end{corollary}
\end{tcolorbox}

\begin{proof}
Denote $\gamma^{(t+1)} = \mathbb{E}_{\xi^{(t)}}\bar{\theta}^{(t+1)} -\bar{\theta}^{(t)}$. We can expand the expected distance as follows:
\begin{align}
    \mathbb{E}_{\xi^{(t)}}\,
    \big\|\bar{\theta}^{(t+1)}-\bar{\theta}^{(t)}\big\|^{2} &= 
    \mathbb{E}_{\xi^{(t)}}\,
    \big\|\bar{\theta}^{(t+1)}\big\|^{2} + 
    \big\|\bar{\theta}^{(t)}\big\|^{2} -2
    (\bar{\theta}^{(t)})^\top \gamma^{(t+1)} \nonumber\\
    &=\operatorname{Tr}\big(\text{Cov}(\bar{\theta}^{(t+1)})\big)
    + \big\|\mathbb{E}_{\xi^{(t)}}\,\bar{\theta}^{(t+1)}\big\|^{2} + 
    \big\|\bar{\theta}^{(t)}\big\|^{2} -2
    (\bar{\theta}^{(t)})^\top \gamma^{(t+1)} \nonumber\\
    & = \operatorname{Tr}\big(\text{Cov}(\bar{\theta}^{(t+1)})\big) + \big\|\mathbb{E}_{\xi^{(t)}}[\bar{\theta}^{(t+1)}-\bar{\theta}^{(t)}]\big\|^{2}\nonumber\\
    & = \operatorname{Tr}\left(\text{Cov}(\frac{\eta}{m}\sum_{k=1}^m \nabla\mathcal{L}(\theta_k^{(t)};\xi_k^{(t)}))\right) + \big\|\mathbb{E}_{\xi^{(t)}}[\bar{\theta}^{(t+1)}-\bar{\theta}^{(t)}]\big\|^{2},
\end{align}
where the second equality follows from the definition of the covariance matrix, namely $$\operatorname{Tr}\big(\text{Cov}(\bar{\theta}^{(t+1)})\big) = \mathbb{E}_{\xi^{(t)}}\,
    \big\|\bar{\theta}^{(t+1)}\big\|^{2} - \big\|\mathbb{E}_{\xi^{(t)}}\,\bar{\theta}^{(t+1)}\big\|^{2},$$ and the final equality is derived from the original update of decentralized SGD (without applying \cref{prop: implicit_bias}):
\[
\bar{\theta}^{(t+1)} = \bar{\theta}^{(t)} - \eta\cdot\frac{1}{m}\sum_{k=1}^m \nabla\mathcal{L}(\theta_k^{(t)};\xi_k^{(t)}).
\]

According to the convexity of the vector norm and the fact that 
\begin{align}
    \operatorname{Tr}\left(\text{Cov}(\frac{1}{m}\sum_{k=1}^m \nabla\mathcal{L}(\theta_k^{(t)};\xi_k^{(t)}))\right) = \mathbb{E}_{\xi^{(t)}} \left\|\frac{1}{m}\sum_{k=1}^m \nabla\mathcal{L}(\theta_k^{(t)};\xi_k^{(t)}) - \frac{1}{m}\sum_{k=1}^m \mathbb{E}_{\xi_k^{(t)}}\nabla\mathcal{L}(\theta_k^{(t)}; \xi_k^{(t)})\right\|^2,
\end{align}
we then complete the proof by applying \cref{prop: implicit_bias} and the bounded noise assumption in \cref{ass: bounded_noise_diversity}.
\end{proof}

\begin{tcolorbox}[notitle, rounded corners, colframe=middlegrey, colback=lightblue, 
       boxrule=2pt, boxsep=0pt, left=0.15cm, right=0.17cm, enhanced, 
       toprule=2pt,
    ]
\begin{corollary} \label{coro:gradient_diff}
Let \(\Gamma^{(t)} = \frac{1}{m} \sum_{k=1}^{m} (\theta_k^{(t)} - \bar{\theta}^{(t)})(\theta_k^{(t)} - \bar{\theta}^{(t)})^{\top} \in \mathbb{R}^{d \times d}\), where \(\bar{\theta}^{(t)} = \frac{1}{m} \sum_{k=1}^{m} \theta_k^{(t)} \in \mathbb{R}^d\) denotes the globally averaged model across \(m\) agents. 
Suppose \cref{ass: regularity} hold. 
Then, for \(\epsilon^{(t)} \sim \mathcal{N}(0, \Gamma^{(t)})\), the expected gradient perturbation satisfies:
\begin{align}
\mathbb{E}_{\epsilon^{(t)} \sim \mathcal{N}(0, \Gamma^{(t)})} &\left[ \nabla \mathcal{L}(\bar{\theta}^{(t)} + \epsilon^{(t)}) \right] - \nabla \mathcal{L}(\bar{\theta}^{(t)})\nonumber\\ &= \frac{1}{2}\nabla \operatorname{Tr} \left( \nabla^2 \mathcal{L}(\bar{\theta}^{(t)}) \Gamma^{(t)} \right) + \mathbb{E}_{\epsilon^{(t)} \sim \mathcal{N}(0, \Gamma^{(t)})} \left[ R_3(\epsilon^{(t)}) \right],
\end{align}
where \(\| R_3(\epsilon^{(t)}) \|\) is bounded by $\frac{L_4}{24} \|\epsilon^{(t)}\|^3$.
\end{corollary}
\end{tcolorbox}

\begin{proof}
We apply the third-order Taylor expansion to \( \nabla \mathcal{L} \) around \( \bar{\theta}^{(t)} \):
\[
\nabla \mathcal{L}(\bar{\theta}^{(t)} + \epsilon^{(t)}) = \nabla \mathcal{L}(\bar{\theta}^{(t)}) + \nabla^2 \mathcal{L}(\bar{\theta}^{(t)}) \epsilon^{(t)} + \frac{1}{2} \nabla^3 \mathcal{L}(\bar{\theta}^{(t)}) [\epsilon^{(t)}, \epsilon^{(t)}] + R_3(\epsilon^{(t)}),
\]
with the remainder:
\[
R_3(\epsilon^{(t)}) = \int_0^1 \frac{(1 - \tau)^3}{6} \nabla^4 \mathcal{L}(\bar{\theta}^{(t)} + \tau \epsilon^{(t)}) [\epsilon^{(t)}, \epsilon^{(t)}, \epsilon^{(t)}] d\tau.
\]
Taking expectations over \(\epsilon^{(t)} \sim \mathcal{N}(0, \Gamma^{(t)})\), since \(\mathbb{E}[\epsilon^{(t)}] = 0\), the linear term vanishes. The quadratic term \(\mathbb{E} \left[ \nabla^3 \mathcal{L}(\bar{\theta}^{(t)}) [\epsilon^{(t)}, \epsilon^{(t)}] \right]\) simplifies to \(\nabla \operatorname{Tr} \left( \nabla^2 \mathcal{L}(\bar{\theta}^{(t)}) \Gamma^{(t)} \right)\) due to properties of the Gaussian distribution. The remainder bound can be bounded as
\[
\| R_3(\epsilon^{(t)}) \| \leq \int_0^1 \frac{(1 - \tau)^3}{6} L_4 \|\epsilon^{(t)}\|^3 d\tau = L_4 \|\epsilon^{(t)}\|^3 \cdot \frac{1}{6} \int_0^1 (1 - \tau)^3 d\tau.
\]
Since \(\int_0^1 (1 - \tau)^3 d\tau = \frac{1}{4}\), we have:
\[
\| R_3(\epsilon^{(t)}) \| \leq L_4 \|\epsilon^{(t)}\|^3 \cdot \frac{1}{6} \cdot \frac{1}{4} = \frac{L_4}{24} \|\epsilon^{(t)}\|^3.
\]
\end{proof}

For comparison, we restate the convergence rate of DSGD by \citet{pmlr-v119-koloskova20a}.

\begin{assumption}[$L$-smoothness]\label{ass: l-smooth}
Each population risk \(\mathcal{L}_k = \mathbb{E}_{\xi_k \sim \mathcal{D}_k} \mathcal{L}(\theta; \xi_k)\) for $k \in \{1,\ldots,m\}$ is continuously differentiable, and there is a constant 
\(L \ge 0\) such that:
\begin{align}
\|\nabla \mathcal{L}_k(\theta) - \nabla \mathcal{L}_k(\vartheta)\|
\le
L\|\theta - \vartheta\|,
\quad
\forall\, \theta,\vartheta \in \mathbb{R}^d.
\end{align}
\end{assumption}

\begin{theorem}[Non-convex Convergence Rate of DSGD \citep{pmlr-v119-koloskova20a}]
Under \cref{ass: mixing}, \cref{ass: l-smooth} and \cref{ass: bounded_noise_diversity}, let the learning rate $\eta$ satisfy 
$\eta \leq \eta_{\max}\;=\;\mathcal{O} \bigl(\tfrac{p}{L}\bigr)$ let 
$\bar{\theta}^{(t)} 
= \tfrac{1}{m}\sum_{k=1}^m \theta_{k}^{(t)}$  
denote the averaged model at the $t$-th step.
To achieve an \(\varepsilon\)-stationary point such that 
\(\frac{1}{T}\sum_{t=0}^{T-1} \mathbb{E}\bigl[\|\nabla \mathcal{L}(\bar{\theta}^{(t)})\|_{2}^2\bigr] \le \varepsilon\), 
the total number of steps \(T\) satisfies:
\[
  T = \mathcal{O}\Bigl(
    \frac{\sigma^2}{m\,\varepsilon^2} +
    \frac{\sqrt{p}\,\sigma + \zeta}{p\,\varepsilon^{3/2}} +
    \frac{1}{p\varepsilon}\Bigr)
  \cdot
  \Bigl(\mathcal{L}\bigl(\theta_{0}\bigr) - \mathcal{L}^{\star}\Bigr).
\]
\end{theorem}

We then provide our main theoretical results as follows.

\begin{tcolorbox}[notitle, rounded corners, colframe=middlegrey, colback=lightblue, 
       boxrule=2pt, boxsep=2pt, left=0.2cm, right=0.2cm, top=0.2cm, bottom=0.2cm, enhanced, 
       toprule=2pt
    ]
\begin{theorem}[Non-convex Convergence Rate of DSGD]
\label{thm:nonconvex-convergence-descent-lemma-appendix}
Suppose \cref{ass: regularity} and \cref{ass: bounded_noise_diversity} hold. 
Consider decentralized SGD (DSGD) with initializations \(\theta_k^{(0)} = \theta^{(0)}\) for all \(k \in \mathcal{V}\), and a constant learning rate satisfying \(\eta < \tfrac{2}{L_2}\). 
Let \(\bar{\theta}^{(t)} = \tfrac{1}{m}\sum_{k=1}^m \theta_{k}^{(t)}\) denote the averaged model at the \(t\)-th step. 
To achieve an \(\varepsilon\)-stationary point such that 
\(\frac{1}{T}\sum_{t=0}^{T-1} \mathbb{E}\bigl[\|\nabla \mathcal{L}(\bar{\theta}^{(t)})\|_{2}^2\bigr] \le \varepsilon\), 
the total number of steps \(T\) satisfies:
\[
  T 
  =\mathcal{O}\Bigl(
    \tfrac{\sigma^2}{m \varepsilon^2}+
    \tfrac{1}{\varepsilon}
    +\colorbox{cyan!10}{$\sum_{t=0}^{T-1}U^{(t)}$}
  \Bigr)
  \cdot
  \bigl(\mathcal{L}(\theta^{(0)}) - \mathcal{L}^{\star}\bigr),
\]
where  \(U^{(t)} = \frac{1}{2}(\eta L_2 - 1)\nabla\mathcal L(\bar\theta^{(t)})^\top \nabla\operatorname{Tr}\bigl(\nabla^2\mathcal L(\bar\theta^{(t)})\,\Gamma^{(t)}\bigr) + \Theta(\Xi_{t}^3)\),
with  $\Gamma^{(t)} = \tfrac{1}{m}\sum_{k=1}^m (\theta_k^{(t)} - \bar{\theta}^{(t)})(\theta_k^{(t)} - \bar{\theta}^{(t)})^\top$ and the consensus distance \(\Xi^2_t = \Tr(\Gamma^{(t)})\).
\end{theorem}
\end{tcolorbox}

\begin{proof}

We structure the proof into several key steps.

\textbf{Step (A): Descent Force Decomposition.}

Based on the \(L_2\)-smoothness (\cref{ass: l-smooth}) of the loss function \(\mathcal{L}\) (as implied by \cref{ass: regularity})), we can apply the first-order Taylor expansion around \(\bar{\theta}^{(t)}\) to establish an upper bound for \(\mathcal{L}(\bar{\theta}^{(t+1)})\):
\begin{align*}
\mathcal{L}\big(\bar{\theta}^{(t+1)}\big)
&\leq
\mathcal{L}\big(\bar{\theta}^{(t)}\big)
+
\nabla \mathcal{L}\big(\bar{\theta}^{(t)}\big)^{\top}
\big(\bar{\theta}^{(t+1)} - \bar{\theta}^{(t)}\big)
+
\tfrac{L_2}{2}\,\big\|\bar{\theta}^{(t+1)} - \bar{\theta}^{(t)}\big\|^{2}.
\end{align*}

According \cref{prop: implicit_bias}, we have
\[
\mathbb{E}_{\xi^{(t)}}[\bar{\theta}^{(t+1)} ]
=
\bar{\theta}^{(t)}
-
\eta\, \big(\nabla\mathcal{L}\big(\bar{\theta}^{(t)}\big)+\frac{1}{2}\nabla\operatorname{Tr}(\nabla^2 \mathcal{L}(\bar{\theta}^{(t)}) \Gamma^{(t)})\big) + {\delta}^{(t)},
\]
where
$\Gamma^{(t)}$ denotes the variance matrix of $\epsilon^{(t)} \sim \mathcal{N}(0, {\Gamma}^{(t)})$ and ${\delta}^{(t)} = \Theta\left( \frac{\eta}{m}\sum_{k=1}^{m}\|\theta_k^{(t)} - \bar{\theta}^{(t)}\|_2^3\right)$ denotes the high-order residuals (see \cref{prop: implicit_bias}).

Substituting this into the previous bound and taking the expectation with respect to random data sampling yields:
\begin{align*}
&\mathbb{E}_{\xi^{(t)}}[\mathcal{L}\big(\bar{\theta}^{(t+1)}\big)]\\
&\leq
\mathcal{L}\big(\bar{\theta}^{(t)}\big)
-
\eta\,
\nabla \mathcal{L}\big(\bar{\theta}^{(t)}\big)^{\top}\,
\left(\nabla\mathcal{L}\big(\bar{\theta}^{(t)}\big)+\frac{1}{2}\nabla\operatorname{Tr}(\nabla^2 \mathcal{L}(\bar{\theta}^{(t)}) \Gamma^{(t)})- {\delta}^{(t)}\right)
+
\mathbb{E}_{\xi^{(t)}}\tfrac{L_2}{2}\,
\big\|\bar{\theta}^{(t+1)}-\bar{\theta}^{(t)}\big\|^{2}.
\end{align*}
According to \cref{coro: distance}, we obtain
\begin{align*}
 \mathbb{E}_{\xi^{(t)}}\,
 \big\|\bar{\theta}^{(t+1)}-\bar{\theta}^{(t)}\big\|^{2} \leq
 \frac{\eta^2\sigma^2}{m} + \eta^2 \big\| \nabla\mathcal{L}\big(\bar{\theta}^{(t)}\big)+\frac{1}{2}\nabla\operatorname{Tr}(\nabla^2 \mathcal{L}(\bar{\theta}^{(t)}) \Gamma^{(t)})
+ {\delta}^{(t)}\big\|^2.
\end{align*}

We can then decompose the squared norm:
\begin{align*}
&\big\|\nabla\mathcal{L}\big(\bar{\theta}^{(t)}\big)+\frac{1}{2}\nabla\operatorname{Tr}(\nabla^2 \mathcal{L}(\bar{\theta}^{(t)}) \Gamma^{(t)})\big\|^{2}\\
& = \frac{1}{4}\big\|\nabla\operatorname{Tr}(\nabla^2 \mathcal{L}(\bar{\theta}^{(t)}) \Gamma^{(t)})\big\|^{2}
+ \big\|\nabla \mathcal{L}\big(\bar{\theta}^{(t)}\big)\big\|^{2} + {\nabla\,\operatorname{Tr}(\nabla^2 \mathcal{L}(\bar{\theta}^{(t)}) \Gamma^{(t)})}^\top\mathcal{L}\big(\bar{\theta}^{(t)}\big).
\end{align*}

Combining the previous steps, we obtain:
\begin{align}
\mathbb{E}_{\xi^{(t)}}\,\mathcal{L}\big(\bar{\theta}^{(t+1)}\big)
&\leq
\mathcal{L}\big(\bar{\theta}^{(t)}\big)
- (\eta-\tfrac{\eta^{2}L_2}{2})
\,\Big\|\nabla \mathcal{L}\big(\bar{\theta}^{(t)}\big)\Big\|^{2}
+
\frac{\eta^{2}L_2}{8}\,\underbrace{\big\|\nabla\,\operatorname{Tr}(\nabla^2 \mathcal{L}(\bar{\theta}^{(t)}) \Gamma^{(t)})\big\|^{2}}_{T_1} \nonumber\\
&\quad
{\frac{1}{2}(-\eta + \eta^{2}L_2})\cdot\,\underbrace{\nabla \mathcal{L}\big(\bar{\theta}^{(t)}\big)^{\top}\,\nabla\,\operatorname{Tr}(\nabla^2 \mathcal{L}(\bar{\theta}^{(t)}) \Gamma^{(t)})}_{T_2}+ \eta\,
\underbrace{\nabla\mathcal{L}\big(\bar{\theta}^{(t)}\big)^{\top}\delta^{(t)} }_{T_3}+\frac{\eta^2\sigma^2}{m}\cdot\frac{L_2}{2}\nonumber\\
&\quad + \eta^2L_2\,\underbrace{\big(\nabla\mathcal{L}\big(\bar{\theta}^{(t)}\big)+\frac{1}{2}\nabla\operatorname{Tr}(\nabla^2 \mathcal{L}(\bar{\theta}^{(t)}) \Gamma^{(t)})\big)^{\top}{\delta}^{(t)}}_{T_4}
+ \frac{\eta^2L_2}{2}\underbrace{\big\|{\delta}^{(t)}\big\|^2}_{T_5} .
\label{eq: decomposition}
\end{align}

We subsequently control terms related to $\mathbb{E}_{\epsilon^{(t)} \sim \mathcal{N}(0, {\Gamma}^{(t)})}\nabla \mathcal{L}\big(\bar{\theta}^{(t+\frac{1}{2})})\big)- \nabla \mathcal{L}\big(\bar{\theta}^{(t)}\big)$ in \cref{eq: decomposition}.

\textbf{Step (B): Control Consensus-Related Terms}

Combining the residual upper bound in \cref{coro:gradient_diff} and with the concavity of ${(\cdot)}^{3/2}$, we can derive
\begin{align*}
\big\|{\delta}^{(t)}\big\|\leq \frac{L_4}{24}\cdot\frac{1}{m}\sum_{k=1}^{m}\big\|\theta_k^{(t)} - \bar{\theta}^{(t)}\big\|^3
\leq \frac{L_4}{24}\cdot\sqrt{m} \big(\frac{1}{m}\sum_{k=1}^{m}\big\|\theta_k^{(t)} - \bar{\theta}^{(t)}\big\|^2\big)^{\frac{3}{2}},
\end{align*}
and thus $T_3\leq \frac{L_1 L_4}{24}\cdot\sqrt{m} \big(\frac{1}{m}\sum_{k=1}^{m}\big\|\theta_k^{(t)} - \bar{\theta}^{(t)}\big\|^2\big)^{\frac{3}{2}}$.

Further, by the convexity of square operation, we have
\begin{align*}
T_5=\big\|{\delta}^{(t)}\big\|^2\leq \frac{m L_4^2} {24^2}\cdot\big(\frac{1}{m}\sum_{k=1}^{m}\big\|\theta_k^{(t)} - \bar{\theta}^{(t)}\big\|^2\big)^{3}.
\end{align*}
According to \cref{ass: regularity}, we can upper-bound $T_1$ as
\begin{align*}
T_1 &= \big\|\nabla\,\operatorname{Tr}(\nabla^2 \mathcal{L}(\bar{\theta}^{(t)}) \Gamma^{(t)})\big\|^2 
= \big\|\nabla^3 \mathcal{L}(\bar{\theta}^{(t)}) : \Gamma^{(t)}\big\|^2 
\leq L_3^2 \cdot \big(\frac{1}{m}\sum_{k=1}^{m}\big\|\theta_k^{(t)} - \bar{\theta}^{(t)}\big\|^2\big)^2,
\end{align*}
where the colon ($:$) represents the tensor contraction of the third-order derivative with the matrix $\Gamma$.

We can also bound $T_4$ as follows:
\begin{align*}
T_4&\leq \big\|\frac{1}{2}\nabla\operatorname{Tr}(\nabla^2 \mathcal{L}(\bar{\theta}^{(t)}) \Gamma^{(t)})+
\nabla \mathcal{L}\big(\bar{\theta}^{(t)}\big)\big\|\cdot\frac{1}{m}\sum_{k=1}^{m}\big\|\theta_k^{(t)} - \bar{\theta}^{(t)}\big\|^3\\
&\leq \big(\frac{L_3}{2} \,\frac{1}{m}\sum_{k=1}^{m}\big\|\theta_k^{(t)} - \bar{\theta}^{(t)}\big\|^2+L_1\big)\frac{1}{m}\sum_{k=1}^{m}\big\|\theta_k^{(t)} - \bar{\theta}^{(t)}\big\|^3\\
&\leq \big(\frac{L_3}{2} \,\frac{1}{m}\sum_{k=1}^{m}\big\|\theta_k^{(t)} - \bar{\theta}^{(t)}\big\|^2+L_1\big)\sqrt{m} \big(\frac{1}{m}\sum_{k=1}^{m}\big\|\theta_k^{(t)} - \bar{\theta}^{(t)}\big\|^2\big)^{\frac{3}{2}}.
\end{align*}

For clarity, we consolidate the terms involving $T_1$ to $T_2$ in \cref{eq: decomposition}:
\begin{align}\label{eq: accumulated terms}
\text{Accumulated $T$-terms}  = 
& (-\eta+ \eta^2 L_2)T_2 + \eta T_3 + \frac{\eta^2 L_2}{8}(T_1 + 8T_4 + 4T_5).
\end{align}
Substituting the \textbf{U}pper bounds for the Accumulated $T$-terms into \cref{eq: decomposition} yields the updated descent inequality:
\begin{align}\label{eq:descent-final}
\mathbb{E}_{\xi^{(t)}}\,\mathcal{L}\bigl(\bar{\theta}^{(t+1)}\bigr)
\le
\mathcal{L}\bigl(\bar{\theta}^{(t)}\bigr)
-\Bigl(\eta - \tfrac{\eta^2 L_2}{2}\Bigr)\,
\bigl\|\nabla \mathcal{L}\bigl(\bar{\theta}^{(t)}\bigr)\bigr\|^2
+  \eta\, U^{(t)}
+ \tfrac{\eta^2\sigma^2L_2}{2},
\end{align}
where $\eta\, U^{(t)}$ denotes the upper bound for \cref{eq: accumulated terms}:
\begin{align}
\eta\, U^{(t)}
 \triangleq &  (\eta^2 L_2 - \eta) T_2 + \frac{\eta \sqrt{m} L_1 L_4}{24} \Xi_{t}^{3}\nonumber\\
&+ \frac{1}{8} \eta^2 L_2 L_3^2 \Xi_{t}^4 + \frac{1}{2}\eta^2 \sqrt{m} L_2 (2L_1 + L_3 \Xi_{t}^2) \Xi_{t}^3 + \frac{\eta^2 m L_2 L_4^2}{1152} \Xi_{t}^6,\nonumber\\
= & \underbrace{(\eta^2 L_2 - \eta)T_2}_{\triangleq\, A^{(t)} \,=\, \Theta(\Xi_t^2)} + 
\underbrace{\Bigl[\bigl( \eta^2 L_2 + \frac{\eta L_4}{24} \bigr) \sqrt{m} L_1 + \frac{\eta^2}{8} L_2 L_3^2 \Xi_{t} + \frac{\eta^2}{2} \sqrt{m} L_2 L_3 \Xi_{t}^2 + \frac{\eta^2 m L_2 L_4^2}{1152} \Xi_{t}^3\Bigr]\Xi_{t}^3}_{\triangleq\,  H^{(t)}\,=\, O(\Xi_t^3)},
\end{align}
and we recall that the consensus distance $\Xi_{t}^2= \frac{1}{m}\sum_{k=1}^{m}\big\|\theta_k^{(t)} - \bar{\theta}^{(t)}\big\|^2$. To facilitate subsequent analysis, we further separate $U^{(t)}$ into an \textbf{A}cceleration term $A^{(t)}$ plus \textbf{H}igh-order terms $H^{(t)}$,

With $U^{(t)}$ serving as a unified proxy for the consensus errors, we obtain the new rate as follows.

\textbf{Step (C): Derive the Convergence Rate}

Starting from the descent inequality \cref{eq:descent-final}:

\[
\mathbb{E}_{\xi^{(t)}}\bigl[\mathcal{L}(\bar{\theta}^{(t+1)})\bigr]
\le
\mathcal{L}(\bar{\theta}^{(t)})
-\Bigl(\eta-\frac{\eta^2L_2}{2}\Bigr)\,
\bigl\|\nabla\mathcal{L}(\bar{\theta}^{(t)})\bigr\|^2
+  \eta\,U^{(t)}
+ \frac{\sigma^2}{m}\,\frac{\eta^2L_2}{2}.
\]
Taking full expectation and summing over \(t=0,\dots,T-1\), we obtain
\[
\sum_{t=0}^{T-1}\Bigl(\eta-\frac{\eta^2L_2}{2}\Bigr)
\mathbb{E}\bigl\|\nabla\mathcal{L}(\bar{\theta}^{(t)})\bigr\|^2
\le
\mathcal{L}(\theta^{(0)}) - \mathbb{E}\bigl[\mathcal{L}(\bar{\theta}^{(T)})\bigr]
+ \eta\,\sum_{t=0}^{T-1}U^{(t)}
+ \frac{\sigma^2\,\eta^2L_2\,T}{2m}.
\]
To ensure the descent property of $-\bigl\|\nabla\mathcal{L}(\bar{\theta}^{(t)})\bigr\|^2$, we have to set \(\eta-\tfrac{\eta^2L_2}{2}\ge 0\), which in turn implies that \(\eta\le\frac{\ell}{L_2}\), with $\ell< 2$.
Under this condition, and denoting
\(\Delta=\mathcal{L}(\bar{\theta}^{(0)})-\mathcal{L}^*\), we obtain
\begin{align}\label{eq: three_factors}
\frac{1}{T}\sum_{t=0}^{T-1}\mathbb{E}\bigl\|\nabla\mathcal{L}(\bar{\theta}^{(t)})\bigr\|^2
\le
 \frac{2\Delta}{(2-\ell)\,\eta\,T}
+ \frac{2}{(2-\ell)\,\eta}\frac{1}{T}\sum_{t=0}^{T-1}U^{(t)}
+ \frac{\sigma^2\,\eta L_2}{(2-\ell)\,m}.
\end{align}
To ensure this is at most \(\varepsilon\), it suffices to enforce
\begin{align}\label{eq: three_errors}
\frac{\sigma^2\,\eta L_2}{(2-\ell)\,m}\leq \frac{\varepsilon}{3},\quad \frac{2}{(2-\ell)}\frac{1}{T}\sum_{t=0}^{T-1}U^{(t)}\le\frac{\varepsilon}{3},\quad \text{and} \quad
\frac{2\Delta}{(2-\ell)\,\eta\,T} \le\frac{\varepsilon}{3}.
\end{align}
To satisfy all three conditions simultaneously, along with a stability condition $\eta \le \frac{\ell}{L_2}$, we should select $\eta$ accordingly:
\[
\eta \;\le\;\min\left\{\frac{\ell}{L_2},\;\frac{(2-\ell)m\varepsilon}{3\sigma^2 L_2}\right\},
\quad \text{and} \quad
T \geq\!\max\left\{\frac{6\,\Delta}{(2-\ell)\eta\varepsilon},\; \frac{6}{(2-\ell)\varepsilon}\sum_{t=0}^{T-1}U^{(t)}\right\}.
\]
To ensure a valid step-size $\eta$ exists, we substitute these three upper bounds into the condition for $T$. This yields three distinct lower bounds on the total number of iterations $T$ that must be satisfied. By rearranging the inequality $T\eta \ge \frac{6\Delta}{\varepsilon}$, we require:
\[
T \;\ge\;\max\left\{\frac{6\Delta L_2}{\ell\,(2-\ell)\,\varepsilon},\; \frac{18\Delta\sigma^2 L_2}{(2-\ell)^2\,m\varepsilon^2},\; \frac{6}{(2-\ell)\varepsilon}\sum_{t=0}^{T-1}U^{(t)}\right\},
\]
where the first two bounds are derived directly by substituting the first two terms from the $\min\{\cdot\}$ operation for $\eta$ into the first lower bound of $T$. 

Therefore, the total number of iterations $T$ should be large enough to satisfy all applicable lower bounds. This leads to the sufficient condition:
\[
T
= \mathcal{O}\Biggl(
\frac{\Delta}{\varepsilon}
+ \frac{\Delta\,\sigma^2}{m\,\varepsilon^2}
+ \frac{1}{\varepsilon}\sum_{t=0}^{T-1}U^{(t)}
\Biggr),
\]
This condition is sufficient to guarantee
\[
\frac{1}{T}\sum_{t=0}^{T-1}\mathbb{E}\bigl\|\nabla\mathcal{L}(\bar{\theta}^{(t)})\bigr\|_2^2
\le\varepsilon.
\]
The proof is now complete.
\end{proof}

\begin{tcolorbox}[notitle, rounded corners, colframe=middlegrey, colback=lightblue, 
       boxrule=2pt, boxsep=0pt, left=0.15cm, right=0.17cm, enhanced, 
       toprule=2pt]
\begin{proposition}\label{prop: critical_appendix}
Suppose \cref{ass: regularity} and \cref{ass: sharpening} hold. Assume  $\eta > 1/L_2$, and assume \(\|\nabla\mathcal L(\bar\theta^{(t)})\|\ge \mu_t>0\) for all \(t\). 
Consider the  matrix \(\Gamma^{(t)} = \tfrac{1}{m}\sum_{k=1}^m (\theta_k^{(t)} - \bar{\theta}^{(t)})(\theta_k^{(t)} - \bar{\theta}^{(t)})^\top\) and its trace \(\Xi_t^2=\operatorname{Tr}(\Gamma^{(t)})\).
Then, for any fixed \(m>0\), there exists \(\Xi_t^2 > 0\) such that 
\begin{inequality}\label{eq:critical_appendix}
\colorbox{cyan!10}{$U^{(t)}$}\triangleq \frac{1}{2}(\eta L_2 - 1)\underbrace{\nabla\mathcal L(\bar\theta^{(t)})^\top \nabla\operatorname{Tr}\bigl(\nabla^2\mathcal L(\bar\theta^{(t)})\,\Gamma^{(t)}\bigr)}_{=\Theta(\Xi_t^2)} + \Theta(\Xi_{t}^3)< 0.
\end{inequality}
\end{proposition}
\end{tcolorbox}


\begin{remark}
    We note that \cref{prop: critical_appendix} does not contradict \cref{eq: three_factors} when both $\Delta$ and $\sigma$ are zero. The condition $\Delta=\mathcal{L}(\bar{\theta}^{(0)})-\mathcal{L}^*=0$ implies that the models are initialized at an optimal point. In \cref{thm:nonconvex-convergence-descent-lemma-appendix}, we assume that all initializations are identical ($\theta_k^{(0)} = \theta^{(0)}, \forall k \in \mathcal{V}$), so it follows that all models begin at the same optimum. Consequently, the consensus error remains zero throughout all iterations, meaning the model covariance matrix $\Gamma^{(t)}$ is the zero matrix and its trace $\Xi_t$ is also zero. Since every component of the term $U^{(t)}$, defined as
    \begin{align}\label{eq: UAH}
    U^{(t)}
     \triangleq  & \underbrace{(\eta L_2 - 1)T_2}_{\triangleq\, A^{(t)} \,=\, \Theta(\Xi_t^2)} + 
    \underbrace{\Bigl[\bigl( \eta L_2 + \frac{ L_4}{24} \bigr) \sqrt{m} L_1 + \frac{1}{8}\eta L_2 L_3^2 \Xi_{t} + \frac{1}{2}\eta \sqrt{m} L_2 L_3 \Xi_{t}^2 + \frac{\eta m L_2 L_4^2}{1152} \Xi_{t}^3\Bigr]\Xi_{t}^3}_{\triangleq\,  H^{(t)}\,=\, O(\Xi_t^3)}.
    \end{align}
\end{remark}
\begin{proof}
The proof relies on establishing that for sufficiently small $\Xi_t$, the negative leading term $A^{(t)}$ in the decomposition of $U^{(t)}$ dominates the higher-order residual term $H^{(t)}$. Specifically, $A^{(t)}$ is of order $\Theta(\Xi_t^2)$, while $H^{(t)}$ is of order $O(\Xi_t^3)$.

\medskip\noindent\textbf{Step (A): Derive Upper Bound on $A^{(t)}$.}
Let $g^{(t)} = \nabla\mathcal L(\bar\theta^{(t)})$ and recall that
\[
T_2 \triangleq (g^{(t)})^\top\nabla\operatorname{Tr}\bigl(\nabla^2\mathcal L(\bar\theta^{(t)})\,\Gamma^{(t)}\bigr).
\]
Based on the definition of $U^{(t)}$ in \cref{eq: UAH}, we have the leading term $A^{(t)} = (\eta L_2 - 1)T_2$. To analyze $T_2$, consider the trilinear form defined on the unit sphere. Let $\delta_k^{(t)}$ be the deviation vectors such that $\Gamma^{(t)}=\frac1m\sum_{k=1}^m\delta_k^{(t)}(\delta_k^{(t)})^\top$ and $\Xi_t^2 = \frac{1}{m}\sum_{k=1}^m \|\delta_k^{(t)}\|^2$. Define the function $F(\delta)$ for $\delta \ne 0$:
\[
F(\delta)=\frac{\nabla^3\mathcal L(\bar\theta^{(t)})[\delta,\delta,g^{(t)}]}{\|g^{(t)}\|\|\delta\|^2}.
\]
Under \cref{ass: sharpening}, we have $\nabla^3\mathcal L(\bar\theta^{(t)})[\delta,\delta,g^{(t)}]<0$, implying $F(\delta) < 0$. Since the unit sphere $S=\{\delta:\|\delta\|=1\}$ is compact, $F$ attains a maximum value $M < 0$. Let $\gamma = -M > 0$. It follows that for any vector $\delta$,
\[
\nabla^3\mathcal L(\bar\theta^{(t)})[\delta,\delta,g^{(t)}]\le -\gamma\,\|g^{(t)}\|\|\delta\|^2.
\]
Substituting this bound into the summation for $T_2$:
\[
T_2 = \frac1m\sum_{k=1}^m\nabla^3\mathcal L(\bar\theta^{(t)})[\delta_k^{(t)},\delta_k^{(t)},g^{(t)}] 
\le \frac1m\sum_{k=1}^m \bigl(-\gamma\|g^{(t)}\|\|\delta_k^{(t)}\|^2\bigr) 
= -\gamma\|g^{(t)}\|\Xi_t^2.
\]
Given the gradient lower bound $\|g^{(t)}\| \ge \mu_t$, we obtain $T_2 \le -\gamma \mu_t \Xi_t^2$. Consequently, assuming the pre-condition $\eta L_2 - 1 > 0$ holds, the leading term $A^{(t)}$ satisfies:
\begin{align}\label{eq:A_bound}
A^{(t)} = (\eta L_2 - 1)T_2 \le -(\eta L_2 - 1)\gamma \mu_t \Xi_t^2.
\end{align}
This confirms that $A^{(t)}$ provides a strictly negative contribution of order $\Theta(\Xi_t^2)$.

\medskip\noindent\textbf{Step (B): Dominance over Higher-Order Residuals.}
We now show that $U^{(t)} = A^{(t)} + H^{(t)} < 0$ for small $\Xi_t$. According to \cref{eq: UAH}, the residual term is given by
\[
H^{(t)} = \underbrace{\left[ \left(\eta L_2 + \frac{L_4}{24}\right)\sqrt{m}L_1 + \frac{1}{8}\eta L_2 L_3^2 \Xi_t + \frac{1}{2}\eta \sqrt{m} L_2 L_3 \Xi_t^2 + \frac{\eta m L_2 L_4^2}{1152}\Xi_t^3 \right]}_{\triangleq P(\Xi_t)}\Xi_t^3.
\]
Here, $P(\Xi_t)$ is a polynomial in $\Xi_t$ with positive coefficients, and thus $H^{(t)} = O(\Xi_t^3)$. The condition $U^{(t)} < 0$ is equivalent to $H^{(t)} < -A^{(t)}$. Using the bound from \cref{eq:A_bound}, it suffices to show:
\[
P(\Xi_t)\Xi_t^3 < (\eta L_2 - 1)\gamma \mu_t \Xi_t^2.
\]
For $\Xi_t > 0$, we divide both sides by $\Xi_t^2$, reducing the condition to:
\[
\Xi_t \cdot P(\Xi_t) < (\eta L_2 - 1)\gamma \mu_t.
\]
Let $Q(u) = u \cdot P(u)$ for $u \ge 0$. Since $P(u)$ is bounded in a neighborhood of zero, we have:
\[
\lim_{u \to 0^+} Q(u) = 0 \cdot \left[\left(\eta L_2 + \frac{L_4}{24}\right)\sqrt{m}L_1\right] = 0.
\]
The term on the right-hand side, $C \triangleq (\eta L_2 - 1)\gamma \mu_t$, is a strictly positive constant. By the continuity of polynomial functions and the intermediate value theorem, there exists a threshold $\delta > 0$ such that for all $0 < \Xi_t < \delta$, the inequality $Q(\Xi_t) < C$ holds. 
This implies that for a sufficiently small consensus error $\Xi_t$, the negative drift from $A^{(t)}$ dominates the residual $H^{(t)}$, ensuring $U^{(t)} < 0$.
\end{proof}
\textbf{Explanation.} The high-level intuition  of \cref{thm:nonconvex-convergence-descent-lemma-appendix} and \cref{prop: critical_appendix} is outlined by the following descent lemma. 
\begin{align}\label{eq:descent_lemma_dec-2-appendix}
    \mathbb{E}_{\xi^{(t)}} \left[ \mathcal{L}(\bar{\theta}^{(t+1)}) \right] \leq \, & \mathcal{L}(\bar{\theta}^{(t)}) 
     -  \underset{\textcolor{textcyan}{>0}}{\colorbox{boxcyan!70}{$\displaystyle \textcolor{textcyan}{\bigl(\eta - \frac{\eta^2 L_2}{2}\bigl)}$}} 
      \underbrace{ \left\| \nabla \mathcal{L}(\bar{\theta}^{(t)}) \right\|^2 }_{\text{Standard Descent}} \nonumber\\
    & + \underset{\textcolor{textpink}{>0}}{\colorbox{boxpink!40}{$\displaystyle \textcolor{textpink}{\bigl(\eta^2 L_2 - \eta\bigl)}$}} 
      \underbrace{ \nabla \mathcal{L}(\bar{\theta}^{(t)})^\top \nabla \text{Tr}\bigl( \nabla^2 \mathcal{L}(\bar{\theta}^{(t)}) \Gamma^{(t)} \bigl)}_{< 0, \text{ \textcolor{textpink}{progressive sharpening}}} 
     + \frac{\eta^2 L_2 \sigma^2}{2m} + \mathcal{O}(\Xi_t^3).
\end{align}
\begin{tcolorbox}[notitle, rounded corners, colframe=middlegrey, colback=lightred, 
       boxrule=2pt, boxsep=0pt, after skip=7pt, left=0.15cm, right=0.17cm, enhanced, 
       toprule=2pt,
    ]
\begin{remark}\label{remark: small_loss}
    To simultaneously satisfy this requirement $\frac{\sigma^2\eta L_2}{(2 - \ell)m} \leq \varepsilon/3$ in \cref{eq: three_errors} and the condition $\eta > 1/L_2$ needed to maintain the descent property introduced by progressive sharpening, we can appropriately scale the number of agents $m$. Specifically, for any arbitrarily small target accuracy $\varepsilon$, $\frac{\sigma^2\eta L_2}{(2 - \ell)m} \leq \varepsilon/3$ is guaranteed to hold as long as the network size satisfies 
    $m > \frac{3\sigma^2}{(2 - \ell)\varepsilon}$.
    In other words, we can achieve the same convergence rate as parallel SGD by adapting the network size $m$ to the desired target accuracy $\varepsilon$.
\end{remark}
\end{tcolorbox}

\begin{tcolorbox}[notitle, rounded corners, colframe=middlegrey, colback=lightblue, 
       boxrule=2pt, boxsep=0pt, left=0.15cm, right=0.17cm, enhanced, 
       toprule=2pt]
\begin{proposition}[Critical Consensus Edge]\label{prop: sufficient-appendix}
Suppose \cref{ass: mixing}-\cref{ass: bounded_noise_diversity} hold. 
Assume $\eta>\frac{1}{L_2}$, and the consensus error  \(\Xi_t \le 1\) for all $t$.
Then, the following condition ensures that the critical \cref{eq:critical} is satisfied:
\begin{align}\label{ineq:sufficient-appendix}
  \sqrt{\frac{24(1-p)\eta^2}{p^2}
  \bigl(\phi^2 + \sigma^2\bigr)} 
  < 
  \min \left\{ 
     \frac{(\eta L_2 - 1)\gamma^* \mu_t}{2(\eta L_2 + \frac{L_4}{24})\sqrt{m}L_1}, \quad 
     \sqrt{\frac{(\eta L_2 - 1)\gamma^* \mu_t}{2 \Sigma_{\mathrm{high}}}} 
  \right\},
\end{align}
where \(\Sigma_{\mathrm{high}} = \frac{1}{8}\eta L_2 L_3^2 + \frac{1}{2}\eta \sqrt{m} L_2 L_3 + \frac{\eta m L_2 L_4^2}{1152}\). 
Here, \(\gamma^*\) denotes the degree of progressive sharpening (see \cref{ass: sharpening}),  $\phi^2$ denotes the uniform upper bound of the averaged squared local gradient norm (i.e., \(\frac{1}{m}\sum_{k=1}^m\|\nabla\mathcal{L}_k(\theta_k^{(t)})\|^2\leq \phi^2\)), and \(\mu_t\) is the lower bound on the global gradient norm (i.e., \(\|\nabla\mathcal L(\bar\theta^{(t)})\|\ge \mu_t>0\)).
\end{proposition}
\end{tcolorbox}
\begin{proof}
The proof establishes that the condition in \cref{ineq:sufficient} suffices to guarantee $U^{(t)} < 0$. Recall from the decomposition in \cref{eq: UAH} that $U^{(t)} = A^{(t)} + H^{(t)}$, where $A^{(t)}$ is the leading descent term and $H^{(t)}$ represents higher-order residuals. We aim to show that the negative drift dominates the error, i.e., $H^{(t)} < -A^{(t)}$.

Using the lower bound on the gradient norm $\|\nabla\mathcal{L}(\bar{\theta}^{(t)})\| \ge \mu_t$ derived in \cref{prop: critical}, the leading term satisfies $A^{(t)} \le -(\eta L_2 - 1)\gamma^* \mu_t \Xi_t^2$. Let $K \triangleq (\eta L_2 - 1)\gamma^* \mu_t$. Under the sharpening regime ($\eta L_2 > 1$), we have $K > 0$. Thus, a sufficient condition for $U^{(t)} < 0$ is:
\begin{align}\label{eq:target_condition}
    H^{(t)} < K \Xi_t^2.
\end{align}

\medskip\noindent\textbf{Step (A): Bounding the residual term.}
Substituting the expansion of $H^{(t)}$ from \cref{eq: UAH} into \cref{eq:target_condition} and dividing both sides by $\Xi_t^2$ (assuming $\Xi_t > 0$), the requirement becomes:
\[
    \underbrace{\left[ \left(\eta L_2 + \tfrac{L_4}{24}\right)\sqrt{m}L_1 + \tfrac{1}{8}\eta L_2 L_3^2 \Xi_t + \tfrac{1}{2}\eta \sqrt{m} L_2 L_3 \Xi_t^2 + \tfrac{\eta m L_2 L_4^2}{1152}\Xi_t^3 \right]}_{\triangleq P(\Xi_t)}\Xi_t < K.
\]
We define $C_{\mathrm{lin}} \triangleq (\eta L_2 + \frac{L_4}{24})\sqrt{m}L_1$ as the coefficient of the linear term. Utilizing the assumption that the consensus error is locally bounded ($\Xi_t \le 1$), we have $\Xi_t^k \le \Xi_t$ for $k \ge 1$. This allows us to upper bound the higher-order polynomial terms using the aggregated coefficient $\Sigma_{\mathrm{high}}$ defined in the proposition:
\[
    P(\Xi_t)\Xi_t \le C_{\mathrm{lin}} \Xi_t + \Sigma_{\mathrm{high}} \Xi_t^2.
\]
Therefore, it suffices to ensure $C_{\mathrm{lin}} \Xi_t + \Sigma_{\mathrm{high}} \Xi_t^2 < K$.

\medskip\noindent\textbf{Step (B).}
To satisfy the inequality above, we employ a budget splitting strategy, requiring both the linear and quadratic components to be bounded by half of the descent budget $K/2$. This yields two separate constraints on $\Xi_t$:
\[
    C_{\mathrm{lin}} \Xi_t < \frac{K}{2} \implies \Xi_t < \frac{K}{2C_{\mathrm{lin}}}, 
    \quad \text{and} \quad
    \Sigma_{\mathrm{high}} \Xi_t^2 < \frac{K}{2} \implies \Xi_t < \sqrt{\frac{K}{2\Sigma_{\mathrm{high}}}}.
\]
Consequently, if $\Xi_t < \min \left\{ \frac{K}{2 C_{\mathrm{lin}}}, \sqrt{\frac{K}{2 \Sigma_{\mathrm{high}}}} \right\}$, then $U^{(t)} < 0$ holds.

Finally, invoking \cref{coro: consensus_distance}, which bounds the consensus error as $\Xi_t \le \sqrt{\frac{24(1-p)\eta^2}{p^2}(\phi^2 + \sigma^2)}$, we see that \cref{ineq:sufficient} ensures $\Xi_t$ falls within the safety region. This completes the proof.
\end{proof}